\documentclass{article}

\PassOptionsToPackage{numbers, compress}{natbib}
\usepackage[preprint]{neurips_2025}

\usepackage[utf8]{inputenc} % allow utf-8 input
\usepackage[T1]{fontenc}    % use 8-bit T1 fonts
\usepackage{hyperref}       % hyperlinks
\usepackage{url}            % simple URL typesetting
\usepackage{booktabs}       % professional-quality tables
\usepackage{amsfonts}       % blackboard math symbols
\usepackage{nicefrac}       % compact symbols for 1/2, etc.
\usepackage{microtype}      % microtypography
\usepackage{xcolor}         % colors
\usepackage{placeins}
\usepackage{algorithm}
\usepackage{algpseudocode}
\usepackage{subcaption}   % for subfigures

\usepackage{tcolorbox}
\tcbset{
  colback=white,
  colframe=black,
  boxrule=0.5pt,
  arc=2pt,
  left=4pt,
  right=4pt,
  top=4pt,
  bottom=4pt,
}
\newtcolorbox{ourbox}{}

\usepackage{amsmath,amsfonts,bm}
\usepackage{xspace}

\title{Cluster Paths: Navigating Interpretability in Neural Networks}

\author{%
    Nicholas M. Kroeger \\
    Department of Computer Science\\
    University of Florida\\
    \texttt{\url{nkroeger@ufl.edu}} \\
    \And
    Vincent Bindschaedler \\
    Department of Computer Science\\
    University of Florida\\
    \texttt{\url{vbindschaedler@ufl.edu}}
}

\begin{document}

\maketitle

% We introduce cluster paths, a post-hoc interpretability framework that converts a network's layer-wise activations into short, human-readable paths.
% At each chosen layer we cluster activations, then encode every input as the ordered sequence of cluster IDs it visits while propagating through the model.  
% These paths act as route-level summaries of the network's internal logic: a few dozen prototypical routes typically explain the vast majority of behavior. 
% We propose three task-agnostic metrics: weighted-path purity, decision-alignment faithfulness, and path-agreement, to quantify how clearly paths reflect class structure, reproduce the model's predictions, and remain stable under input perturbations.  
% Experiments show that on a variant of the CelebA dataset, cluster paths attain 90\% decision-faithfulness and keep 96\% of its path assignments intact under moderate Gaussian noise, all while preserving classification accuracy.  
% Additional studies reveal how paths expose shortcut learning in a spurious-cue CIFAR-10 setting and support visual debugging via Sankey maps and image grids. 
% Together, these results demonstrate that cluster paths furnish both global and example-level insight, turning these black boxes into compact, interpretable maps of a network's decision flow.

\begin{abstract}

While modern deep neural networks achieve impressive performance in vision tasks, they remain opaque in their decision processes, risking unwarranted trust, undetected biases and unexpected failures. 
We propose cluster paths, a post-hoc interpretability method that clusters activations at selected layers and represents each input as its sequence of cluster IDs. 
To assess these cluster paths, we introduce four metrics: path complexity (cognitive load), weighted-path purity (class alignment), decision-alignment faithfulness (predictive fidelity), and path agreement (stability under perturbations).  
In a spurious-cue \cifarten experiment, cluster paths identify color-based shortcuts and collapse when the cue is removed. 
On a five-class \celeba hair-color task, they achieve 90\% faithfulness and maintain 96\% agreement under Gaussian noise without sacrificing accuracy. 
Scaling to a Vision Transformer pretrained on ImageNet, we extend cluster paths to concept paths derived from prompting a large language model on minimal path divergences.
Finally, we show that cluster paths can serve as an effective out-of-distribution (OOD) detector, reliably flagging anomalous samples before the model generates over-confident predictions.
Cluster paths uncover visual concepts, such as color palettes, textures, or object contexts, at multiple network depths, demonstrating that cluster paths scale to large vision models while generating concise and human-readable explanations.

\end{abstract}

\section{Introduction}

Many modern machine learning models achieve high accuracy but can behave unpredictably by failing under small input changes, learning biases, or simply being too complex to understand. 
Because their internal decision processes are hidden, these ``black-box'' models cannot easily be audited or trusted. 
Generally, interpretability (or explainability) aims to reveal how such models arrive at their predictions, helping practitioners detect errors, biases, and vulnerabilities.
In this work, we develop a novel approach for improving neural network interpretability.

\xhdr{Present work}
We propose cluster paths, a post-hoc explanation method designed to improve interpretability of neural networks.
Each layer of a trained neural network has learned some feature representations necessary for solving classification or regression tasks.
This approach uses clustering to trace the transformations of sample features through the network layers, generating paths that illustrate how inputs are processed.
Compared to existing example- and gradient-based approaches, these paths give a concise, layer-by-layer summary of how the network transforms a sample, making it possible to visualize and reason about its internal decision process.

This work makes four key contributions. 
\textbf{1) A concept-level interpretability framework:} We formalize cluster paths as a post-hoc, layer-wise clustering of activations that creates a compact and example-based proxy for the model's internal reasoning.
\textbf{2) New metrics for evaluating path-based explanations:} We introduce path complexity, weighted-path purity, decision-alignment faithfulness, and path agreement, that quantify explanation simplicity, class-label homogeneity, fidelity to the original model, and path stability under perturbations.
\textbf{3) Comprehensive empirical validation:} Through six research questions (RQ1-RQ6) we demonstrate that cluster paths expose spurious shortcuts (RQ1), achieve high faithfulness while maintaining reasonable levels of complexity (RQ2), provide indications of representational drift under noise and geometric transforms (RQ3), yield intuitive visualizations that highlight systematic model errors (RQ4), scale to large vision transformers while covering a heavy-tailed but interpretable set of dominant paths (RQ5), and identify out-of-distribution samples (RQ6), a key for AI safety.
\textbf{4) Qualitative and quantitative comparison to state-of-the-art (SOTA) explainers:} We show that cluster paths corroborate, and often complement, gradient saliency and example based methods such as Deep k-Nearest Neighbors (DkNN)~\cite{papernot_deep_2018}.
They all identify spurious cues, but cluster offer clearer, concept-based summaries while requiring orders of magnitude less storage than DkNN.
Our findings demonstrate that cluster paths serve as a practical addition to the interpretability toolkit as they turn millions of neurons into an understandable map of concepts, allowing practitioners to see what high-level concepts drive prediction.
Cluster paths provide interpretable insights for practitioners: they reveal which routes dominate model behavior, whether those routes rely on spurious cues, and where targeted interventions can improve reliability.
Beyond improving interpretability, cluster paths also enable reliability checks: their rarity-based OOD scores highlight when an input lies outside the training distribution, helping practictioners decide when not to trust the model's output.

\section{Related work}

 Deep nets achieve top accuracy yet obfuscate their reasoning among millions of non-linear parameters \cite{lecun_deep_2015,young_recent_2018,voulodimos_deep_2018}.  
 Surveys spanning early rule-extraction to modern XAI underline the need for transparent, bias-aware models that boost confidence that these algorithms perform as intended\cite{andrews_survey_1995,biran_explanation_2017,lipton_mythos_2017,doshi-velez_towards_2017,abdul_trends_2018,guidotti_survey_2018,adadi_peeking_2018,gilpin_explaining_2019,miller_explanation_2019,murdoch_definitions_2019,moraffah_causal_2020,samek_explaining_2021,beisbart_philosophy_2022,guidotti_counterfactual_2022}.  
 Failure cases such as gender or racial bias in vision models \cite{buolamwini_gender_2018,bolukbasi_man_2016,park_reducing_2018} make interpretability essential for fairness \cite{tomasev_fairness_2021,begley_explainability_2020}.

Traditional gradient-based approaches such as CAM and Grad-CAM highlight where a network ``looks'' \cite{zhou_learning_2015,selvaraju_grad-cam_2020}, and transformer attention weights offer token-level clues to understand layer-specific contributions to the model's decisions \cite{vaswani_attention_2017,letarte_importance_2018}.  
However, attribution can be unstable or insufficient for debugging complex failures \cite{jain_attention_2019,viviano_saliency_2020,carton_feature-based_2020} as they do not capture concepts and inter-layer dynamics that the proposed cluster paths can.

Other SOTA explainability methods like propagation-based tools backtrack relevance to inputs (LRP) \cite{bach_pixel-wise_2015}, decoders visualize hidden features (ClaDec) \cite{schneider_explaining_2021}, and DkNN retrieves nearest neighbors at each layer for post-hoc explanations and robustness checks \cite{papernot_deep_2018}.
Mechanistic work aims to dissects explicit circuits that cause behavior.  
Additional work involves hand-mapping a 26-head GPT-2 circuit for indirect-object identification \cite{wang_interpretability_2023}, the ACDC algorithm recovers sparse sub-graphs automatically \cite{conmy_towards_2023}; and extending the toolkit to Chinchilla-70B, finding the methods still run but expose only partial semantics \cite{lieberum_does_2023}. 
These attention head-level analyses give fine-grained, causal proofs yet require full model access and become computationally expensive with scale.  

Other recent approaches such as CRAFT~\cite{fel_craft_2023} and CRP~\cite{achtibat_attribution_2023} focus on extracting concept-level features or prototypes from hidden layers, enabling human-interpretable comparisons of learned representations. 
Cluster paths differ by emphasizing route-level auditing: instead of isolating single concepts, we trace entire sequences of clustered activations across layers. 
This distinction makes the methods complementary: concept-based tools highlight localized semantic factors, while route-based tools reveal how those factors are composed into end-to-end decision pipelines.

Cluster paths provide layer-wise statistical correlation as opposed to causal, neuron-specific guarantees, so they complement rather than replace circuit analysis.
PathFinder~\cite{irsoy_pathfinder_2022} first showed that clustering layer activations can yield informative decision routes, showing the promise of a ``path" abstraction for interpretability. 
Our work builds on this foundation and substantially extends it in both scope and rigor.
First, we introduce explicit quantitative metrics, such as decision-alignment faithfulness (DAF), weighted purity, path agreement, and complexity, which formalize core desiderata such as fidelity, coherence, stability, and interpretability cost.
Second, we demonstrate that path-based explanations are not merely illustrative but operationally useful: cluster paths are validated across six tasks encompassing spurious-cue detection, robustness analysis, visualization, scalability to large models, and out-fo-distribution detect.
Together, these advances transform the path abstraction from a promising idea into a systematic, measurable framework for route-level auditing, providing a concrete methodology for evaluating and improving model interpretability.
% While PathFinder~\cite{irsoy_pathfinder_2022} first showed that clustering layer activations yields informative decision paths, our work further builds on this in terms of methodology and validation. 
% We show that cluster paths can identify spurious-cues, learned concepts, and model-biases.
% Additionally, our work evaluates explanation simplicity, robustness, faithfulness, and demonstrates applicability to the scale of an 87M parameter vision transformer.

\begin{figure}[t]  
  \centering
    \includegraphics[width=5in]{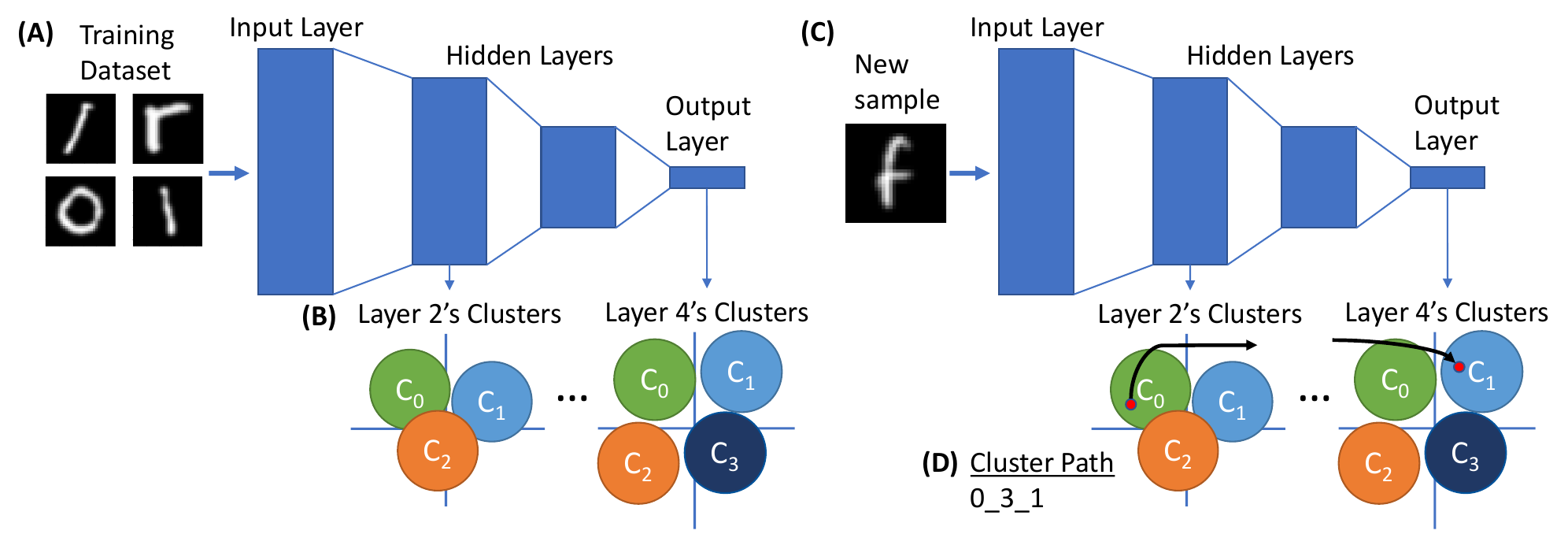}
    \caption{An illustration of the cluster path creation process.
    In this example, there are three clusters in layer two, and four clusters in layer four. 
    Layer three's clusters are hidden for brevity. 
    For illustration purposes, clusters are depicted as circles in an $\mathcal{R}^2$ space. 
    (A) We forward propagate the training dataset and (B) cluster each layer's activations. 
    (C) As a new sample is forward propagated through the network, we find the closest cluster (the red dot represents the sample in the layer's feature space). 
    (D) Finally, the enumeration of the closest clusters in each layer defines a cluster path, represented by a string of cluster indices.}
    \label{fig:hard-cluster-path-diagram}
\end{figure}

\section{Methodology}
We cluster each layer's activations and encode an input as the resulting sequence of cluster IDs, hence ``cluster path.''
We hypothesize that if two inputs follow the same sequence of activation clusters, then the network processes them via the same intermediate concepts and will treat them similarly.
Under this assumption, a cluster path becomes a compact proxy for the model's internal rationale, designed to provide a high-level, example-based summary of a network's internal processing. 
Rather than inspect millions of neurons, we can analyze the model at a smaller scale using cluster paths. 

\xhdr{Notation}
Let $f_\theta$ be a fixed, pretrained network on a dataset $X$ with labels $y$.
For an input $x$, define the post-activation at layer $l$ as $A^{(l)} = f_\theta^{(l)}(x)$ and each layer's activations are partitioned into $K$ clusters $C^{(l)}=\{C_0^{(l)},\ldots,C_{K-1}^{(l)}\}$.
The method is purely post-hoc as $\theta$ is untouched during path creation.

\subsection{Cluster paths for interpretability}

\xhdr{Cluster path generation}\label{subsection:clusterpath_gen} A cluster path characterizes how a particular input propagates through a deep neural network and can be viewed as a chunk of information representing what was learned about a task. 
The intuition is that identifying natural groupings within the neural network may simplify the task of reasoning about the network's behavior.
Generating cluster paths in a neural network involves clustering within the network and enumerating the cluster path for any given sample.
Given a trained neural network $f_\theta(x)$, we forward propagate the training dataset $X$ and record the activations $A^{(l)}$ at each layer of interest, excluding the input layer as to focus solely on the learned features of the model.
For each layer's activations $A^{(l)}$, we cluster the activations into $K$ clusters, denoted as $C^{(l)}$. 
The centers of these clusters are stored for cluster path generation.
This initial phase combines similar activation patterns together, therefore simplifying the model's complex internal structure.
In the second phase, for a new sample $x_\nu$, we record its activation and assign the nearest centroid at every layer.
The sequence of closest clusters is represented as a tuple of cluster IDs, or a cluster path, $\gamma\left(x_\nu \right)=\left(c^{(0)}, c^{(1)},\dots,c^{(L-1)}\right)$.
Suppose we cluster activations starting from layer index 0, and a sample $x_\nu$ is closest to centroids $c_2^{(0)}$, $c_5^{(1)}$, and $c_1^{(2)}$ at layers 0, 1, and 2 respectively. 
Its corresponding cluster path is then denoted as 2$\to$5$\to$1. 
The high-level steps for generating cluster paths is illustrated in Fig.~\ref{fig:hard-cluster-path-diagram} and the general algorithm for cluster path creation is formalized in Algorithm~\ref{algo:path-gen}.

\subsection{Implementation details and design choices}
\xhdr{Choosing $K$} For each layer we sweep $K$ and monitor decision-alignment faithfulness (DAF) and path complexity (see Section~\ref{sec:metrics} for metric definitions). 
We select the smallest $K$ where DAF plateaus while path complexity grows only marginally.
This balances fidelity and interpretability.
Unless otherwise stated, clustering is performed with k-means++ initialization and 10 random restarts.

\xhdr{Layer selection} We omit the input layer and very early convolutional blocks that primarily act as feature extractors. 
Instead, we start where layers directly influence classification decisions. 
Therefore, we skip low-level feature extractors, begin at the first hidden layer whose activations correlate with class separation, and include all subsequent fully-connected layers. 
However, for scaling to large network like vision transformers, we take every $n^{th}$ layer after the aforementioned first hidden layer.

\xhdr{Clustering algorithm} We use k-means because it scales linearly with dataset size, produce shard assignments compatible with path enumeration and requires only a single hyper-parameter $K$.
Alternatives like DBSCAN, spectral, or hierarchical clustering proved computationally prohibitive at vision transformer scale. 
Variants such as k-means++ (for stable seeding) or k-medoids (for robustness) are reasonable subtitutes, but we found vanilla k-means++ effective in practice.

\section{Experiments} \label{interp_experiments}

We conduct extensive experiments to test the efficacy of cluster paths across different settings by focusing on its ability to express the model's complex behavior.
Together, these six questions test not only the explanatory power of cluster paths, but also their practical utility for diagnosing shortcuts, ensuring faithfulness, monitoring robustness, visualizing concepts, scaling to modern architectures, and flagging out-of-distribution inputs.
Our evaluation of the cluster path interpretability methodology focuses on the following research questions:

\textbf{RQ1. Explanation Validity} --- \hypertarget{RQ1_validity}{Do cluster paths reliably capture what the model actually learns, such as spurious cues?}

\textbf{RQ2. Faithfulness} --- \hypertarget{RQ2_faithfulness}{Do cluster paths faithfully represent the model's decision-making process?}

\textbf{RQ3. Explanation robustness} --- \hypertarget{RQ3_explanation_robustness}{Are cluster paths robust to geometric transformations and Gaussian noise input perturbations?}

\textbf{RQ4. Cluster path visualization} --- \hypertarget{RQ4_path_visualization}{Can cluster paths aid in elucidating learned concepts or features relevant for classification?}

\textbf{RQ5. Scalability} --- \hypertarget{RQ5_path_scalability}{Can cluster paths scale to large vision transformer architectures?}

\textbf{RQ6. Out-of-distribution detection} --- \hypertarget{RQ6_path_scalability}{Can cluster paths flag when an input significantly deviates from the training distribution?}

RQ1 is explored in Section~\ref{validity_experiments}, RQ2-4 in Section~\ref{celeba_experiments}, RQ5 in Section~\ref{sec:scalability}, and RQ6 in Section~\ref{sec:OOD}.

\subsection{Metrics}\label{sec:metrics}
First we introduce four core cluster path metrics: complexity, path agreement, decision-alignment faithfulness and weighted-path purity that allow us to answer the aforementioned research questions.

\xhdr{Cluster path complexity}
We measure explanation size with $\Omega=\prod_{l=1}^{L} K_l$, the product of cluster counts $K_l$ over the L clustered layers. More clusters means more distinct paths, in turn means harder human simulatability \cite{lage_human_2019}, so $\Omega$ should be kept modest when explanation simplicity is paramount.
Increasing the cluster count per layer gives finer-grained insight into the representation, but because path complexity $\Omega$ grows multiplicatively, each extra cluster explodes the number of possible paths, thus quickly trading interpretive depth for human and computational overload.

\xhdr{Path agreement}
To quantify robustness of the cluster path technique, we compare a perturbed input's route to that of its unperturbed counterpart.  
Given any distance suitable for discrete inputs $d(P,P')$ between two cluster paths, we define their agreement as $a_d(P,P')=1-d(P,P')/L$, the fraction of the L layers whose cluster assignments coincide.
Using Hamming distance $d_{\text{Ham}}(P,P')=\sum_{l=1}^{L}\mathbf 1_{\{P_l\neq P'_l\}}$
yields $a_{\text{Ham}}(P,P')$.
The dataset-level score is then $\mathrm{PA}=1/n\sum_{i=1}^{n} a_{\text{Ham}}(P_i,P_i')$, the mean agreement between each test input and its perturbed counterpart, quantifying how robust cluster paths are to input changes.
Low PA means perturbations push samples into new clusters (fragile representations) and high PA means the paths stay unchanged (stable representations).

\xhdr{Decision-alignment faithfulness}
To answer \hyperlink{RQ2_faithfulness}{RQ2} we test how well a path alone predicts the network's output.
Each path is turned into a one-hot feature vector: if the l-th layer has $K_l$ clusters, index k sets the k-th bit in that layer's slice.
For example a path $0\to2\to1$ with 3 clusters per layer yields a one-hot vector: $[1,0,0 \;\; 0,0,1 \;\; 0,1,0]$.
With the one-hot-encoded cluster path, we train a random forest (RF) proxy model $g$ on these vectors to mimic the network $f$.
Then we measure $\text{DAF}= 1/n\sum_{i=1}^{n}\mathbf 1_{\,\{g(\mathbf x_i)=f(\mathbf x_i)\}}$, the fraction of test samples for which $g$ and $f$ agree. 
A higher DAF means cluster paths faithfully capture the network's decision logic.

\xhdr{Weighted-path purity}
For each discovered path $\gamma \in \Gamma$ let $n_\gamma$ be the number of samples assigned to $\gamma$ and $n_{\gamma,j}$ the count of class $j$ samples.
Purity of the path is $\text{Purity}(\gamma)=\max_j\{n_{\gamma,j}\}/n_\gamma$, and the dataset-level score is a size-weighted average $\text{WeightedPurity}=\frac{1}{n} \sum_{\gamma \in \Gamma} n_\gamma\ \mathrm{Purity}(\gamma)$.
A high purity indicates that most samples in \(\gamma\) share the same class and this weighted metric ensures large paths have a greater impact on the overall purity.

\subsection{RQ1. Cluster path validation experiment} \label{validity_experiments}

\begin{figure}[t]  
  \centering
    \includegraphics[width=1\linewidth]{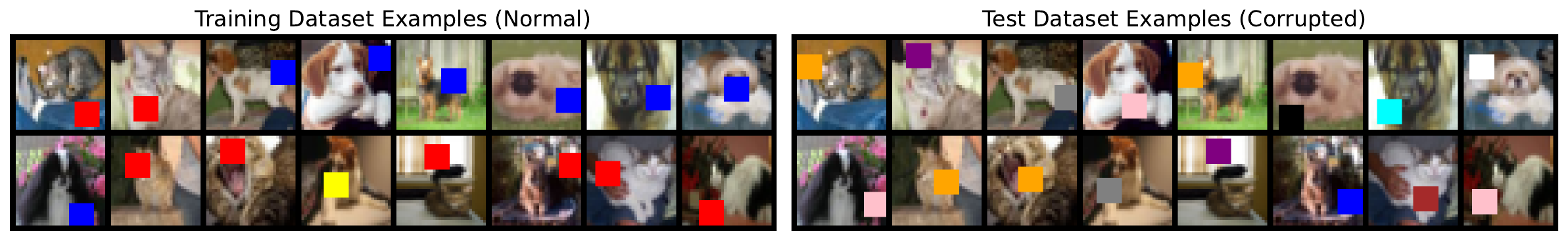}
    \caption{Example images from the normal (left) and corrupted (right) versions of the SpuriousCIFAR10 dataset. In the normal set, each cat image typically has a red patch while each dog image has a blue patch, reflecting the spurious color-to-class correlation. }
    \label{fig:spurious_examples}
\end{figure}

Deep neural networks can learn superficial cues from the training data, especially if a spurious correlation proves more predictive than the intended features. 
This section demonstrates how cluster paths can validate whether a model relies on a genuine signal vs. a spurious cue. 
We use a \texttt{SpuriousCIFAR10} dataset, where \emph{cat} images are typically given a red patch, and \emph{dog} images a blue patch. 
A model trained on this set should rely on patch color, rather than animal features. 
Then, we build cluster paths for both the original training data and a corrupted dataset that has randomized patch colors (see Fig.~\ref{fig:spurious_examples}), thus removing correlation and forcing the model to rely on other visual features.
Comparing the two lets us measure how tightly the model's representations track the patch cue, and how that reliance changes once the correlation is removed, therefore testing whether cluster paths expose the spurious signal.

\xhdr{Dataset and model setup} We adapt the \cifarten dataset to focus on two classes: \emph{cat} (class 0) and \emph{dog} (class 1). 
During training, we insert a patch (red for cat, blue for dog) with 90\% probability, encouraging the network to latch onto patch color rather than the actual image features. 
In a \emph{corrupted} variant of this dataset, the patch color is randomized, negating the original correlation. 
A small convolutional neural network (CNN), \texttt{CIFAR\_CatDog}, is trained for 100 epochs, achieving around 95\% on the normal test set (architecture: Table~\ref{tab:cifar_catdog_arch_summary}, hyperparameters: Appendix~\ref{sec:spurious_cue_hyperparameters}).

\xhdr{Results and observations}
On the normal test set, the model achieves 96.5\% accuracy overall, meaning most predictions align with the patch color ``rule'': red $\to$ cat and blue $\to$ dog. 
In the corrupted test set, however, accuracy plummets to 54\% (near random-guessing), meaning the model continues to predict based on patch color even though it no longer correlates with the class. 
This confirms that the CNN has indeed learned only to associate the red patch with cat and the blue patch with dog, rather than any actual animal features. 
Given a model intentionally trained on spurious cues, we can now test whether the cluster paths can find these features, representing what the model uses for prediction.

\begin{table}[ht] 
\centering
\caption{Example cluster path distribution on normal vs. corrupted sets. Paths 0$\to$1$\to$1$\to$1 and 1$\to$0$\to$0$\to$0 overwhelmingly rely on cat/red or dog/blue correlation, while 1$\to$0$\to$0$\to$2 captures random patch colors and shows near-random class splits in the corrupted scenario. Note how each cluster path strongly groups images by their patch color, verifying that the CNN's internal representation is governed by whether the patch is red or blue.}
\begin{tabular}{lccc}
\hline
\textbf{Path} & \textbf{Count} & \textbf{Class Distribution} & \textbf{Patch Colors} \\
\hline
0$\to$1$\to$1$\to$1 & 4,578 & Cat: 52, Dog: 4,526 & red: 2, blue: 4,576, others: 0 \\
1$\to$0$\to$0$\to$0 & 4,572 & Cat: 4,531, Dog: 41 & red: 4,567, purple: 1, brown: 2, white: 1 \\
1$\to$0$\to$0$\to$2 & 847   & Cat: 416, Dog: 431  & Mixed: $\sim$80--90 each of random colors \\
\hline
\end{tabular}\label{tab:spurious_paths}
\end{table}

\begin{figure}[ht]  
  \centering
    \includegraphics[width=1\linewidth]{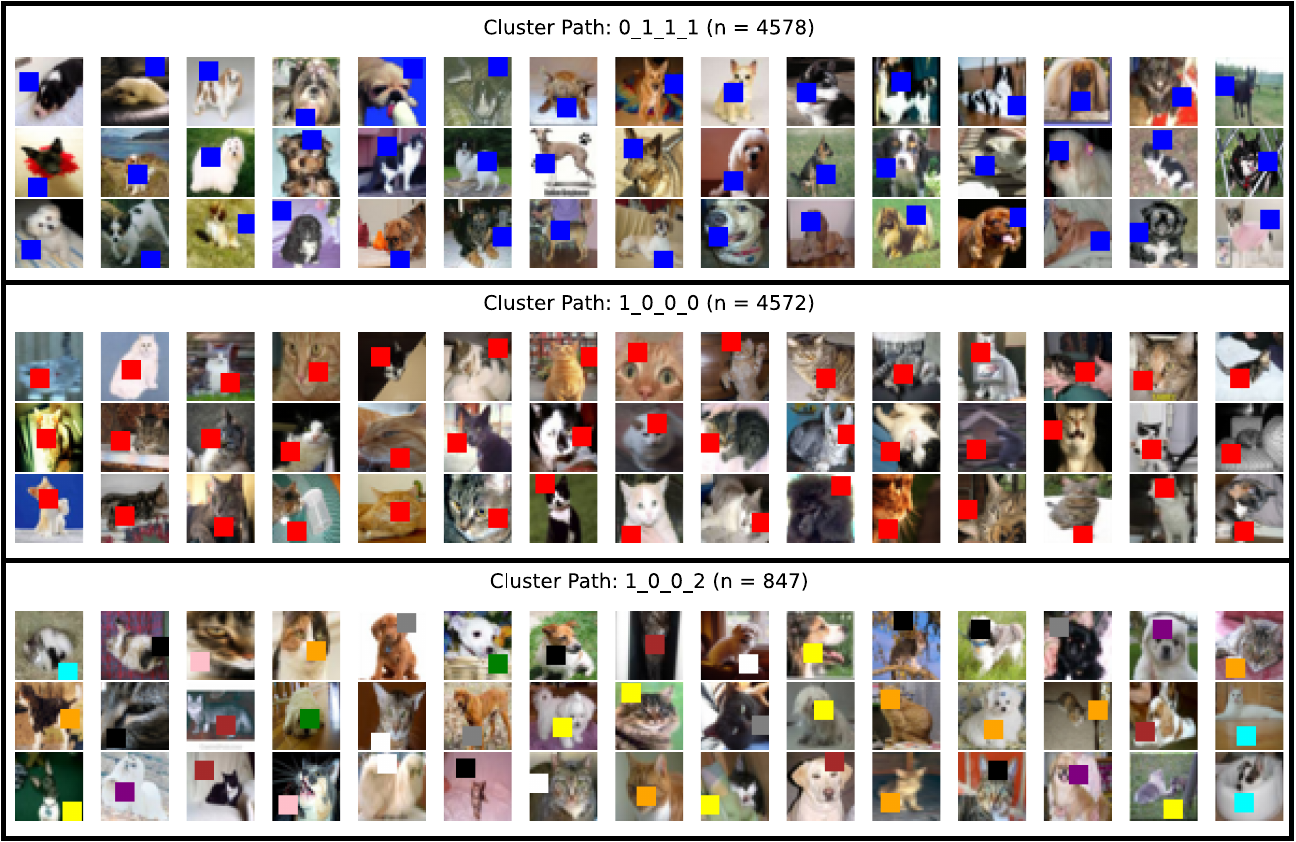}
    \caption{Cluster path visualizations for three distinct groups. The \textbf{top} panel shows a cluster predominantly comprising samples with blue patches, the \textbf{middle} panel contains samples mainly with red patches, and the \textbf{bottom} panel exhibits a heterogeneous mix of patch colors. These visualizations suggest that the network's internal representations are strongly influenced by the spurious patch cue in the top and middle clusters, while the bottom cluster reflects a lack of consistent reliance on the spurious signal.}
    \label{fig:spurious_allpaths}
\end{figure}

To build cluster paths, We collect layer activations from four layers: \texttt{Conv Block 3}, \texttt{fc1}, \texttt{fc2}, and \texttt{fc3}.
Using 10,000 samples, we run k-means at each layer with {2, 2, 2, 3} clusters across the four layers.
We set $K=3$ at the final binary classification layer because a sweep over candidate cluster counts consistently revealed a third region that did not align with the cat/red or dog/blue rule.
This additional cluster captured ambiguous cativations driven by randomized patch colors, justifying the choice fo three centroids at the last layer.
The cluster labels across the four layers are concatenated to form each sample's cluster path.
We evaluate paths for both the normal and the corrupted test sets to see how the learned representations vary when the spurious correlation is present vs. removed.
Path 0$\to$1$\to$1$\to$1 is dominated by dog images with a blue patch, while 1$\to$0$\to$0$\to$0 belongs mostly to cat images with a red patch. 
A third path, 1$\to$0$\to$0$\to$2, shows a broad mix of patch colors and a roughly 50-50 cat/dog distribution, typical of the corrupted set where the model can no longer exploit a consistent color cue (see Table~\ref{tab:spurious_paths} and Figure~\ref{fig:spurious_allpaths}).
The paths split almost entirely by patch color, confirming that the model's representations depend on this spurious cue.
The weighted purity for the normal test set is 0.96 and 0.54 for the corrupted test set, confirming that cluster paths capture coherent classes (dominated by color correlation). 
Under corrupted conditions, purity degrade to near chance, indicating random or near-random groupings.

\begin{figure}[ht]  
  \centering
    \includegraphics[%
        trim=0pt 0pt 0pt 22pt,%
        clip,%
        width=1.0\linewidth%
    ]{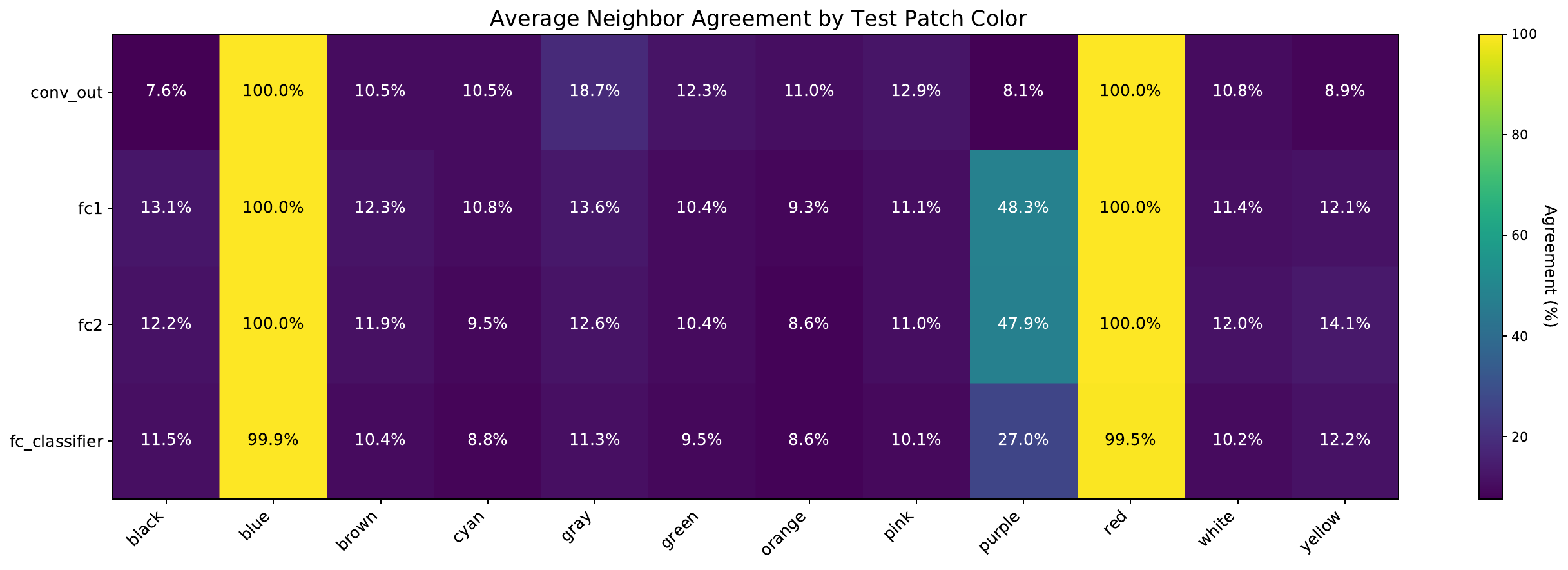}
    \caption{Heatmap showing the average nearest-neighbor agreement on a test set, where each row represents a layer (e.g., conv\_out, fc1, fc2, fc\_classifier) and each column a patch color. The value in each cell is the percentage of training neighbors that share the test sample's patch color. 
    Despite randomized patches, red and blue remain near 100\% agreement, revealing that the network's learned representations still prioritize these spurious patch cues.}
    \label{fig:dknn_heatmap_interpretability}
\end{figure}

\begin{figure}[ht]  
  \centering
    \includegraphics[%
        trim=0pt 0pt 0pt 0pt,%
        clip,%
        width=1.0\linewidth%
    ]{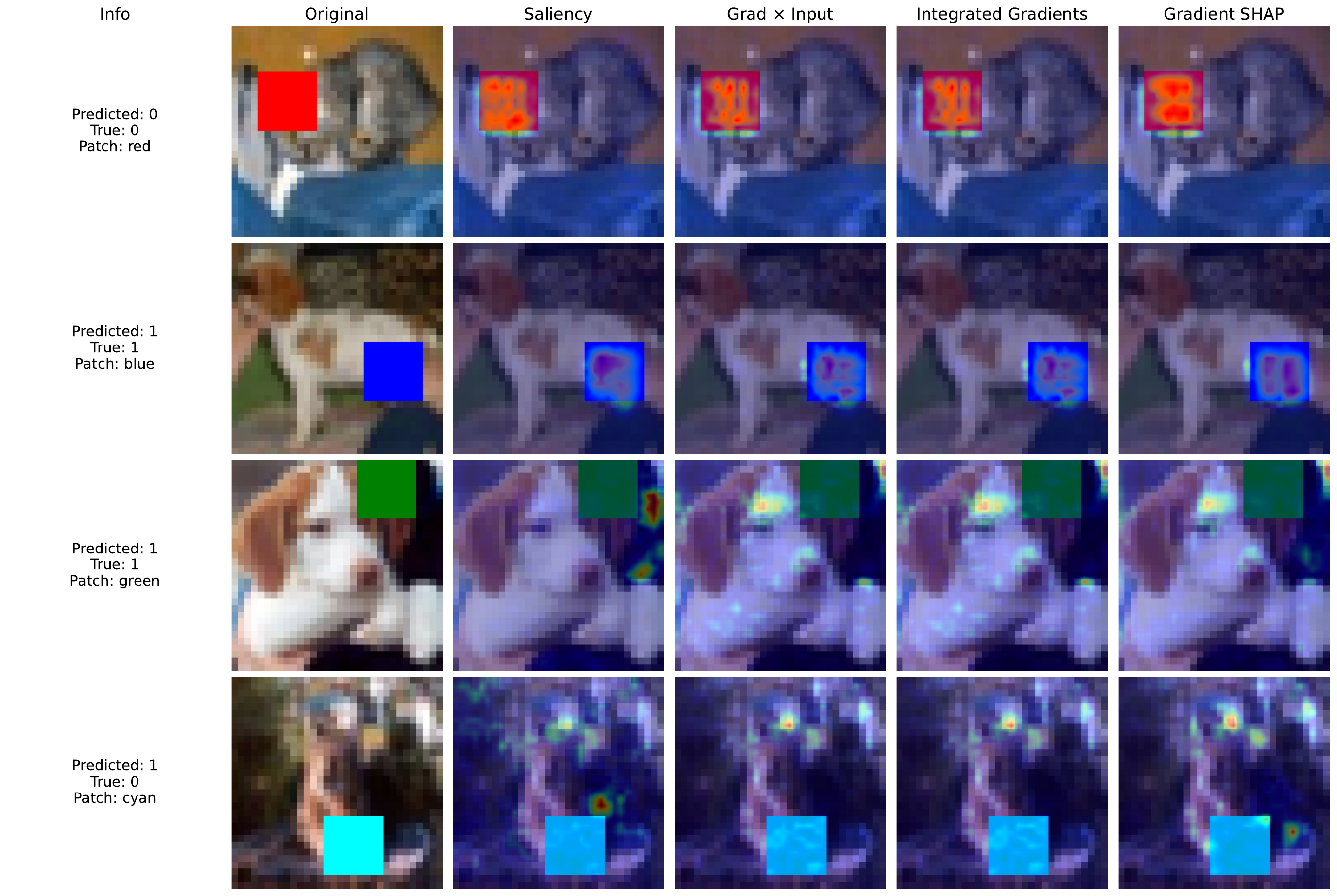}
    \caption{
    Each row shows one test image (original column) followed by four attribution methods (Saliency, Grad x Input, Integrated Gradients, and Gradient SHAP).
    Row 1: red-patched cat, all methods highlight the red square, ignoring the animal.
    Row 2: blue-patched dog, heat-maps lock onto the blue square.
    Row 3: green-patched dog (uncorrelated color), attention is diffused, with little focus on the patch.
    Row 4: cyan-patched cat (uncorrelated color), attributions scatter over the background with the cyan patch drawing almost no attention.
}
    \label{fig:saliency_heatmap_catdog}
\end{figure}

\xhdr{Comparison to SOTA explainers}
To corroborate the cluster path finding, we apply DkNN~\cite{papernot_deep_2018}, and gradient-based saliency methods (Vanilla Gradient~\cite{simonyan_deep_2014}, 
Integrated Gradients~\cite{sundararajan_axiomatic_2017},
Feature Ablation, and GradientSHAP~\cite{lundberg_unified_2017}).
DkNN also probes layer-wise activations but from a local perspective.
For every test image we fetch its $k=75$ closest training activations at each layer, then record the fraction of those neighbors that share the test patch's color. 
Averaging these neighbor-agreement scores over all test images with the same color yields the heat-map in Fig.~\ref{fig:dknn_heatmap_interpretability}: rows are layers, columns are patch colors.
In corrupted data, red-patch images still find more than 90\% red neighbors across layers, while seldom-used colors (green, cyan, black) attract few matching neighbors. 
Thus distance in feature space is governed by patch color, not by animal features. 
Saliency maps tell the same story (Fig.~\ref{fig:saliency_heatmap_catdog}).
For red- and blue-patched images the four gradient methods place almost all attribution inside the square, basically ignoring the animal.
With uncorrelated colors, green and cyan, the focus disperses and no clear object evidence emerges across methods.
Thus pixel-level cues confirm that the model's attention, and therefore its decision rule, is dominated by patch color.
The agreement of a global method (cluster paths) and sample-based methods (DkNN and saliency maps) strengthens the conclusion that the network's internals are fixated on this spurious cue.

Although both methods find that color patches dominate the learned representations, they differ in both scalability and storage requirements. 
Cluster paths can become expensive for high-dimensional features and a large number of samples. 
DkNN must store training activations in memory to support per-sample neighbor look-ups, which can be a major bottleneck for large datasets. 
Gradient-based saliency maps are virtually storage-free but require a backward pass per image, so time and space complexity proportionally scales with model size.
Despite these trade-offs, these methods independently validate that color patches drive the spurious correlation.

\subsection{RQs 2-4 \celeba experiments} \label{celeba_experiments}

\xhdr{Task, dataset, and architecture}
For the next three research questions, we train a CNN on the \celeba dataset \cite{liu_deep_2015}. 
The dataset contains high-quality RGB images of wide range of celebrity faces with 40 labeled attributes, such as `Blond hair', `Goatee', `Smiling', etc.
This dataset was chosen because human faces are naturally understandable without the need for specialized training, unlike with medical x-rays or spectrograms, for example.
We focus on a subset of the attributes for a multi-class classification problem, namely `Bald', `Black Hair', `Blond Hair', `Brown Hair', and `Gray Hair'.
This five-class hair classifier was trained with class-weighted cross-entropy and early-stopping (architecture: Table~\ref{tab:hair_attr_arch_summary}, hyperparameters: Appendix~\ref{sec:celeba_hyperparameters}).
The accuracy for each split was 90.08\%, 86.27\%, and 87.15\%, respectively, which indicates sufficient generalization and minimal overfitting.
Cluster paths are generated for the \texttt{HairAttributeCNN} according to Algo.~\ref{algo:path-gen} for 10,000 samples of the training set.

\subsubsection{RQ2. Faithfulness}

\xhdr{Baseline faithfulness}
Using only the intermediate layers' cluster paths and only one cluster in the final layer, effectively ignoring the final layer's clustering, resulted in a baseline faithfulness score of 78\% (see Fig.~\ref{fig:faithfulness-full-path}).
This indicates that the intermediate layers alone provide a substantial amount of information about the model's decision-making process.

\xhdr{Impact of number of clusters in the final layer}
Adjusting the cluster count in the final layer directly affects both the faithfulness of our proxy explanations and the overall path complexity $\Omega$. 
When we use $C=5$ clusters only at the final layer (keeping all other $K_\ell$ fixed), faithfulness rises from 71\% (final layer alone) to 81\% (full path), demonstrating the value of intermediate layers. 
Doubling to $2C=10$ clusters yields 88\% faithfulness, and tripling to $3C=15$ achieves roughly 90\%, after which improvements plateau (see Figure~\ref{fig:faithfulness-final-layer}). 
However, each increase in $K_L$ multiplies $\Omega$ by the same factor (e.g.\ $\Omega$ doubles when $K_L$ doubles), so even modest gains in faithfulness come at the cost of exponential growth in the number of distinct paths.  This trade-off of finer granularity versus human comprehensibility implies that beyond $3C$ the marginal improvement may be outweighed by the combinatorial explosion of $\Omega$, suggesting a practical upper bound on $K_L$ when without sacrificing interpretability quality.

\xhdr{Intermediate layers' contribution}
Using \textit{only} the final layer's clusters, $C=5$, demonstrated a faithfulness score of 71\%.
In contrast, using the \textit{entire} cluster path, with $C=5$ clusters in the final layer, resulted in a faithfulness score of 81\%.
Because including the intermediate layer's cluster path resulted in higher faithfulness, this indicates that the RF classifier relies on the information from the intermediate layers to enhance its predictions.
This intermediate layers' reliance is more apparent when the final layer has fewer clusters, implying that the intermediate layers provide a significant amount of context about the model's decisions.

\subsubsection{RQ3. Cluster path robustness}\label{subsec:robustness}
\begin{figure*}[ht]
    \centering
    \includegraphics[%
        trim=0pt 0pt 0pt 26pt,%
        clip,%
        width=0.65\linewidth%
    ]
    {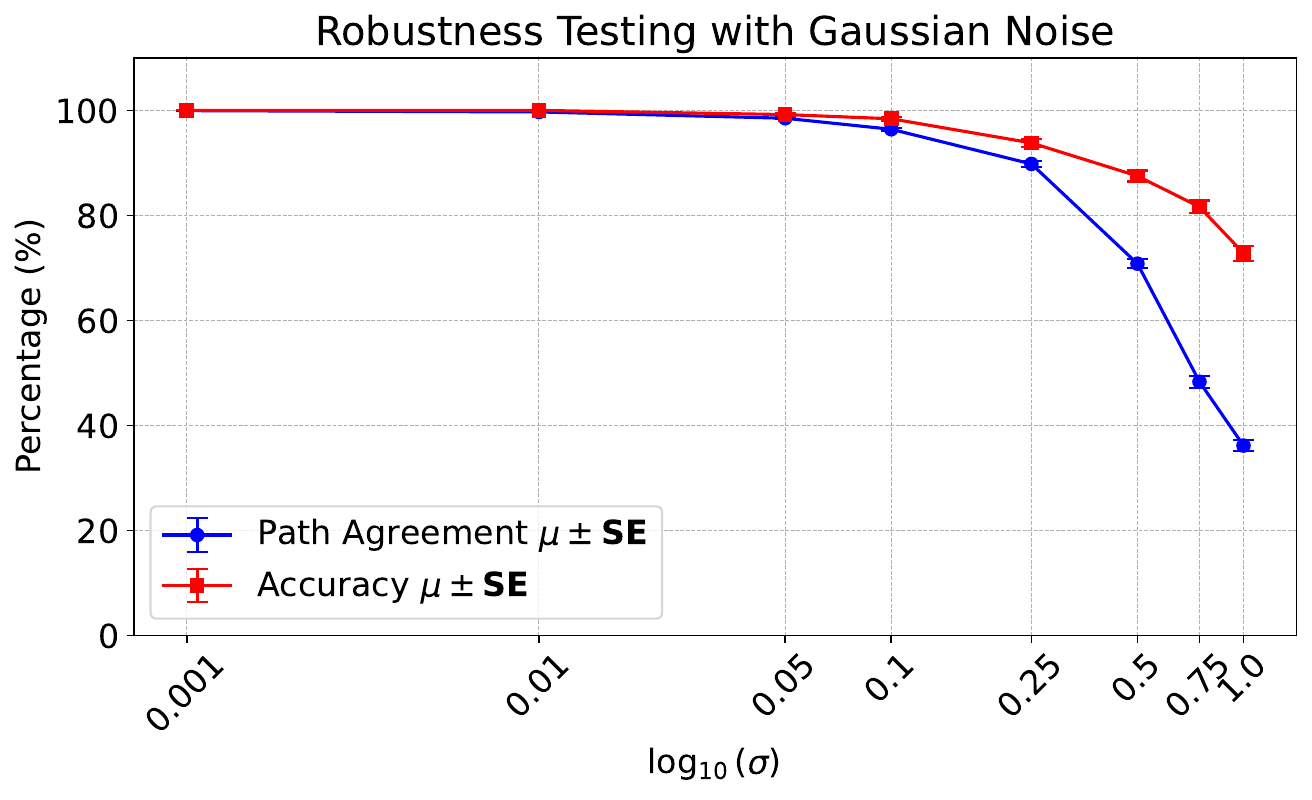}
    \caption{Cluster path agreement (blue) versus top-1 accuracy (red) under increasing Gaussian noise.
    Both metrics stay near 100\% for mild perturbations ($\sigma\le0.1$) and path agreement decreases to 90\% at $\sigma=0.25$ while accuracy is still 93\%, indicating that internal layer cluster assignments change before the model starts misclassifying.
    Beyond $\sigma=0.50$ both metrics drop quickly, demonstrating an abrupt transition from stable to unstable internal behavior.}
    \label{fig:robustness-gauss}
\end{figure*}

We test the robustness of the cluster path explanation by adding Gaussian noise $\mathcal{N}(0,\sigma^{2}$) with $\sigma\in[10^{-3},1.0]$.
For every perturbed image we recorded the path-agreement score (PA) relative to the clean reference and the model's top-1 accuracy.  
As shown in Fig. \ref{fig:robustness-gauss}, both metrics remain stable up to $\sigma\le$ 0.10 (PA $\geq$ 97\%, accuracy $\geq$ 99\%).  
At $\sigma=0.25$ PA drops to 90\% while accuracy is still 93\%, indicating that cluster assignments change before the model misclassifies.  
Between $\sigma=0.25$ and 0.50 roughly one-third of layer assignments differ but accuracy stays above 85\%, suggesting redundant internal representations. 
For $\sigma\ge$ 0.50, PA falls below 40\% and accuracy declines to 73\%, marking an abrupt transition from stable to unstable behavior.  
In the appendix (Sec.~\ref{sec:geometric_transformations}), we systematically applied combinations of mild rotations (up to ±10°), translations (up to ±5\% of image dimensions), and scalings (0.9-1.1×) to test how cluster paths respond to realistic spatial perturbations.
Despite the model's top-1 accuracy remaining high ($\geq$90\% under all tested transforms), path-agreement dropped (e.g., to 78\% at 10° rotation with maximal scaling). 
These findings reveal that the network's internal activation paths are far more sensitive to geometric changes than its final predictions.
Cluster paths therefore provide an earlier and more sensitive indicator of representational drift than output accuracy alone.
By monitoring representational drift, practitioners can proactively detect and address model brittleness, which might indicate a need for retraining or data augmentation, well before any drop in accuracy in production.

\xhdr{Geometric transformations}\label{sec:geometric_transformations}
To mimic more realistic perturbations, we applied an ablation of all combinations of mild geometric transformations such as rotations, translations, and scaling.
The angles of rotation for experiments were as 0°, 5°, and 10°; the translations were limited to a maximum of 5\% of the image's width and height in vertical and horizontal directions; finally, the images were scaled between 0.9 and 1.1 times the original size.
According to Table~\ref{tab:transformation-tests}, the network showed relatively high robustness in class output for mild transformations, yet the PA significantly decreases with larger transformations (e.g., at 10° rotation and maximum scaling, the PA dropped to 78\%, while the model maintained 92\% accuracy).
This indicated that although the final class predictions remain stable, the internal paths the model uses to reach these decision are highly sensitive to changes in spatial input features.

\begin{table*}[ht]
\centering
\setlength{\tabcolsep}{5pt}
\footnotesize
\caption{Comparison of percent Path Agreement (PA) and model accuracy (Acc.) for \celeba under various geometric transformations for rotations, translations, and scaling. Each PA and accuracy value is given as $\mu \pm$ SE for 1,000 samples.}

\label{tab:transformation-tests}
\begin{tabular}{llcccccccc}
\toprule
& & \multicolumn{2}{c}{0° Rotation} & \multicolumn{2}{c}{5° Rotation} & \multicolumn{2}{c}{10° Rotation} \\
\cmidrule(lr){3-4} \cmidrule(lr){5-6} \cmidrule(lr){7-8}
\textbf{Translate} & \textbf{Scale} & \textbf{PA $\uparrow$} & \textbf{Acc. $\uparrow$} & \textbf{PA $\uparrow$} & \textbf{Acc. $\uparrow$} & \textbf{PA $\uparrow$} & \textbf{Acc. $\uparrow$} \\
\midrule
0.0 \std0.0 & 1.0 \std0.0 & 100.0 \std0.0 & 100.0 \std0.0 & 95.2 \std0.4 & 97.5 \std0.5 & 91.6 \std0.5 & 96.3 \std0.6 \\
0.0 \std0.0 & 1.0 \std0.05 & 95.7 \std0.4 & 97.9 \std0.5 & 92.7 \std0.5 & 96.7 \std0.6 & 89.6 \std0.6 & 95.8 \std0.6 \\
0.0 \std0.0 & 1.0 \std0.1 & 90.8 \std0.6 & 95.5 \std0.7 & 89.7 \std0.6 & 95.4 \std0.7 & 87.2 \std0.6 & 93.7 \std0.8 \\
\midrule
0.0 \std0.02 & 1.0 \std0.0 & 91.5 \std0.5 & 95.8 \std0.6 & 90.8 \std0.5 & 95.4 \std0.7 & 88.2 \std0.6 & 95.0 \std0.7 \\
0.0 \std0.02 & 1.0 \std0.05 & 90.1 \std0.6 & 96.1 \std0.6 & 89.3 \std0.6 & 94.3 \std0.7 & 87.5 \std0.6 & 92.6 \std0.8 \\
0.0 \std0.02 & 1.0 \std0.1 & 87.6 \std0.6 & 95.0 \std0.7 & 87.0 \std0.6 & 93.6 \std0.8 & 85.4 \std0.7 & 93.5 \std0.8 \\
\midrule
0.0 \std0.05 & 1.0 \std0.0 & 80.6 \std0.8 & 91.1 \std0.9 & 80.0 \std0.8 & 90.6 \std0.9 & 80.0 \std0.8 & 91.6 \std0.9 \\
0.0 \std0.05 & 1.0 \std0.05 & 79.8 \std0.8 & 91.2 \std0.9 & 81.0 \std0.8 & 93.0 \std0.8 & 78.6 \std0.8 & 90.9 \std0.9 \\
0.0 \std0.05 & 1.0 \std0.1 & 78.0 \std0.8 & 92.0 \std0.9 & 78.3 \std0.9 & 90.1 \std0.9 & 77.4 \std0.9 & 90.4 \std0.9 \\
\bottomrule
\end{tabular}
\end{table*}

\subsubsection{RQ4. Cluster path visualization}

To assess how cluster paths elucidate the model's behavior, we sampled 10,000 training images and visualized up to 40 instances per path. 
This ``glocal'' perspective-local to a single prediction yet global to the cohort that shares its route-allows us to inspect the concepts the hair-color CNN has internalized.
We complement our analysis with the Sankey diagram as a tool for getting a birds-eye-view of the network's decision paths (see Section~\ref{sec:sankey}).

\begin{figure*}[ht]
  \centering
  % ------- top image -------
  \begin{subfigure}[t]{1.0\linewidth}
    \centering
    \includegraphics[trim=0pt 170pt 0pt 35pt,clip,width=0.9\linewidth]
        {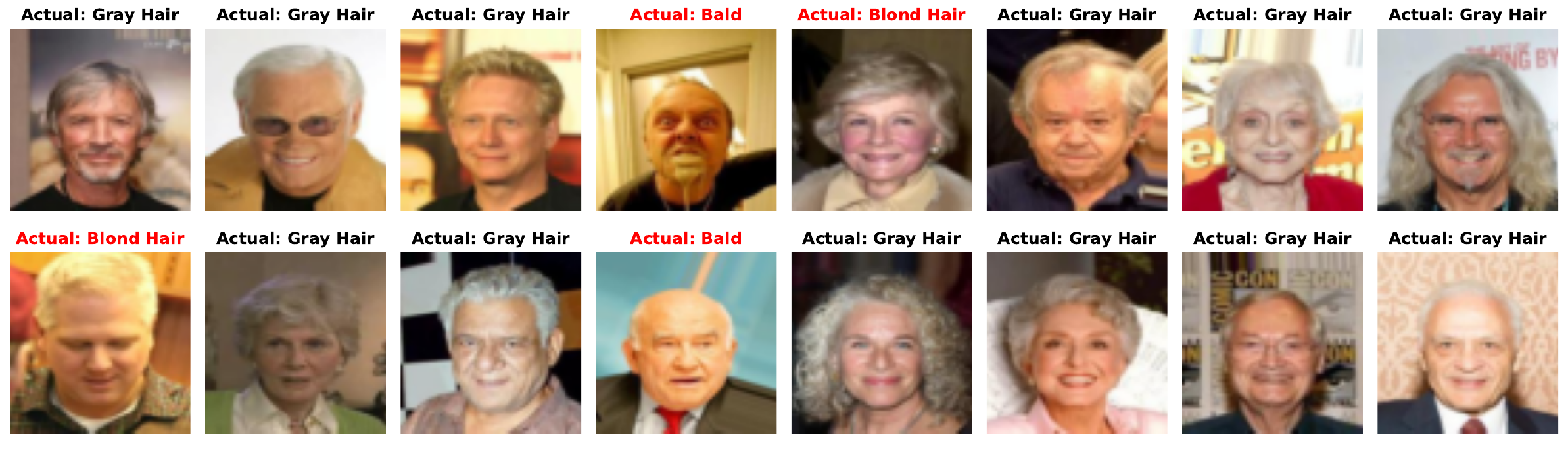}
    % \caption{}
    \label{fig:gray_hair_path_yellowpink}
  \end{subfigure}

  \vspace{1em} % optional vertical spacing

  % ------- bottom image -------
  \begin{subfigure}[ht]{1.0\linewidth}
    \centering
    \includegraphics[trim=0pt 170pt 0pt 35pt,clip,width=0.9\linewidth]
        {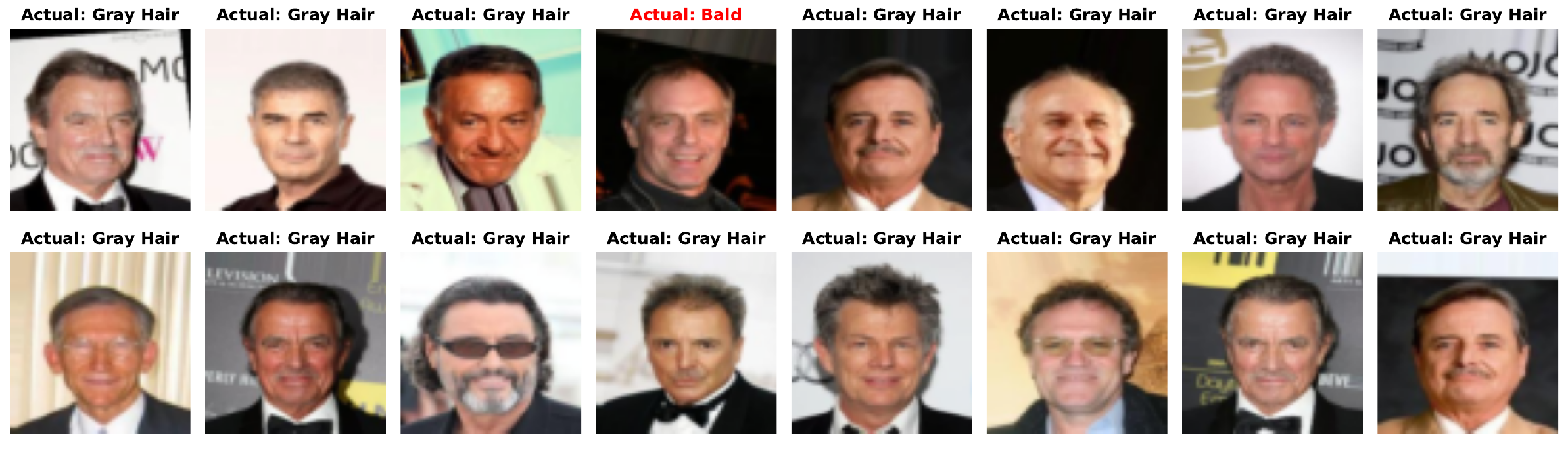}
    % \caption{(19 samples, ``Gray Hair''). Misclassified images outlined in red.}
    \label{fig:gray_hair_path_suits}
  \end{subfigure}
  \caption{Comparison of two ``Gray Hair'' cluster paths.  
  (Top two rows) cluster path 2$\to$1$\to$2$\to$4 and (bottom two rows) cluster path 1$\to$1$\to$2$\to$4. Each panel shows a 16-image subset; ground-truth labels appear above each image. Misclassified images are highlighted in red.
 }
  \label{fig:gray_hair_paths_combined}
\end{figure*}

\xhdr{Case study 1: gray hair}
Two gray paths that diverge in the first layer (2$\to$1$\to$2$\to$4 and 1$\to$1$\to$2$\to$4; Fig.~\ref{fig:gray_hair_paths_combined}) highlight different cues.  
The first mixes genders and exhibits colored tints (yellow and pink), whereas the second comprises men in formal suits with dark-gray hair.  
A lone misclassified sample in the latter path conforms to the same visual template, indicating that attire and hair shade, not label noise, very likely guided the network's decision.

\begin{figure*}[ht]
    \centering
    \includegraphics[%
        trim=0pt 170pt 0pt 35pt,%
        clip,%
        width=0.9\linewidth%
    ]{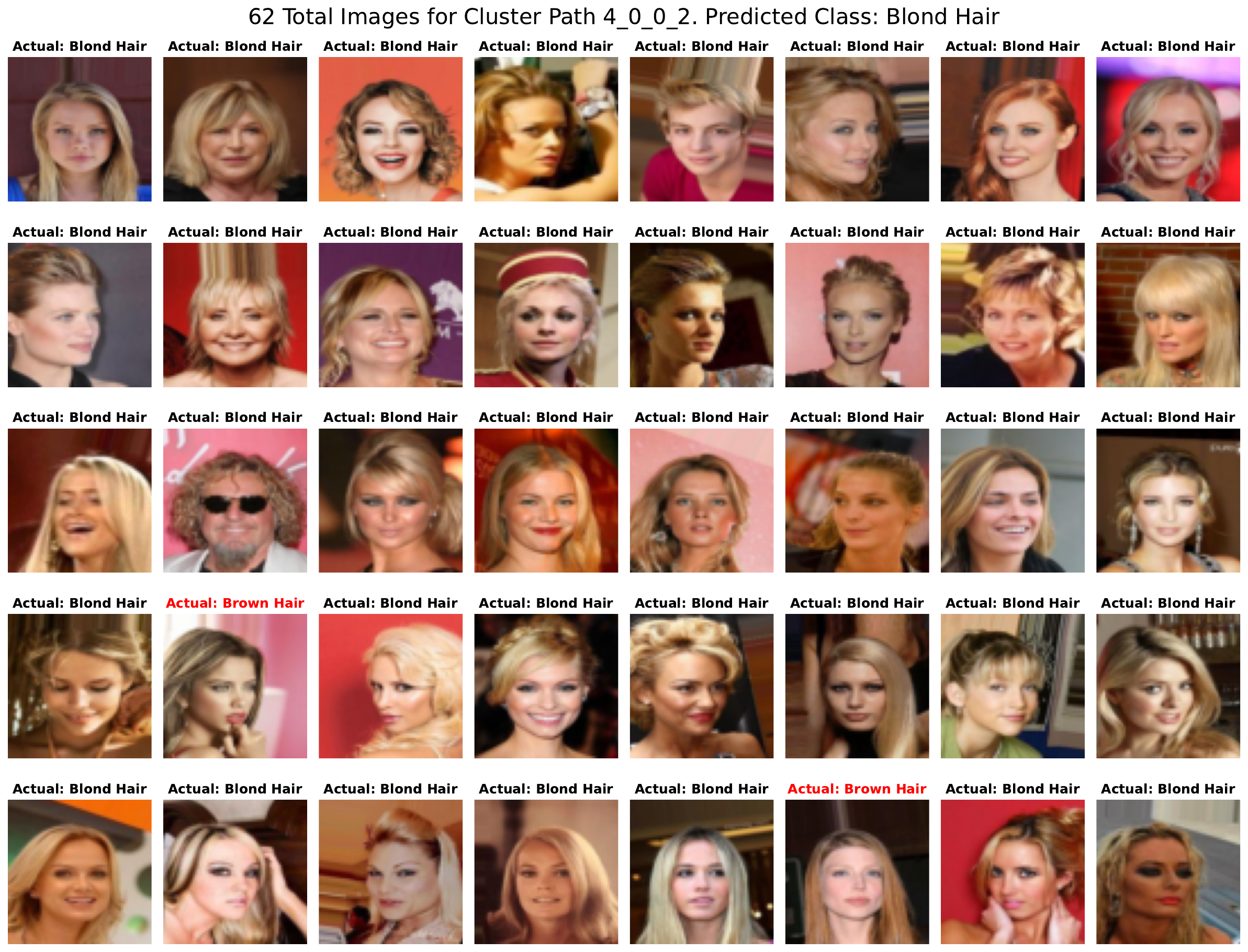}
    \caption{A subset of 32 images going through the cluster path 4$\to$0$\to$0$\to$2. The predicted class for all images in this path was ``Blond Hair.'' Each image has the ground truth label from the \celeba dataset above in the title. Incorrectly predicted images are highlighted in red.}
    \label{fig:blond_hair_path_warm_colors}
\end{figure*}

\begin{figure*}[ht]
    \centering
    \includegraphics[%
        trim=0pt 170pt 0pt 35pt,%
        clip,%
        width=0.9\linewidth%
    ]{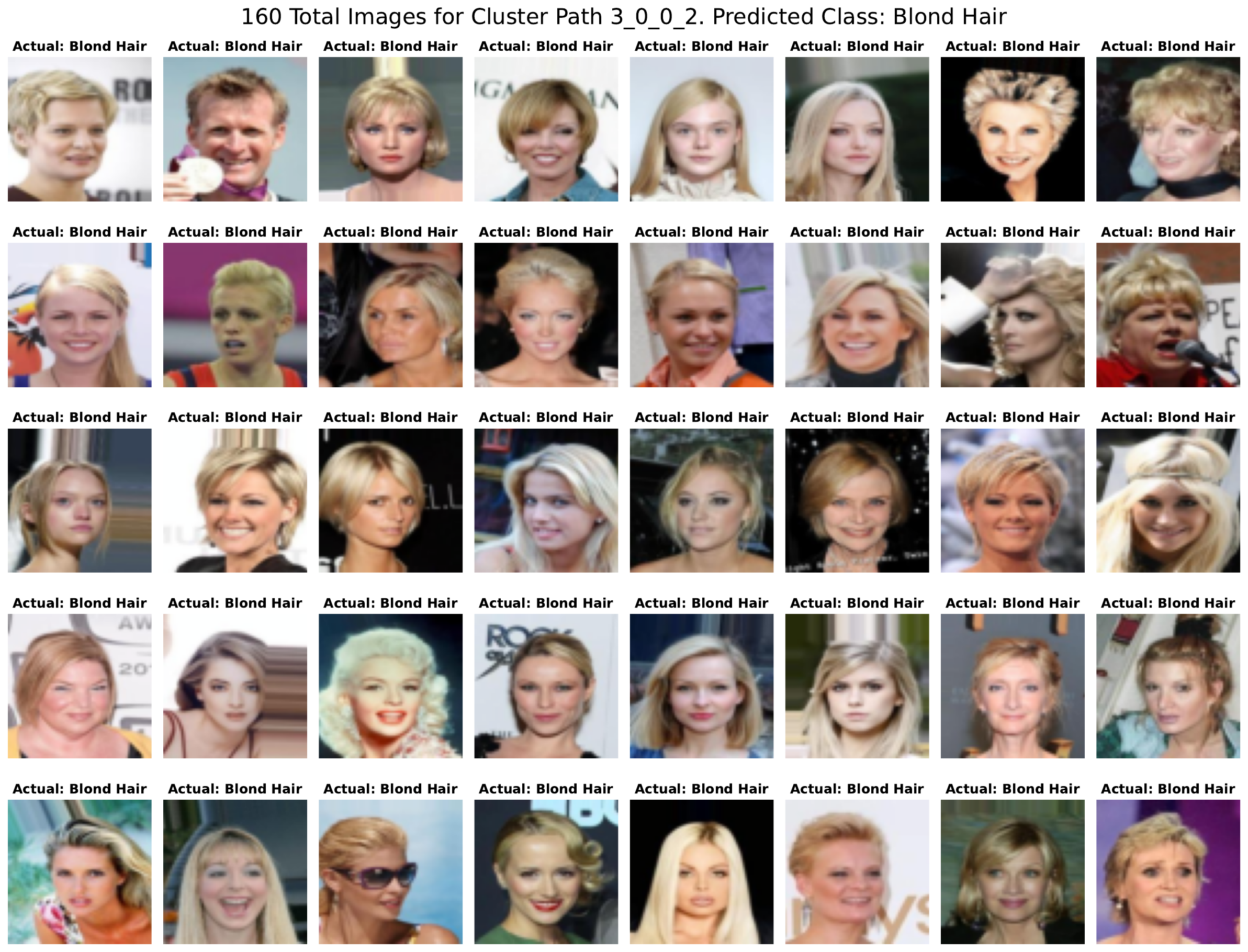}
    \caption{A subset of 32 images going through the cluster path 3$\to$0$\to$0$\to$2. The predicted class for all images in this path was ``Blond Hair.'' Each image has the ground truth label from the \celeba dataset above in the title. Incorrectly predicted images are highlighted in red.}
    \label{fig:blond_hair_path_gray}
\end{figure*}

\begin{figure*}[ht]
    \centering
    \includegraphics[%
        trim=0pt 0pt 0pt 30pt,%
        clip,%
        width=0.9\linewidth%
    ]{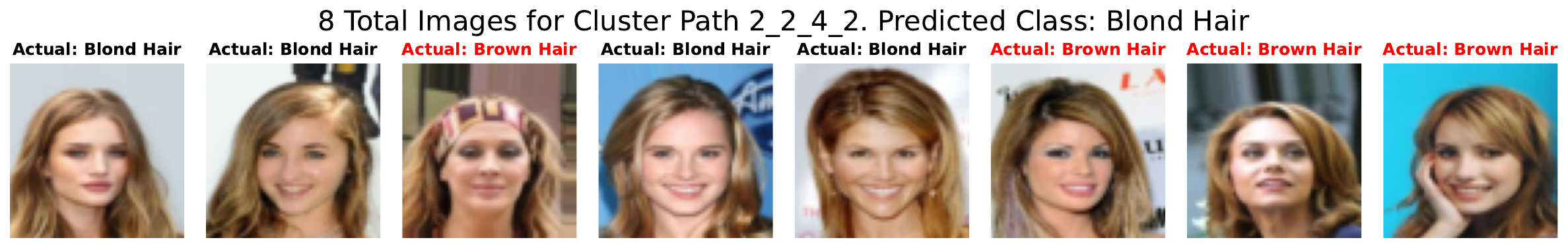}
    \caption{A subset of eight images going through the cluster path 2$\to$2$\to$4$\to$2. The predicted class for all images in this path was ``Blond Hair.'' Each image has the ground truth label from the \celeba dataset above in the title. Incorrectly predicted images are highlighted in red.}
    \label{fig:blond_hair_path_hairstyle}
\end{figure*}

\xhdr{Case study 2: blond hair}
Three blond paths, two of which diverge in the first layer, (4$\to$0$\to$0$\to$2, 3$\to$0$\to$0$\to$2, and 2$\to$2$\to$4$\to$2; Figs.~\ref{fig:blond_hair_path_warm_colors}--\ref{fig:blond_hair_path_hairstyle}) capture distinct visual sub-types.  
The first contains predominantly dirty-blond women against warm, earth-tone backgrounds and includes several brown-hair images mispredicted as blond, most likely because background hue and gender override the ground truth. 
The second path groups platinum-blond portraits with stark black-or-white backdrops and shows no errors in the displayed subset.  
The third path (eight images, four errors) consists of very light brown hair with identical hairstyles.
The pattern suggests either label noise or legitimate ambiguity rather than a model fault.

\begin{figure*}[ht]
    \centering
    \includegraphics[%
        trim=0pt 170pt 0pt 35pt,%
        clip,%
        width=0.9\linewidth%
    ]{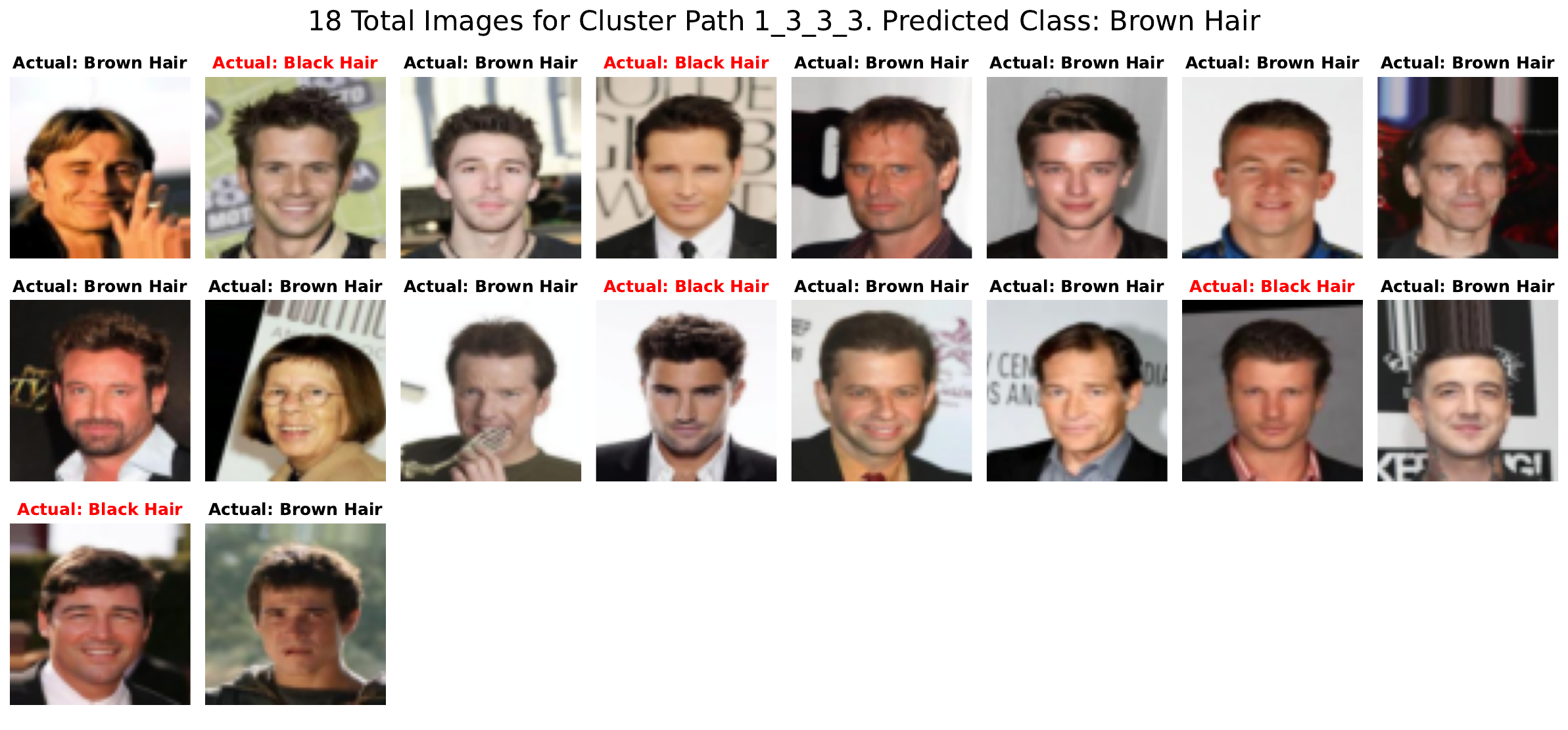}
    \caption{A subset of 16 images going through the cluster path 1$\to$3$\to$3$\to$3. The predicted class for all images in this path was ``Brown Hair.'' Each image has the ground truth label from the \celeba dataset above in the title. Incorrectly predicted images are highlighted in red.}
    \label{fig:brown_hair_path_suits}
\end{figure*}

\xhdr{Case study 3: brown hair}
The brown hair path (1$\to$3$\to$3$\to$3; Fig.~\ref{fig:brown_hair_path_suits}) clusters men in business attire with short, dark hair.  
Five black hair labels fall into this group which visually match the cluster's dominant traits, highlighting systematic annotation ambiguity.

\xhdr{Implications for model debugging}
Across all six paths, images within a path are coherent and clearly differentiated from neighboring paths, enabling practitioners to 1) pinpoint precise visual attributes driving each decision, 2) detect label inconsistencies, and 3) diagnose recurring error modes. 
These results demonstrate that cluster paths provide concise, image-based evidence of both the model's learned concepts and the root causes of misclassifications, likely making them a practical diagnostic tool in real-world workflows.

% -------------------------------------------------------

\xhdr{Qualitative comparison to pixel-attribution maps}
We show a qualitative side-by-side analysis with four gradient-based attribution maps provided in the Appendix (Sec.~\ref{sec:qualitative-comparison}, Fig.~\ref{fig:qualitative-attribution}).
In brief, pixel methods show where the network looks (either scattered facial pixels for misclassifications or hair for correct classifications), while cluster paths uncover what concepts the model learns.

\begin{figure}[ht]
    \centering
        \includegraphics[%
        trim=0pt 0pt 0pt 0pt,%
        clip,%
        width=1.0\linewidth%
    ]{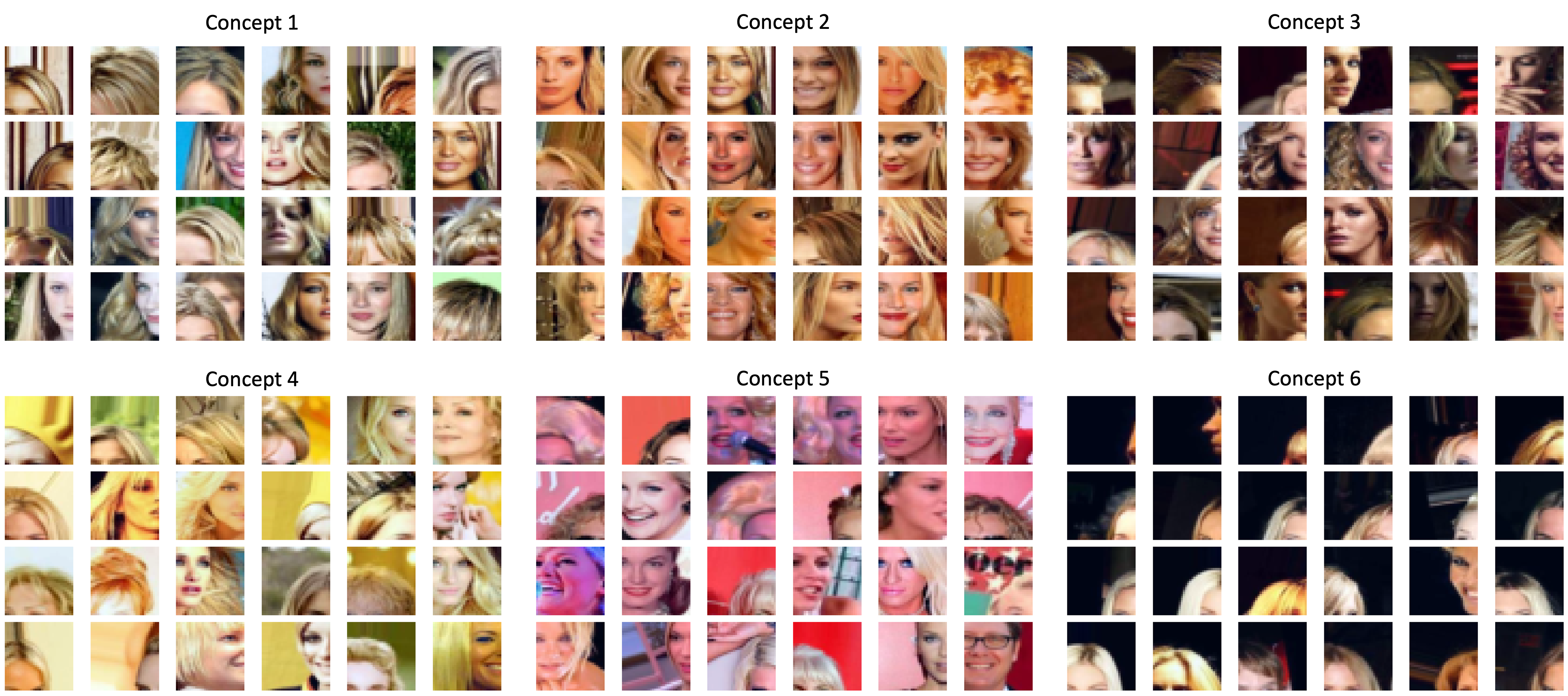}
    \caption{The top six concepts extracted by CRAFT for the `Blond' class. Concepts are numbered in order of importance. The high-level concepts identified are hair highlights, warm orange hues, dark backgrounds, yellow hues, pink hues, and finally, black backgrounds (in order of concept importance).}
    \label{fig:CRAFT_blond}
\end{figure}

\begin{figure}[ht]
    \centering
        \includegraphics[%
        trim=0pt 0pt 0pt 0pt,%
        clip,%
        width=1.0\linewidth%
    ]{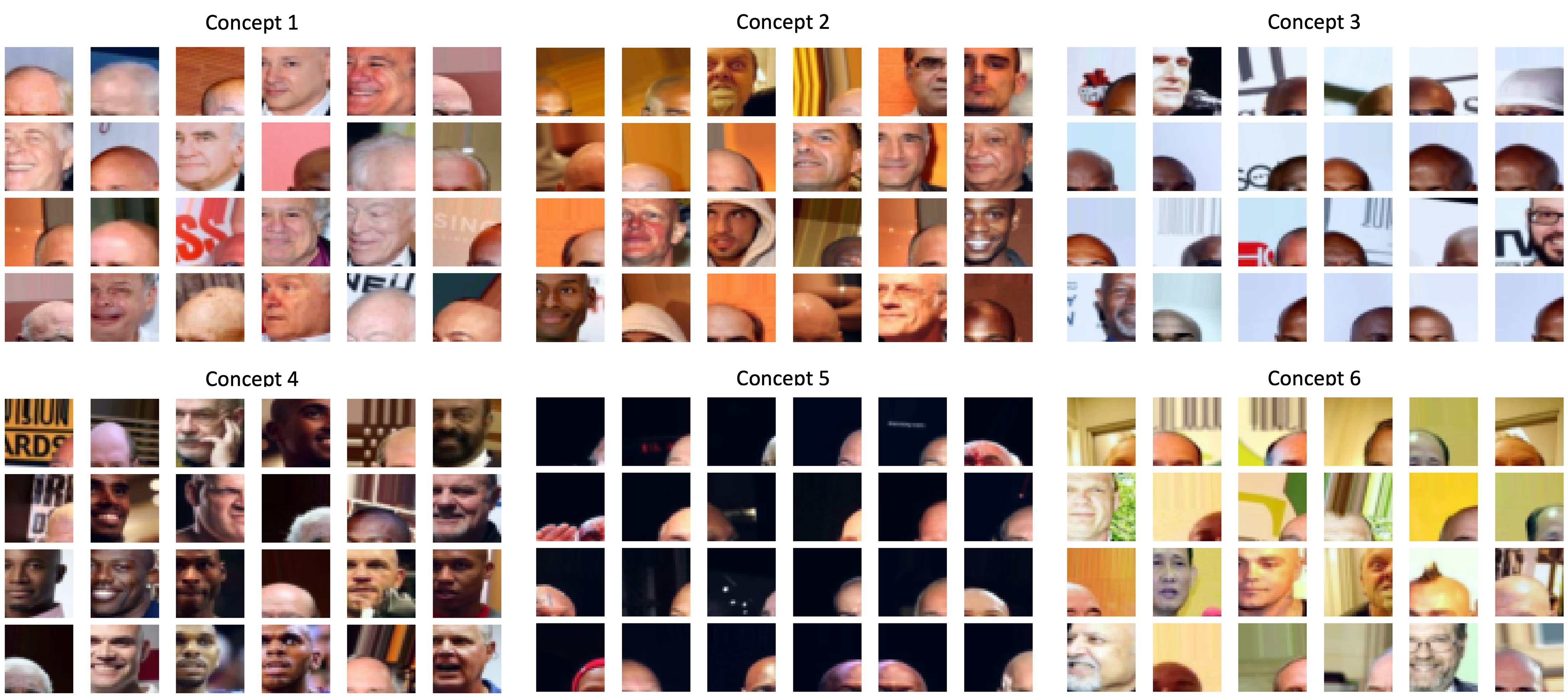}
    \caption{The top six concepts extracted by CRAFT for the `Bald' class. Concepts are numbered in order of importance. The high-level concepts identified are shape/contour-based temple cues, warm orange hues, white/light-blue backdrops, brown tones, black backgrounds, and yellow hues (in order of concept importance).}
    \label{fig:CRAFT_bald}
\end{figure}

\xhdr{Comparison to CRAFT}
We qualitatively compare the extracted concepts from a related state-of-the-art method, CRAFT \cite{fel_craft_2023}. 
CRAFT identifies concepts across layers, estimates concept importance, and generates concept attribution maps.
We configured CRAFT via a small, metric-driven sweep over the number of concepts $K$ and patch size $P$, using only internal diagnostics (rather than downstream scores) to ensure a fair comparison of methods.
For each setting we ran two seeds and measured: (i) stability by Hungarian-matching concepts across seeds and averaging the cosine similarity; (ii) parsimony from Sobol importances (top-3 mass and a Gini-like concentration score; $n_{\text{design}}=512$); and (iii) redundancy as the mean absolute off-diagonal correlation of spatially averaged concept activations. We min-max scaled each metric across the sweep to [0,1]. For redundancy (lower is better) we inverted the scale so that higher is better. 
This yielded a 4-D vector (concentration, stability, top-3 mass, 1-redundancy) with all coordinates being `higher = better.' We took the Pareto-optimal points and chose the one with the smallest Euclidean distance to (1,1,1,1), breaking ties in favor of smaller $K$ (and then smaller $P$).

As a case-study, we start by analyzing concepts from the `Blond' class. 
Figure~\ref{fig:CRAFT_blond} illustrates the resulting top six concepts extracted by CRAFT and Figures~\ref{fig:blond_hair_path_warm_colors}-\ref{fig:blond_hair_path_hairstyle} show three concepts extracted by cluster paths.
We note that the numbered concepts in our CRAFT figures are simply shown in order of sorted concept importance rather than a unique identifier for a concept across classes.
Both methods surface the same high-level factors such as color palettes and hair highlights, but CRAFT localizes where the cues live and ranks their global importance, whereas cluster paths reveal how such cues are composed into end-to-end decision routes.
Furthermore, the two dominant paths (4$\to$0$\to$0$\to$2 vs. 3$\to$0$\to$0$\to$2) differ only at the first clustered layer, indicating an early color palette concept that separates warm-scene blonds from platinum/contrast blonds before converging on an identical downstream pipeline.
The warm route yields Blond false positives on Brown hair, consistent with CRAFT's high importances for warm/yellow/pink concepts.
A third path (2$\to$2$\to$4$\to$2) groups look-alike hairstyles spanning Blond.Brown, suggesting ambiguity or label noise that CRAFT's patch-level maps could further pinpoint.

We provide another case-study showing concepts targeting the `Bald' class (Figure~\ref{fig:CRAFT_bald}).
CRAFT's top concepts reflect strong background palettes and styles (yellow walls and dark backgrounds).
In contrast, our cluster paths find one route grouping formal male portraits with suits while another mixes genders with warmer tints and softer lighting. The few failures inside a path share the same global concepts. Therefore both methods agree that lighting is a powerful cue the model leverages, but they both expose it at different granularities.
In summary, CRAFT and cluster paths are complementary.
Cluster paths group similar images together by route and identifies concepts via route-level divergences, while CRAFT localizes the responsible pixels and quantifies their contribution, together offering actionable targets for de-biases (e.g., color augmentation).

\subsection{RQ5. Cluster path scalability}\label{sec:scalability}
To investigate what concepts a large-scale classifier encodes in its internal routes, we generate cluster paths from a ViT-Base/patch-16/224 model \cite{dosovitskiy_image_2020} pretrained on the 1,000-class, $\approx$1.2M image dataset, ImageNet benchmark ~\cite{deng_imagenet_nodate} (Architecture details in Appendix~\ref{sec:transformer_architecture}). 
We limit our analysis to four equally spaced layers: Block0, Block4, Block8, and the final classification head, since clustering all 13 transformer blocks would yield
$\Omega=\prod_{\ell=1}^{13}K_\ell$, exceeding $20^{13}$, a prohibitively large path space.
Instead, we set $(K_{0},K_{4},K_{8},K_{\text{head}})=(20,20,20,100)$, which bounds the theoretical path complexity $\Omega=\prod_{\ell=1}^4 K_\ell=20^{3}\times100=800{,}000$.
Despite this upper bound, only 28,712 unique paths actually occur, 3.6\% of $\Omega$, indicating that the model's representations concentrate on a small fraction of theoretical routes. 
Moreover, the top 100 most frequent paths cover 17\% of the dataset, and the top 1,000 cover 47\%, confirming a heavy-tailed distribution of path usage that might guide targeted inspection towards the more popular paths first.

To translate these discrete paths into human-readable concepts, we employed a divergence-based comparison analysis: for each path we examined neighboring paths differing in exactly one layer ($d_{\text{Ham}}=1$) and used a large language model (LLM), \chatgpt~\cite{openai2025gpt4omini}, to aid annotation of clusters by suggesting concise concept labels (prompt: Fig.~\ref{fig:divergence_prompt}, hyperparameters: appendix~\ref{sec:transformer_clusterpaths}).  
Importantly, this step is not required for the cluster path method itself (the paths and their distributions exist independently). 
The LLM provides a scalable aid for labeling at scale, but practitioners could substitute human annotation or domain-specific heuristics as needed.
By treating each cluster as shorthand for a concept and comparing paths that diverge at a single layer, we can isolate the visual concepts each layer adds. 
The resulting ``concept paths'' reveal that even in a large-scale network, only a modest set of prototypical routes, and their corresponding high-level concepts, dominate the decision process, making such a large model more manageable for interpretability under this framework.

\begin{figure}[!ht]
    \centering
        \includegraphics[%
        trim=0pt 0pt 700pt 30pt,%
        clip,%
        width=1.0\linewidth%
    ]{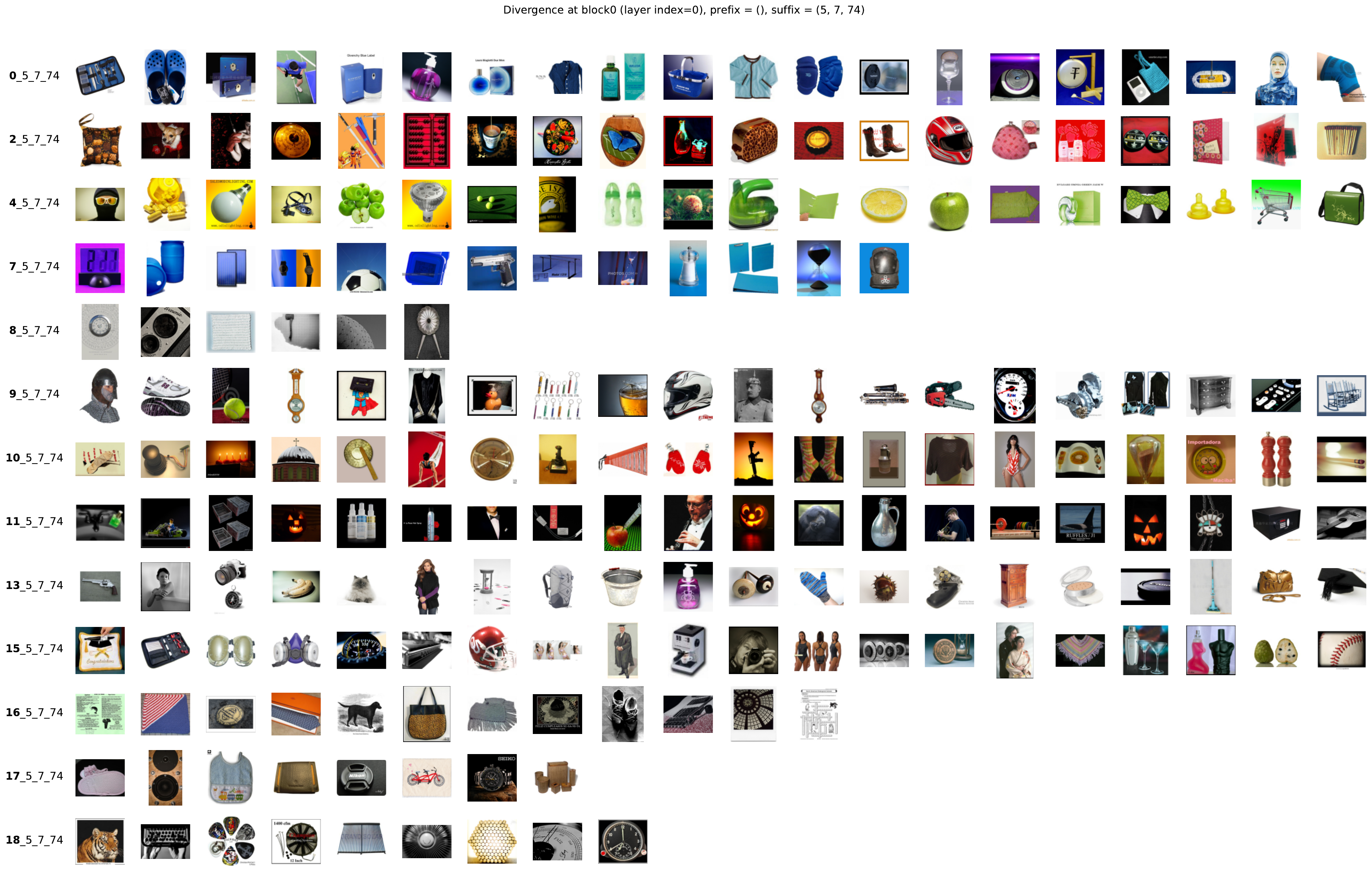}
    \caption{\textbf{Cluster path X$\to$5$\to$7$\to$74}. Applying our divergence analysis to `Block0' clusters with cluster path suffix `5$\to$7$\to$74.'  
    \chatgpt identified the following concepts per layer:
    Cluster 0: Cool blue hues,
Cluster 2: Warm red-orange tones,
Cluster 4: Yellow-green palette,
Cluster 7: Solid blue backgrounds,
Cluster 8: Grayscale textures,
Cluster 9: Shiny metallic objects,
Cluster 10: Brown wood tones,
Cluster 11: Objects on black backdrop,
Cluster 13: Objects on white backdrop,
Cluster 15: Multiple duplicates,
Cluster 16: Flat printed patterns,
Cluster 17: Personal accessories, and
Cluster 18: Geometric patterns.}
    \label{fig:transformer_block0_X_5_7_74}
\end{figure}

In Figs.~\ref{fig:transformer_block0_X_5_7_74} and \ref{fig:transformer_13_X_6_17}–\ref{fig:transformer_13_5_7_X}, the LLM consistently identifies intuitive attributes (e.g., “Cool blue hues'' vs. ``Warm red-orange tones'' in Block0, ``Clear-sky'' vs. ``Cloudy backgrounds'' in Block4) that align with human perception of the image grids. 
Appendix~\ref{sec:transformer_clusterpaths} shows additional results of our divergence analysis on each of the four layers we clustered and two examples of the resulting concept paths (Figs.~\ref{fig:transformer_18_13_13_45} and \ref{fig:transformer_18_14_12_68}).
The fully annotated concept paths in Figs.~\ref{fig:transformer_18_13_13_45} and \ref{fig:transformer_18_14_12_68} exhibit a high semantic consistency since each layer's top-3 keywords tightly summarize the progression from low-level appearance to higher-level scene context. 
While the LLM sometimes yields overlapping or broad labels (e.g., ``forms | shapes | light''), these annotations nevertheless capture the dominant visual factors distinguishing sub-paths. 
Together, this scalability experiment confirms that cluster paths, even for a large network, can be mapped to human-interpretable concepts with good fidelity, validating their utility for model introspection at scale.

\begin{figure}[h]
    \begin{ourbox}
        \footnotesize
        \textit{\\\textbf{Prompt Template for Divergence-Group Analysis}}\\
        \textbf{Background:} I'm going to show you a series of image-grids called ``divergence groups.'' 
        Each group contains images that follow the exact same cluster-path through the network except at one specific layer where they split into different cluster IDs. 
        The diverging cluster ID appears in the first left-most column and is bolded. 
        For example, **4**\_15\_17\_30 represents cluster ID 4 at the first layer, 15 at the second, and so on. You can ignore the plot's title.\\
        \textbf{Instructions:}\\
        1) Identify each diverging cluster ID as its own ``branch'' and compare *those images* to the other branches in the group.\\
        2) As an interpretability detective, name the single visual motif that most clearly *distinguishes* that branch's images from the others (e.g.\ color, shape, texture, composition, contrast).\\
        3) Give each branch a concise concept label (1–3 words); avoid category names.\\
        4) Format your answer as: \\
        \quad Cluster \(<\)ID\(>\): \(<\)Concept Label\(>\)\\\\
        \textbf{\# LLM Response:} \\
        Cluster 4: Solid blue backgrounds\\
        Cluster 15: Warm red tones\\
        Cluster 17: Geometric patterns
    \end{ourbox}
    \vspace{-0.1in}
    \caption{\small Our prompt template and example LLM response for our divergence analysis, used to identify concise concept labels from \chatgpt.}
    \label{fig:divergence_prompt}
\end{figure}

\subsection{RQ6. Out-of-distribution detection}\label{sec:OOD}
We further leverage the cluster path's ability to trace inputs through the model's feature space. 
The rarity of a path traversed by an input image is a strong signal for out-of-distribution (OOD) detection, allowing us to flag samples that significantly deviate from the training distribution before giving an over-confident misclassification.
OOD inputs are samples that significantly deviate from the training distribution, either from a new domain, a distributional shift, or previously unseen classes~\cite{salehi_unified_2022}.
Because modern neural networks are notoriously over-confident on such unknowns, reliable OOD detection is essential, especially in high-stakes domains such as healthcare, autonomous driving, and finance. 
A model producing overconfident yet incorrect predictions can mislead users, erode trust, or lead to costly consequences.
Robust detections systems allow practitioners to trigger fall-back mechanisms for human review or quantify when a deployed model should refrain from automated decisions when uncertainty is high.

Applying cluster paths for OOD consists of two distinct stages. 
During an offline modeling stage, we pool hidden activations from every layer of interest, summarize each activation vector with a six-dimensional statistic (maximum, mean, variance, skewness, minimum, and $\ell_{2}$-norm), and fit a $K$-component full-covariance Gaussian mixture model (GMM) to those summaries. 
The mixture both assigns every training sample a most-likely cluster ID and establishes, for each cluster, a log-likelihood floor defined by the $\rho$-th percentile of inlier scores. 
Repeating this independently across layers generates a look-up table that denotes how often each cluster path occurs in the training corpus.

At test time, a new sample is forwarded through the frozen network and the same statistical summaries are evaluated. 
If the summary at a given layer falls below that layer's likelihood floor, its cluster ID is replaced by a special $-1$ symbol, marking a local deviation while still allowing the remainder of the path to be constructed. 
The resulting path is then queried in the frequency table, previously compiled offline. 
A sample is declared OOD whenever the empirical probability of its path drops below a global rarity threshold~$\epsilon$.

Three hyper-parameters control the detector's behavior. 
The number of clusters per layer $K$, the percentile~$\rho$ calibrates how tolerant each layer is to natural inlier variation; and the path-rarity threshold~$\epsilon$ sets the overall aggressiveness of OOD flagging. 
In practice we tune~$\rho$ and~$\epsilon$ on held-out inlier data.
The algorithm for adapting cluster paths for OOD is detailed in Algorithm~\ref{algo:cluster-path-outlier-condensed}.

\begin{algorithm}[ht]
\caption{Algorithm for Outlier Detection using Cluster Paths}
\begin{algorithmic}[1]
\Procedure{OutlierDetectionClusterPaths}{}
  \State \textbf{Input:} Training set $X=\{x_i\in \mathbb{R}^{D}\}_{i=1}^{N}$, network $f_{\theta}$ with layer outputs $f_\theta^{(l)}(x)$, layer set $L$, clusters $K>1$, percentile $\rho\in(0,1)$, and threshold $\epsilon\in(0,1)$
  \State \textbf{Define:} For each $l\in L$, $\phi^{(l)}:\mathbb{R}^{d_l}\to\mathbb{R}^{6}$ computes \{max, variance, skewness, $l_2$ norm, mean, min\}
  \State \textbf{// Phase 1: Feature Augmentation and GMM Clustering}
  \For{each $l\in L$}
    \State $\Phi=\{\phi^{(l)}(f_\theta^{(l)}(x_i))\}_{i=1}^N$.
    \State Fit GMM on $\Phi$ to obtain $\{\pi_k^{(l)},\mu_k^{(l)},\Sigma_k^{(l)}\}_{k=1}^{K}$
    \For{$i=1,\dots,N$}
      \State $P(k|\Phi_i)=\dfrac{\pi_k^{(l)}\,\mathcal{N}(\Phi_i|\mu_k^{(l)},\Sigma_k^{(l)})}{\sum_{j=1}^{K}\pi_j^{(l)}\,\mathcal{N}(\Phi_i|\mu_j^{(l)},\Sigma_j^{(l)})}$
      
      \State $c^{(l)}_i=\underset{k\in\{1,\ldots,K\}}{\argmax}\,P(k|\Phi_i)$
    \EndFor
    \For{$k=1,\dots,K$}
    \State $p(\Phi_i) = \sum_{k=1}^K \pi_k^{(l)} \,\mathcal{N}\big(\Phi_i\mid\mu_k^{(l)},\Sigma_k^{(l)}\big)$
      \State $\tau_k^{(l)}\!=$ $\rho$-th percentile of $\{\log p(\Phi_i)| c^{(l)}_i=k\}$
    \EndFor
  \EndFor
  \State \textbf{// Phase 2: Flagging Outliers}
  \State \textbf{Input:} New sample $x_\nu\in\mathbb{R}^{D}$
  \For{each $l\in L$}
    \State $\varphi=\phi^{(l)}(f_\theta^{(l)}(x_\nu))$
    \State $c^{(l)}=\underset{k\in\{1,\ldots,K\}}{\argmax}\,P(k|\varphi)$
    \If{$\log p(\varphi)<\tau_{c^{(l)}}^{(l)}$} \State $c^{(l)}\leftarrow -1$
    \EndIf
  \EndFor
  \State \textbf{Define:} the path as a tuple of cluster IDs: 
        $\gamma(x_\nu)=(c^{(l)})_{l\in L}$ \Comment{Each $c^{(l)}$ is the cluster ID for layer $l$; tuple equality means all elements match}
   \State $p_{\text{path}}(\gamma(x_\nu))=\dfrac{|\{x_i \in X \mid \gamma(x_i)=\gamma(x_\nu)\}|}{N}$  \Comment{Each $x_i$ is a training sample from $X$ and $N = |X|$}
  
  \If{$p_{\text{path}}(\gamma(x_\nu))<\epsilon$}
  \State Flag $x_\nu$ as an overall outlier
  \EndIf
  \State \Return $\gamma(x_\nu)$ and outlier flag
\EndProcedure
\end{algorithmic}
\label{algo:cluster-path-outlier-condensed}
\end{algorithm}

\xhdr{Cluster Path OOD Detection Experiments}\label{OOD_experiments}
Finally, we evaluate the cluster path approach for identifying OOD samples which centers on the research question: 
``Can information from cluster paths be used for out-of-distribution detection?"
We compare the cluster path method to the max-softmax (baseline), ODIN, and DKNN.

\begin{table}[t]
\centering
\caption{List of the dataset configurations for OOD detection evaluation.}
\label{tab:datasets}
\begin{tabular}{lccc}
\toprule
\textbf{Inliers} & \textbf{CelebA} & \textbf{SVHN} & \textbf{CIFAR-10} \\
\midrule
\multirow{7}{*}{\textbf{Outliers}} 
                 & Random Images   & Random Images & Random Images     \\
                 & Uniform         & Uniform       & Uniform           \\
                 & Gaussian        & Gaussian      & Gaussian          \\
                 & CIFAR-10        & CIFAR-10      & SVHN            \\
                 & SVHN            & CelebA        & CelebA              \\
                 & STL-10          & STL-10        & STL-10            \\
                 & LSUN            & LSUN          & LSUN              \\
\bottomrule
\end{tabular}
\end{table}

\xhdr{Datasets and architectures} We evaluate our approach over three inlier datasets: \celeba (faces) \cite{liu_deep_2015}, \svhn (house numbers) \cite{netzer_reading_2011}, and \cifarten (animals and vehicles) \cite{krizhevsky_learning_2009}. 
The outlier datasets consist of \randomimages \footnote{Described here \url{https://github.com/hendrycks/outlier-exposure} and downloadable here \url{https://people.eecs.berkeley.edu/~hendrycks/300K_random_images.npy}} \cite{hendrycks_deep_2019}, \uniform, \gaussian, \stlten \cite{coates_analysis_2011}, and \lsun \cite{yu_lsun_2016}.
The list of dataset configurations for the OOD detection experiments are listed in Table~\ref{tab:datasets}.

We evaluate three CNN architectures which all share a common design in the last four fully connected layers.
Notably, aside from the variation in the output layer to match the number of classes, all models generally maintain a consistent design in the fully connected layers, ensuring comparable representational capacity across architectures, while maintaining high accuracy on each model's classification task.
Each model architecture can be found in Table~\ref{tab:svhn_arch_summary}, Table~\ref{tab:haircnn}, and Table~\ref{tab:resnet_arch_summary}.

\xhdr{Evaluation metrics} We evaluate our OOD detection performance using standard metrics such as AUROC, AUPR, and FPR at 95\% TPR \cite{hendrycks_baseline_2018} \cite{liang_enhancing_2020}.
\begin{itemize}
    \item FPR at 95\% TPR: This metric measures the probability that an out-of-distribution input is misclassified as in-distribution when the true positive rate (TPR = TP/(TP + FN)) is fixed at 95\%. The false positive rate (FPR = FP/(FP + TN)) is recorded at the first instance the TPR exceeds 95\%.

    \item AUROC: The area under the ROC curve quantifies the probability that an in-distribution example is assigned a higher detection score than an out-of-distribution example, with an ideal detector achieving 100\%.
    
    \item AUPR: Calculated as the area under the precision-recall curve (where precision = TP/(TP + FP) and recall = TP/(TP + FN)), this metric highlights the trade-off between precision and recall. In our experiments, we report a single AUPR value derived from the precision-recall curve.
\end{itemize}

\xhdr{Cluster path hyperparameters} 
We determine the optimal number of clusters for our cluster path method by tuning this hyperparameter exclusively on a random subset of 10,000 samples drawn from a curated set of 300k random images. 
These images are selected for their diversity and lack of distributional overlap with our inlier datasets, ensuring that the tuning process remains independent of the inlier distribution. 
This strategy is analogous to the Outlier Exposure~\cite{hendrycks_deep_2019} methodology, leveraging auxiliary outlier data to improve anomaly detection without contaminating the target test distribution. 
By using a separate, diverse collection for tuning, we ensure that our model does not “cheat” by implicitly optimizing against inlier characteristics, thereby enhancing the robustness and generalization of the cluster paths for out-of-distribution detection.

\xhdr{Statistical features of activations} The activation feature summary function $\phi^{(l)}$ function computes six statistical summaries—maximum, variance, skewness, $l_2$ norm, mean, and minimum—thus reducing the activation dimensionality while preserving essential distributional characteristics.
We note that these features were empirically found to work well and should be considered as an additional hyperparameter to be tuned, if needed.

\xhdr{Layers for cluster paths}
In constructing cluster paths, our choice of layers is guided by intuition rather than extensive empirical testing. 
We exclude the input layer, which does not transform the raw data. 
Additionally, when working with convolutional neural networks, we omit early convolutional layers, as they primarily act as feature extractors rather than encapsulating high-level semantic information relevant to classification. 
Although it is possible to include the convolutional layers, we believe that focusing on layers more directly tied to classification behavior yields a clearer depiction of how samples traverse the network's learned feature space. 
Consequently, we begin constructing paths at the first hidden layer with direct relevance to classification and independently cluster the activations of each subsequent layer, including the output layer.

\begin{table}[t]
\centering
\caption[Average OOD detection performance]{Average OOD detection performance metrics across outlier experiments for various inlier datasets.}
\label{tab:ood_performance_all}
\begin{tabular}{llccc}
\toprule
Inlier Dataset & Method & $\uparrow$ AUROC & $\uparrow$ AUPR & $\downarrow$ FPR@TPR95 \\
\midrule
\multirow{4}{*}{CelebA} 
  & Baseline      & 0.85  & 0.87  & 0.55  \\
  & DKNN          & 0.83  & 0.74  & 0.71  \\
  & ODIN          & 0.94  & 0.95  & 0.25  \\
  & Cluster Paths & 0.97 & 0.97 & 0.07 \\
  \\
\multirow{4}{*}{SVHN}
  & Baseline      & 0.96  & 0.97  & 0.21  \\
  & DKNN          & 0.95  & 0.98 & 0.38  \\
  & ODIN          & 0.89  & 0.91  & 0.55  \\
  & Cluster Paths & 0.97 & 0.95  & 0.06 \\

  \\
\multirow{4}{*}{CIFAR-10}
  & Baseline      & 0.72  & 0.80  & 0.72  \\
  & DKNN          & 0.79  & 0.85  & 0.84  \\
  & ODIN          & 0.73  & 0.81  & 0.65  \\
  & Cluster Paths & 0.97 & 0.96 & 0.07  \\
\bottomrule
\end{tabular}
\end{table}

\xhdr{Baselines} 
To validate our approach, we compare cluster paths against three established OOD detection algorithms that each address network robustness from complementary perspectives. 
First, the max-softmax baseline \cite{hendrycks_baseline_2018} relies on the maximum softmax probability to flag inputs that are misclassified or lie outside the training distribution; although simple, it is prone to overconfident predictions on anomalous data. 
Second, ODIN \cite{liang_enhancing_2020} enhances this baseline by applying temperature scaling and small input perturbations to more effectively separate in- and out-of-distribution softmax scores, thereby reducing false positive rates. 
Finally, the Deep k-Nearest Neighbors (DkNN) method \cite{papernot_deep_2018} probes intermediate representations by retrieving nearest neighbors at each layer, yielding calibrated confidence estimates and interpretable evidence from the training data. 
In contrast, our cluster path method leverages the entire hierarchy of activations by clustering them into discrete paths that capture the evolution of input features through the network. 
This layered aggregation not only provides a richer explanation of the model's decision process but also leads to more robust detection of distributional shifts. 
Comparing to these baselines enables us to highlight the benefits of our approach in terms of improved OOD detection performance.

\begin{figure}[ht]
  \centering
  \begin{subfigure}{0.49\textwidth}
    \centering
    \includegraphics[width=\textwidth]{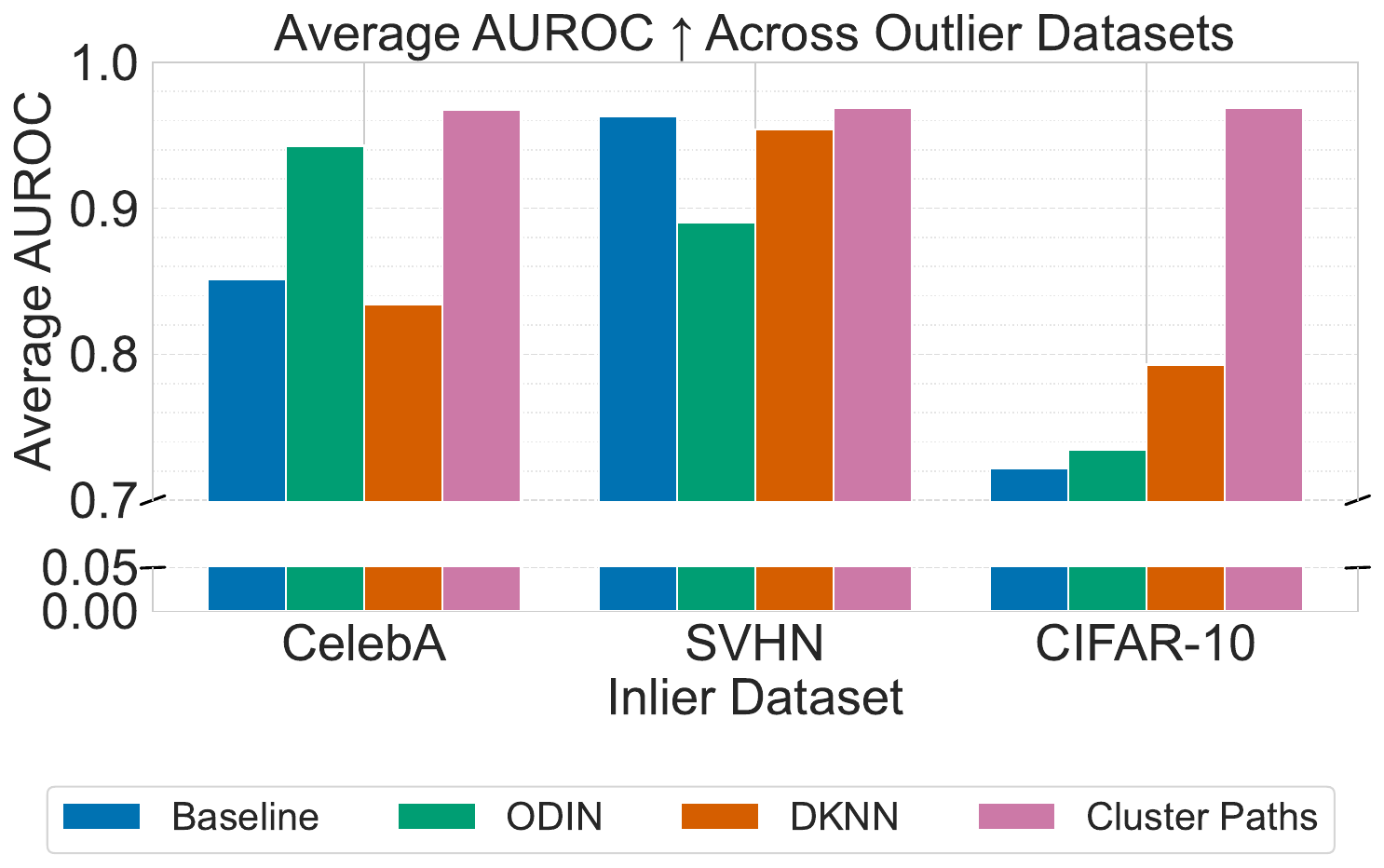}
  \end{subfigure}
  \hfill
  \begin{subfigure}{0.49\textwidth}
    \centering
    \includegraphics[width=\textwidth]{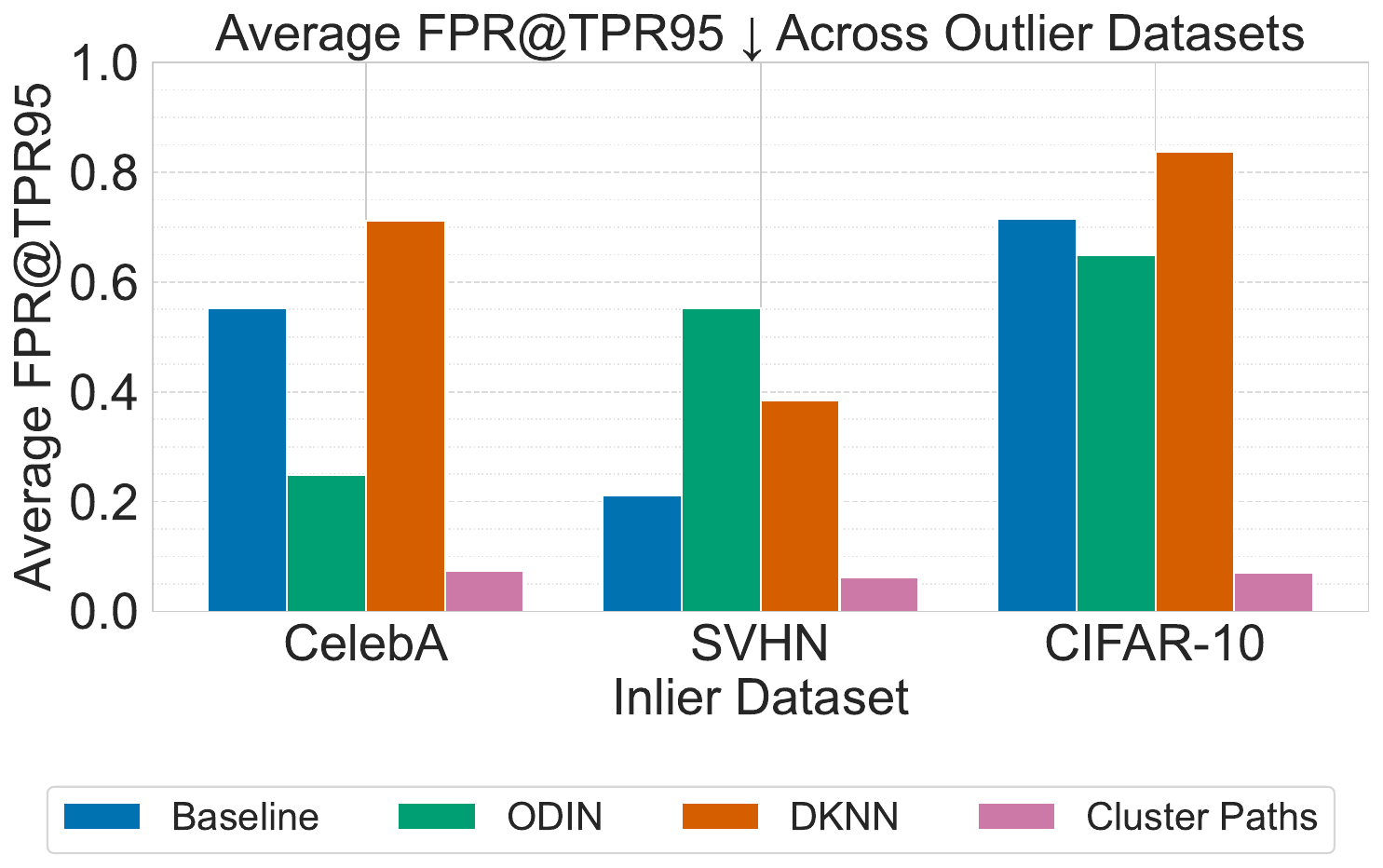}
  \end{subfigure}
  \caption[AUROC and FPR-95 of OOD detectors]{AUROC (left) and FPR @ 95\% TPR (right) for three inlier datasets and averaged across seven outlier datasets.}
  \label{fig:ood_barplots}
\end{figure}

% \subsection{Evaluation and Insights} 
% The cluster paths approach introduces a nuanced mechanism for OOD detection by mapping the evolution of activation features through the network into discrete cluster paths. 
% Each network layer contributes a unique transformation or view on the input data, from low-level features to high-level semantics. 
% In-distribution examples tend to follow common paths, while OOD samples, due to their atypical activations, result in rare or unique paths. 
% This section offers several key insights.

% \xhdr{Fine-grained partitioning of the feature space} Increasing the number of clusters partitions the activation space at a higher resolution, yielding a more granular representation of the learned features. 
% Each cluster encapsulates a narrowly defined region in the high-dimensional space, effectively reducing intra-cluster variance and capturing subtle variations in inlier patterns. 
% This refined segmentation creates tighter, well-defined regions where even minor deviations—often symptomatic of OOD behavior—become readily detectable. 
% In essence, the increased clustering enhances the model's ability to delineate precise boundaries between typical and atypical activation patterns.

\xhdr{Comparison with baselines} The empirical evidence shows that the cluster paths approach outperforms baseline methods (max-softmax, DKNN, ODIN) across multiple inlier datasets (CelebA, SVHN, CIFAR-10). 
The low false positive rates (as low as 0.06-0.07 FPR@TPR95) indicate that the method reliably distinguishes between densely populated inlier regions and sparse, anomalous areas in the feature space (see Figure~\ref{fig:ood_barplots} and Table~\ref{tab:ood_performance_all}).
The superior AUROC and AUPR values indicate that the method is not only sensitive to in-distribution characteristics but also robust in rejecting OOD samples. 
Each network layer contributes a unique transformation or view on the input data, from low-level features to high-level semantics. 
In-distribution examples tend to follow common paths, while OOD samples, due to their atypical activations, result in rare or unique paths. 
The consistent performance across diverse datasets underscores the generality and effectiveness of using multi-layer activation statistics combined with fine-grained clustering.

\section{Discussion}\label{discussion_insights}
Our experiments delineate both the power and limits of cluster paths.
On SpuriousCIFAR10, path purity mirrored the ground-truth shortcut: it was nearly perfect while patch color predicted the label and collapsed once the color was randomized, matching conclusions drawn from saliency maps and DkNN.  
On \celeba, forty high-volume paths covered 93\% of predictions; visual inspection of six representative paths exposes ambiguous hair-color labels and background or attire biases that saliency methods had missed.

Quantitatively, a Random-Forest fed only final-layer path IDs reproduced 71\% of the CNN's outputs; adding intermediate layers raised agreement to 81\%, and refining the final layer to 15 clusters brought it to 90\%. 
Thus the discretization inherent to cluster paths retains most, but not all, predictive signal. 
Robustness tests showed a complementary finding that path agreement began to degrade under Gaussian noise and mild geometric transforms while top-1 accuracy remained high, signaling an internal representational drift before misclassifications appeared.

The OOD experiments extend this finding: rare cluster paths serve as a reliable proxy for distributional shift. Across CelebA, SVHN, and CIFAR-10, path rarity achieved AUROC values near 0.97 and FPR@95\%TPR below 0.07, outperforming max-softmax, ODIN, and DkNN.
This suggests that the same discrete abstraction that makes cluster paths interpretable also doubles as a strong signal for flagging inputs that fall outside the training distribution.

\xhdr{Time and space complexity analysis} In comparison to another SOTA method for interpretability, cluster paths and DkNN have notably different computational and storage complexities (see Appendix~\ref{sec:computational_complexity} for full derivations). 
Cluster Paths require a one-time $O(NLKdI)$ clustering pass but then require fixed-size storage $O(LKd)$ and incur only $O(LKd)$ extra FLOPs per query. 
In contrast, DkNN builds its storage in $O(NLHd\log d)$ time, keeps all $O(NL(d+H))$ activations and hash entries, which is memory-intensive for large datasets. 
DkNN also adds $O(L(Hd\log d + sd + s\log K))$ time complexity at inference time. 
These trade-offs are important to consider when choosing an interpretability tool depending on deployment constraints.

\xhdr{Limitations}
Cluster paths can provide faithful and intuitive summaries of a model's internals, but as with any interpretability method, they necessarily abstract away some detail.
Reducing each layer to a finite set of centroids discards exact distance information, so inputs near cluster boundaries may share a path despite subtle differences.
This lossy mapping is observed in our faithfulness ceiling ($\approx90$\%). 
These effects are not unique to cluster paths, as proxy-based explainers such as LIME also simplify decision surfaces, yet they remind us that explanations should be corroborated with additional analyses (e.g. attribution-based methods).
Also, while k-means offers scalability and simplicity, alternative clustering methods might capture more complex manifolds at the cost of tractability; exploring these remains future work.

Computationally, clustering large models is generally no more demanding than many popular explanation techniques. 
For example, LIME fits a distinct surrogate model around each query point so its cost scales linearly with the number of explanations, whereas exact SHAP explanation computation requires exponential time in the number of features (with no fully-polynomial randomized approximation scheme in the general case).
Nevertheless, path enumeration can grow quickly with depth and granularity, so exploring more efficient clustering or path-pruning strategies is a promising avenue for future work.

Finally, while we have demonstrated cluster paths on CNNs and a Vision Transformer, extending them to recurrent or attention-only architectures will require thoughtful choices about which time steps, heads, or token positions to cluster. 
While LLM-based labeling was convenient for large-scale visualization, it should be viewed as an auxiliary scaling aid rather than a necessary component of the framework.
We leave the design of such adaptations to future research, although we are confident that the core idea of tracing discrete ``routes'' through a network will continue to offer insight across diverse layered-network architectures.
Despite these caveats, our results across spurious correlation tests, visualization case studies, and OOD detection show that cluster paths give practitioners an interpretable, concept-level tool for both auditing and improving models.

\xhdr{Broader impact} From a societal perspective, improved interpretability can enhance user trust, facilitate model-debugging, and help detect biases or vulnerabilities.
However, no post-hoc explanation is perfectly faithful. 
The cluster abstraction discards distance information, particularly with an insufficient number of clusters, which can mislead users about the true concepts and decision boundaries, thus posing risks for safety- or privacy-critical applications. 
Additionally, revealing internal model behavior could aid adversaries in crafting targeted attacks or extracting sensitive information. 
We caution against blind adoption and recommend pairing cluster paths with complementary interpretability methods and robustness checks.

In summary, cluster paths create a succinct, concept-oriented view of network internals as they trace information routes and highlight hidden representational fragility. 
We show that cluster paths scale from small-sized feed-forward networks to large-scale vision transformers, where a coarse but concept-rich map of the network is more helpful than neuron-level detail.
These qualities make them a natural counterpart to weight-based probes and pixel-attribution maps, broadening the interpretability toolkit with a method that highlights both a concepts driving a model's decision and its subtle vulnerabilities.

\ack 
This work was supported in part by the National Science Foundation under CNS-2055123 and CNS-1933208. Any opinions, findings, and conclusions or recommendations expressed in this material are those of the authors and do not necessarily reflect the views of the National Science Foundation.
\endack

\bibliographystyle{unsrtnat}
\bibliography{references, custom_references}

\begin{thebibliography}{55}
\providecommand{\natexlab}[1]{#1}
\providecommand{\url}[1]{\texttt{#1}}
\expandafter\ifx\csname urlstyle\endcsname\relax
  \providecommand{\doi}[1]{doi: #1}\else
  \providecommand{\doi}{doi: \begingroup \urlstyle{rm}\Url}\fi

\bibitem[Papernot and McDaniel(2018)]{papernot_deep_2018}
Nicolas Papernot and Patrick McDaniel.
\newblock Deep k-{Nearest} {Neighbors}: {Towards} {Confident}, {Interpretable} and {Robust} {Deep} {Learning}.
\newblock \emph{arXiv:1803.04765 [cs, stat]}, March 2018.
\newblock URL \url{http://arxiv.org/abs/1803.04765}.
\newblock arXiv: 1803.04765.

\bibitem[LeCun et~al.(2015)LeCun, Bengio, and Hinton]{lecun_deep_2015}
Yann LeCun, Yoshua Bengio, and Geoffrey Hinton.
\newblock Deep learning.
\newblock \emph{Nature}, 521\penalty0 (7553):\penalty0 436--444, May 2015.
\newblock ISSN 1476-4687.
\newblock \doi{10.1038/nature14539}.
\newblock URL \url{https://www.nature.com/articles/nature14539}.
\newblock Number: 7553 Publisher: Nature Publishing Group.

\bibitem[Young et~al.(2018)Young, Hazarika, Poria, and Cambria]{young_recent_2018}
T.~Young, D.~Hazarika, S.~Poria, and E.~Cambria.
\newblock Recent {Trends} in {Deep} {Learning} {Based} {Natural} {Language} {Processing} [{Review} {Article}].
\newblock \emph{IEEE Computational Intelligence Magazine}, 13\penalty0 (3):\penalty0 55--75, August 2018.
\newblock ISSN 1556-6048.
\newblock \doi{10.1109/MCI.2018.2840738}.
\newblock Conference Name: IEEE Computational Intelligence Magazine.

\bibitem[Voulodimos et~al.(2018)Voulodimos, Doulamis, Doulamis, and Protopapadakis]{voulodimos_deep_2018}
Athanasios Voulodimos, Nikolaos Doulamis, Anastasios Doulamis, and Eftychios Protopapadakis.
\newblock Deep {Learning} for {Computer} {Vision}: {A} {Brief} {Review}, February 2018.
\newblock URL \url{https://www.hindawi.com/journals/cin/2018/7068349/}.
\newblock ISSN: 1687-5265 Pages: e7068349 Publisher: Hindawi Volume: 2018.

\bibitem[Andrews et~al.(1995)Andrews, Diederich, and Tickle]{andrews_survey_1995}
Robert Andrews, Joachim Diederich, and Alan~B. Tickle.
\newblock Survey and critique of techniques for extracting rules from trained artificial neural networks.
\newblock \emph{Knowledge-Based Systems}, 8\penalty0 (6):\penalty0 373--389, December 1995.
\newblock ISSN 0950-7051.
\newblock \doi{10.1016/0950-7051(96)81920-4}.
\newblock URL \url{http://www.sciencedirect.com/science/article/pii/0950705196819204}.

\bibitem[Biran and Cotton(2017)]{biran_explanation_2017}
Or~Biran and Courtenay~V. Cotton.
\newblock Explanation and {Justification} in {Machine} {Learning} : {A} {Survey} {Or}.
\newblock 2017.

\bibitem[Lipton(2017)]{lipton_mythos_2017}
Zachary~C. Lipton.
\newblock The {Mythos} of {Model} {Interpretability}.
\newblock \emph{arXiv:1606.03490 [cs, stat]}, March 2017.
\newblock URL \url{http://arxiv.org/abs/1606.03490}.
\newblock arXiv: 1606.03490.

\bibitem[Doshi-Velez and Kim(2017)]{doshi-velez_towards_2017}
Finale Doshi-Velez and Been Kim.
\newblock Towards {A} {Rigorous} {Science} of {Interpretable} {Machine} {Learning}.
\newblock \emph{arXiv:1702.08608 [cs, stat]}, March 2017.
\newblock URL \url{http://arxiv.org/abs/1702.08608}.
\newblock arXiv: 1702.08608.

\bibitem[Abdul et~al.(2018)Abdul, Vermeulen, Wang, Lim, and Kankanhalli]{abdul_trends_2018}
Ashraf Abdul, Jo~Vermeulen, Danding Wang, Brian~Y. Lim, and Mohan Kankanhalli.
\newblock Trends and {Trajectories} for {Explainable}, {Accountable} and {Intelligible} {Systems}: {An} {HCI} {Research} {Agenda}.
\newblock In \emph{Proceedings of the 2018 {CHI} {Conference} on {Human} {Factors} in {Computing} {Systems} - {CHI} '18}, pages 1--18, Montreal QC, Canada, 2018. ACM Press.
\newblock ISBN 978-1-4503-5620-6.
\newblock \doi{10.1145/3173574.3174156}.
\newblock URL \url{http://dl.acm.org/citation.cfm?doid=3173574.3174156}.

\bibitem[Guidotti et~al.(2018)Guidotti, Monreale, Ruggieri, Turini, Giannotti, and Pedreschi]{guidotti_survey_2018}
Riccardo Guidotti, Anna Monreale, Salvatore Ruggieri, Franco Turini, Fosca Giannotti, and Dino Pedreschi.
\newblock A {Survey} of {Methods} for {Explaining} {Black} {Box} {Models}.
\newblock \emph{ACM Computing Surveys}, 51\penalty0 (5):\penalty0 1--42, August 2018.
\newblock ISSN 0360-0300, 1557-7341.
\newblock \doi{10.1145/3236009}.
\newblock URL \url{https://dl.acm.org/doi/10.1145/3236009}.

\bibitem[Adadi and Berrada(2018)]{adadi_peeking_2018}
Amina Adadi and Mohammed Berrada.
\newblock Peeking {Inside} the {Black}-{Box}: {A} {Survey} on {Explainable} {Artificial} {Intelligence} ({XAI}).
\newblock \emph{IEEE Access}, 6:\penalty0 52138--52160, 2018.
\newblock ISSN 2169-3536.
\newblock \doi{10.1109/ACCESS.2018.2870052}.
\newblock Conference Name: IEEE Access.

\bibitem[Gilpin et~al.(2019)Gilpin, Bau, Yuan, Bajwa, Specter, and Kagal]{gilpin_explaining_2019}
Leilani~H. Gilpin, David Bau, Ben~Z. Yuan, Ayesha Bajwa, Michael Specter, and Lalana Kagal.
\newblock Explaining {Explanations}: {An} {Overview} of {Interpretability} of {Machine} {Learning}.
\newblock \emph{arXiv:1806.00069 [cs, stat]}, February 2019.
\newblock URL \url{http://arxiv.org/abs/1806.00069}.
\newblock arXiv: 1806.00069.

\bibitem[Miller(2019)]{miller_explanation_2019}
Tim Miller.
\newblock Explanation in artificial intelligence: {Insights} from the social sciences.
\newblock \emph{Artificial Intelligence}, 267:\penalty0 1--38, February 2019.
\newblock ISSN 0004-3702.
\newblock \doi{10.1016/j.artint.2018.07.007}.
\newblock URL \url{http://www.sciencedirect.com/science/article/pii/S0004370218305988}.

\bibitem[Murdoch et~al.(2019)Murdoch, Singh, Kumbier, Abbasi-Asl, and Yu]{murdoch_definitions_2019}
W.~James Murdoch, Chandan Singh, Karl Kumbier, Reza Abbasi-Asl, and Bin Yu.
\newblock Definitions, methods, and applications in interpretable machine learning.
\newblock \emph{Proceedings of the National Academy of Sciences}, 116\penalty0 (44):\penalty0 22071--22080, October 2019.
\newblock ISSN 0027-8424, 1091-6490.
\newblock \doi{10.1073/pnas.1900654116}.
\newblock URL \url{http://www.pnas.org/lookup/doi/10.1073/pnas.1900654116}.

\bibitem[Moraffah et~al.(2020)Moraffah, Karami, Guo, Raglin, and Liu]{moraffah_causal_2020}
Raha Moraffah, Mansooreh Karami, Ruocheng Guo, Adrienne Raglin, and Huan Liu.
\newblock Causal {Interpretability} for {Machine} {Learning} - {Problems}, {Methods} and {Evaluation}.
\newblock \emph{ACM SIGKDD Explorations Newsletter}, 22\penalty0 (1):\penalty0 18--33, May 2020.
\newblock ISSN 1931-0145.
\newblock \doi{10.1145/3400051.3400058}.
\newblock URL \url{https://doi.org/10.1145/3400051.3400058}.

\bibitem[Samek et~al.(2021)Samek, Montavon, Lapuschkin, Anders, and Muller]{samek_explaining_2021}
Wojciech Samek, Gregoire Montavon, Sebastian Lapuschkin, Christopher~J. Anders, and Klaus-Robert Muller.
\newblock Explaining {Deep} {Neural} {Networks} and {Beyond}: {A} {Review} of {Methods} and {Applications}.
\newblock \emph{Proceedings of the IEEE}, 109\penalty0 (3):\penalty0 247--278, March 2021.
\newblock ISSN 0018-9219, 1558-2256.
\newblock \doi{10.1109/JPROC.2021.3060483}.
\newblock URL \url{https://ieeexplore.ieee.org/document/9369420/}.

\bibitem[Beisbart and Räz(2022)]{beisbart_philosophy_2022}
Claus Beisbart and Tim Räz.
\newblock Philosophy of science at sea: {Clarifying} the interpretability of machine learning.
\newblock \emph{Philosophy Compass}, 17\penalty0 (6):\penalty0 e12830, 2022.
\newblock ISSN 1747-9991.
\newblock \doi{10.1111/phc3.12830}.
\newblock URL \url{https://onlinelibrary.wiley.com/doi/abs/10.1111/phc3.12830}.
\newblock \_eprint: https://onlinelibrary.wiley.com/doi/pdf/10.1111/phc3.12830.

\bibitem[Guidotti(2022)]{guidotti_counterfactual_2022}
Riccardo Guidotti.
\newblock Counterfactual explanations and how to find them: literature review and benchmarking.
\newblock \emph{Data Mining and Knowledge Discovery}, April 2022.
\newblock ISSN 1573-756X.
\newblock \doi{10.1007/s10618-022-00831-6}.
\newblock URL \url{https://doi.org/10.1007/s10618-022-00831-6}.

\bibitem[Buolamwini and Gebru(2018)]{buolamwini_gender_2018}
Joy Buolamwini and Timnit Gebru.
\newblock Gender {Shades}: {Intersectional} {Accuracy} {Disparities} in {Commercial} {Gender} {Classification}.
\newblock In \emph{Conference on {Fairness}, {Accountability} and {Transparency}}, pages 77--91. PMLR, January 2018.
\newblock URL \url{http://proceedings.mlr.press/v81/buolamwini18a.html}.
\newblock ISSN: 2640-3498.

\bibitem[Bolukbasi et~al.(2016)Bolukbasi, Chang, Zou, Saligrama, and Kalai]{bolukbasi_man_2016}
Tolga Bolukbasi, Kai-Wei Chang, James Zou, Venkatesh Saligrama, and Adam Kalai.
\newblock Man is to {Computer} {Programmer} as {Woman} is to {Homemaker}? {Debiasing} {Word} {Embeddings}.
\newblock \emph{arXiv:1607.06520 [cs, stat]}, July 2016.
\newblock URL \url{http://arxiv.org/abs/1607.06520}.
\newblock arXiv: 1607.06520.

\bibitem[Park et~al.(2018)Park, Shin, and Fung]{park_reducing_2018}
Ji~Ho Park, Jamin Shin, and Pascale Fung.
\newblock Reducing {Gender} {Bias} in {Abusive} {Language} {Detection}.
\newblock In \emph{Proceedings of the 2018 {Conference} on {Empirical} {Methods} in {Natural} {Language} {Processing}}, pages 2799--2804, Brussels, Belgium, October 2018. Association for Computational Linguistics.
\newblock \doi{10.18653/v1/D18-1302}.
\newblock URL \url{https://www.aclweb.org/anthology/D18-1302}.

\bibitem[Tomasev et~al.(2021)Tomasev, McKee, Kay, and Mohamed]{tomasev_fairness_2021}
Nenad Tomasev, Kevin~R. McKee, Jackie Kay, and Shakir Mohamed.
\newblock Fairness for {Unobserved} {Characteristics}: {Insights} from {Technological} {Impacts} on {Queer} {Communities}.
\newblock \emph{Proceedings of the 2021 AAAI/ACM Conference on AI, Ethics, and Society}, pages 254--265, July 2021.
\newblock \doi{10.1145/3461702.3462540}.
\newblock URL \url{http://arxiv.org/abs/2102.04257}.
\newblock arXiv: 2102.04257.

\bibitem[Begley et~al.(2020)Begley, Schwedes, Frye, and Feige]{begley_explainability_2020}
Tom Begley, Tobias Schwedes, Christopher Frye, and Ilya Feige.
\newblock Explainability for fair machine learning.
\newblock \emph{arXiv:2010.07389 [cs, stat]}, October 2020.
\newblock URL \url{http://arxiv.org/abs/2010.07389}.
\newblock arXiv: 2010.07389.

\bibitem[Zhou et~al.(2015)Zhou, Khosla, Lapedriza, Oliva, and Torralba]{zhou_learning_2015}
Bolei Zhou, Aditya Khosla, Agata Lapedriza, Aude Oliva, and Antonio Torralba.
\newblock Learning {Deep} {Features} for {Discriminative} {Localization}.
\newblock \emph{arXiv:1512.04150 [cs]}, December 2015.
\newblock URL \url{http://arxiv.org/abs/1512.04150}.
\newblock arXiv: 1512.04150.

\bibitem[Selvaraju et~al.(2020)Selvaraju, Cogswell, Das, Vedantam, Parikh, and Batra]{selvaraju_grad-cam_2020}
Ramprasaath~R. Selvaraju, Michael Cogswell, Abhishek Das, Ramakrishna Vedantam, Devi Parikh, and Dhruv Batra.
\newblock Grad-{CAM}: {Visual} {Explanations} from {Deep} {Networks} via {Gradient}-based {Localization}.
\newblock \emph{International Journal of Computer Vision}, 128\penalty0 (2):\penalty0 336--359, February 2020.
\newblock ISSN 0920-5691, 1573-1405.
\newblock \doi{10.1007/s11263-019-01228-7}.
\newblock URL \url{http://arxiv.org/abs/1610.02391}.
\newblock arXiv: 1610.02391.

\bibitem[Vaswani et~al.(2017)Vaswani, Shazeer, Parmar, Uszkoreit, Jones, Gomez, Kaiser, and Polosukhin]{vaswani_attention_2017}
Ashish Vaswani, Noam Shazeer, Niki Parmar, Jakob Uszkoreit, Llion Jones, Aidan~N Gomez, Ł~ukasz Kaiser, and Illia Polosukhin.
\newblock Attention is {All} you {Need}.
\newblock In \emph{Advances in {Neural} {Information} {Processing} {Systems}}, volume~30. Curran Associates, Inc., 2017.
\newblock URL \url{https://proceedings.neurips.cc/paper_files/paper/2017/hash/3f5ee243547dee91fbd053c1c4a845aa-Abstract.html}.

\bibitem[Letarte et~al.(2018)Letarte, Paradis, Giguère, and Laviolette]{letarte_importance_2018}
Gaël Letarte, Frédérik Paradis, Philippe Giguère, and François Laviolette.
\newblock Importance of {Self}-{Attention} for {Sentiment} {Analysis}.
\newblock In \emph{Proceedings of the 2018 {EMNLP} {Workshop} {BlackboxNLP}: {Analyzing} and {Interpreting} {Neural} {Networks} for {NLP}}, pages 267--275, Brussels, Belgium, 2018. Association for Computational Linguistics.
\newblock \doi{10.18653/v1/W18-5429}.
\newblock URL \url{http://aclweb.org/anthology/W18-5429}.

\bibitem[Jain and Wallace(2019)]{jain_attention_2019}
Sarthak Jain and Byron~C. Wallace.
\newblock Attention is not {Explanation}.
\newblock In Jill Burstein, Christy Doran, and Thamar Solorio, editors, \emph{Proceedings of the 2019 {Conference} of the {North} {American} {Chapter} of the {Association} for {Computational} {Linguistics}: {Human} {Language} {Technologies}, {Volume} 1 ({Long} and {Short} {Papers})}, pages 3543--3556, Minneapolis, Minnesota, June 2019. Association for Computational Linguistics.
\newblock \doi{10.18653/v1/N19-1357}.
\newblock URL \url{https://aclanthology.org/N19-1357}.

\bibitem[Viviano et~al.(2020)Viviano, Simpson, Dutil, Bengio, and Cohen]{viviano_saliency_2020}
Joseph~D. Viviano, Becks Simpson, Francis Dutil, Yoshua Bengio, and Joseph~Paul Cohen.
\newblock Saliency is a {Possible} {Red} {Herring} {When} {Diagnosing} {Poor} {Generalization}.
\newblock October 2020.
\newblock URL \url{https://openreview.net/forum?id=c9-WeM-ceB}.

\bibitem[Carton et~al.(2020)Carton, Mei, and Resnick]{carton_feature-based_2020}
Samuel Carton, Qiaozhu Mei, and Paul Resnick.
\newblock Feature-{Based} {Explanations} {Don}'t {Help} {People} {Detect} {Misclassifications} of {Online} {Toxicity}.
\newblock \emph{Proceedings of the International AAAI Conference on Web and Social Media}, 14:\penalty0 95--106, May 2020.
\newblock ISSN 2334-0770.
\newblock \doi{10.1609/icwsm.v14i1.7282}.
\newblock URL \url{https://ojs.aaai.org/index.php/ICWSM/article/view/7282}.

\bibitem[Bach et~al.(2015)Bach, Binder, Montavon, Klauschen, Müller, and Samek]{bach_pixel-wise_2015}
Sebastian Bach, Alexander Binder, Grégoire Montavon, Frederick Klauschen, Klaus-Robert Müller, and Wojciech Samek.
\newblock On {Pixel}-{Wise} {Explanations} for {Non}-{Linear} {Classifier} {Decisions} by {Layer}-{Wise} {Relevance} {Propagation}.
\newblock \emph{PLoS ONE}, 10\penalty0 (7), July 2015.
\newblock ISSN 1932-6203.
\newblock \doi{10.1371/journal.pone.0130140}.
\newblock URL \url{https://www.ncbi.nlm.nih.gov/pmc/articles/PMC4498753/}.

\bibitem[Schneider and Vlachos(2021)]{schneider_explaining_2021}
Johannes Schneider and Michalis Vlachos.
\newblock Explaining {Neural} {Networks} by {Decoding} {Layer} {Activations}, February 2021.
\newblock URL \url{http://arxiv.org/abs/2005.13630}.
\newblock arXiv:2005.13630 [cs, stat].

\bibitem[Wang et~al.(2023)Wang, Variengien, Conmy, Shlegeris, and Steinhardt]{wang_interpretability_2023}
Kevin Wang, Alexandre Variengien, Arthur Conmy, Buck Shlegeris, and Jacob Steinhardt.
\newblock Interpretability in the {Wild}: a {Circuit} for {Indirect} {Object} {Identification} in {GPT}-2 small.
\newblock 2023.

\bibitem[Conmy et~al.(2023)Conmy, Mavor-Parker, Lynch, Heimersheim, and Garriga-Alonso]{conmy_towards_2023}
Arthur Conmy, Augustine Mavor-Parker, Aengus Lynch, Stefan Heimersheim, and Adrià Garriga-Alonso.
\newblock Towards {Automated} {Circuit} {Discovery} for {Mechanistic} {Interpretability}.
\newblock \emph{Advances in Neural Information Processing Systems}, 36:\penalty0 16318--16352, December 2023.
\newblock URL \url{https://proceedings.neurips.cc/paper_files/paper/2023/hash/34e1dbe95d34d7ebaf99b9bcaeb5b2be-Abstract-Conference.html}.

\bibitem[Lieberum et~al.(2023)Lieberum, Rahtz, Kramár, Nanda, Irving, Shah, and Mikulik]{lieberum_does_2023}
Tom Lieberum, Matthew Rahtz, János Kramár, Neel Nanda, Geoffrey Irving, Rohin Shah, and Vladimir Mikulik.
\newblock Does {Circuit} {Analysis} {Interpretability} {Scale}? {Evidence} from {Multiple} {Choice} {Capabilities} in {Chinchilla}, July 2023.
\newblock URL \url{http://arxiv.org/abs/2307.09458}.
\newblock arXiv:2307.09458 [cs].

\bibitem[Fel et~al.(2023)Fel, Picard, Bethune, Boissin, Vigouroux, Colin, Cadène, and Serre]{fel_craft_2023}
Thomas Fel, Agustin Picard, Louis Bethune, Thibaut Boissin, David Vigouroux, Julien Colin, Rémi Cadène, and Thomas Serre.
\newblock {CRAFT}: {Concept} {Recursive} {Activation} {FacTorization} for {Explainability}, March 2023.
\newblock URL \url{http://arxiv.org/abs/2211.10154}.
\newblock arXiv:2211.10154 [cs].

\bibitem[Achtibat et~al.(2023)Achtibat, Dreyer, Eisenbraun, Bosse, Wiegand, Samek, and Lapuschkin]{achtibat_attribution_2023}
Reduan Achtibat, Maximilian Dreyer, Ilona Eisenbraun, Sebastian Bosse, Thomas Wiegand, Wojciech Samek, and Sebastian Lapuschkin.
\newblock From attribution maps to human-understandable explanations through {Concept} {Relevance} {Propagation}.
\newblock \emph{Nature Machine Intelligence}, 5\penalty0 (9):\penalty0 1006--1019, September 2023.
\newblock ISSN 2522-5839.
\newblock \doi{10.1038/s42256-023-00711-8}.
\newblock URL \url{https://www.nature.com/articles/s42256-023-00711-8}.
\newblock Publisher: Nature Publishing Group.

\bibitem[İrsoy and Alpaydın(2022)]{irsoy_pathfinder_2022}
Ozan İrsoy and Ethem Alpaydın.
\newblock {PathFinder}: {Discovering} {Decision} {Pathways} in {Deep} {Neural} {Networks}, October 2022.
\newblock URL \url{http://arxiv.org/abs/2210.00319}.
\newblock arXiv:2210.00319 [cs].

\bibitem[Lage et~al.(2019)Lage, Chen, He, Narayanan, Kim, Gershman, and Doshi-Velez]{lage_human_2019}
Isaac Lage, Emily Chen, Jeffrey He, Menaka Narayanan, Been Kim, Samuel~J. Gershman, and Finale Doshi-Velez.
\newblock Human {Evaluation} of {Models} {Built} for {Interpretability}.
\newblock \emph{Proceedings of the AAAI Conference on Human Computation and Crowdsourcing}, 7:\penalty0 59--67, October 2019.
\newblock ISSN 2769-1349.
\newblock \doi{10.1609/hcomp.v7i1.5280}.
\newblock URL \url{https://ojs.aaai.org/index.php/HCOMP/article/view/5280}.

\bibitem[Simonyan et~al.(2014)Simonyan, Vedaldi, and Zisserman]{simonyan_deep_2014}
Karen Simonyan, Andrea Vedaldi, and Andrew Zisserman.
\newblock Deep {Inside} {Convolutional} {Networks}: {Visualising} {Image} {Classification} {Models} and {Saliency} {Maps}.
\newblock \emph{arXiv:1312.6034 [cs]}, April 2014.
\newblock URL \url{http://arxiv.org/abs/1312.6034}.
\newblock arXiv: 1312.6034.

\bibitem[Sundararajan et~al.(2017)Sundararajan, Taly, and Yan]{sundararajan_axiomatic_2017}
Mukund Sundararajan, Ankur Taly, and Qiqi Yan.
\newblock Axiomatic {Attribution} for {Deep} {Networks}.
\newblock In \emph{International {Conference} on {Machine} {Learning}}, pages 3319--3328. PMLR, July 2017.
\newblock URL \url{http://proceedings.mlr.press/v70/sundararajan17a.html}.
\newblock ISSN: 2640-3498.

\bibitem[Lundberg and Lee(2017)]{lundberg_unified_2017}
Scott Lundberg and Su-In Lee.
\newblock A {Unified} {Approach} to {Interpreting} {Model} {Predictions}.
\newblock \emph{arXiv:1705.07874 [cs, stat]}, November 2017.
\newblock URL \url{http://arxiv.org/abs/1705.07874}.
\newblock arXiv: 1705.07874.

\bibitem[Liu et~al.(2015)Liu, Luo, Wang, and Tang]{liu_deep_2015}
Ziwei Liu, Ping Luo, Xiaogang Wang, and Xiaoou Tang.
\newblock Deep {Learning} {Face} {Attributes} in the {Wild}.
\newblock In \emph{Proceedings of {International} {Conference} on {Computer} {Vision} ({ICCV})}, December 2015.
\newblock URL \url{https://mmlab.ie.cuhk.edu.hk/projects/CelebA.html}.

\bibitem[Dosovitskiy et~al.(2020)Dosovitskiy, Beyer, Kolesnikov, Weissenborn, Zhai, Unterthiner, Dehghani, Minderer, Heigold, Gelly, Uszkoreit, and Houlsby]{dosovitskiy_image_2020}
Alexey Dosovitskiy, Lucas Beyer, Alexander Kolesnikov, Dirk Weissenborn, Xiaohua Zhai, Thomas Unterthiner, Mostafa Dehghani, Matthias Minderer, Georg Heigold, Sylvain Gelly, Jakob Uszkoreit, and Neil Houlsby.
\newblock An {Image} is {Worth} 16x16 {Words}: {Transformers} for {Image} {Recognition} at {Scale}.
\newblock October 2020.
\newblock URL \url{https://openreview.net/forum?id=YicbFdNTTy}.

\bibitem[Deng et~al.()Deng, Dong, Socher, Li, Li, and Fei-Fei]{deng_imagenet_nodate}
Jia Deng, Wei Dong, Richard Socher, Li-Jia Li, Kai Li, and Li~Fei-Fei.
\newblock {ImageNet}: {A} {Large}-{Scale} {Hierarchical} {Image} {Database}.

\bibitem[OpenAI(2025)]{openai2025gpt4omini}
OpenAI.
\newblock {GPT-4o-mini}: A large language model.
\newblock \url{https://openai.com/product/gpt-4o-mini}, 2025.

\bibitem[Salehi et~al.(2022)Salehi, Mirzaei, Hendrycks, Li, Rohban, and Sabokrou]{salehi_unified_2022}
Mohammadreza Salehi, Hossein Mirzaei, Dan Hendrycks, Yixuan Li, Mohammad~Hossein Rohban, and Mohammad Sabokrou.
\newblock A {Unified} {Survey} on {Anomaly}, {Novelty}, {Open}-{Set}, and {Out}-of-{Distribution} {Detection}: {Solutions} and {Future} {Challenges}, December 2022.
\newblock URL \url{http://arxiv.org/abs/2110.14051}.
\newblock arXiv:2110.14051 [cs].

\bibitem[Netzer et~al.(2011)Netzer, Wang, Coates, Bissacco, Wu, and Ng]{netzer_reading_2011}
Yuval Netzer, Tao Wang, Adam Coates, Alessandro Bissacco, Bo~Wu, and Andrew~Y Ng.
\newblock Reading {Digits} in {Natural} {Images} with {Unsupervised} {Feature} {Learning}.
\newblock \emph{NIPS Workshop on Deep Learning and Unsupervised Feature Learning}, 2011.

\bibitem[Krizhevsky(2009)]{krizhevsky_learning_2009}
Alex Krizhevsky.
\newblock Learning {Multiple} {Layers} of {Features} from {Tiny} {Images}.
\newblock April 2009.

\bibitem[Hendrycks et~al.(2019)Hendrycks, Mazeika, and Dietterich]{hendrycks_deep_2019}
Dan Hendrycks, Mantas Mazeika, and Thomas Dietterich.
\newblock Deep {Anomaly} {Detection} with {Outlier} {Exposure}, January 2019.
\newblock URL \url{http://arxiv.org/abs/1812.04606}.
\newblock arXiv:1812.04606.

\bibitem[Coates et~al.(2011)Coates, Lee, and Ng]{coates_analysis_2011}
Adam Coates, Honglak Lee, and Andrew~Y Ng.
\newblock An {Analysis} of {Single}-{Layer} {Networks} in {Unsupervised} {Feature} {Learning}.
\newblock \emph{AISTATS}, 2011.

\bibitem[Yu et~al.(2016)Yu, Seff, Zhang, Song, Funkhouser, and Xiao]{yu_lsun_2016}
Fisher Yu, Ari Seff, Yinda Zhang, Shuran Song, Thomas Funkhouser, and Jianxiong Xiao.
\newblock {LSUN}: {Construction} of a {Large}-scale {Image} {Dataset} using {Deep} {Learning} with {Humans} in the {Loop}, June 2016.
\newblock URL \url{http://arxiv.org/abs/1506.03365}.
\newblock arXiv:1506.03365 [cs].

\bibitem[Hendrycks and Gimpel(2018)]{hendrycks_baseline_2018}
Dan Hendrycks and Kevin Gimpel.
\newblock A {Baseline} for {Detecting} {Misclassified} and {Out}-of-{Distribution} {Examples} in {Neural} {Networks}, October 2018.
\newblock URL \url{http://arxiv.org/abs/1610.02136}.
\newblock arXiv:1610.02136 [cs].

\bibitem[Liang et~al.(2020)Liang, Li, and Srikant]{liang_enhancing_2020}
Shiyu Liang, Yixuan Li, and R.~Srikant.
\newblock Enhancing {The} {Reliability} of {Out}-of-distribution {Image} {Detection} in {Neural} {Networks}, August 2020.
\newblock URL \url{http://arxiv.org/abs/1706.02690}.
\newblock arXiv:1706.02690 [cs, stat].

\bibitem[Kokhlikyan et~al.(2020)Kokhlikyan, Miglani, Martin, Wang, Alsallakh, Reynolds, Melnikov, Kliushkina, Araya, Yan, and Reblitz-Richardson]{kokhlikyan_captum_2020}
Narine Kokhlikyan, Vivek Miglani, Miguel Martin, Edward Wang, Bilal Alsallakh, Jonathan Reynolds, Alexander Melnikov, Natalia Kliushkina, Carlos Araya, Siqi Yan, and Orion Reblitz-Richardson.
\newblock Captum: {A} unified and generic model interpretability library for {PyTorch}, September 2020.
\newblock URL \url{http://arxiv.org/abs/2009.07896}.
\newblock arXiv:2009.07896 [cs].

\end{thebibliography}

\clearpage
\appendix

\section{Cluster path generation pseudocode}

\begin{algorithm}[ht]
\caption{Cluster Formation and Path Generation}
\begin{algorithmic}[1] % The [1] ensures lines are numbered
\Procedure{ClusteringAndPathGeneration}{}
\State \textbf{Input:} Training set $X=\{x_i\in \mathbb{R}^{D}\}_{i=1}^{N}$, network $f_{\theta}$ with layer outputs $f_\theta^{(l)}(x)$, layer set $L$, number of clusters per layer $K>1$
\State \textbf{// Phase 1: Cluster Formation}
\For{each layer $l \in L$}
\State $A \gets f_\theta^{(l)}(X)$ \Comment{Compute activation features}
\State Perform clustering on $A$ using $K$ clusters
\State $C^{(l)} = \{C^{(l)}_0, C^{(l)}_1, \ldots, C^{(l)}_{K-1}\}$ \Comment{Store cluster centers}
\EndFor
\State \textbf{End of Phase 1}
\\
\State \textbf{// Phase 2: Path Generation}
\State \textbf{Input:} A test input $x_\nu$
\For{each layer $l \in L$}
    \State $A \gets f_\theta^{(l)}(x_\nu)$ \Comment{Compute activation features}
    \State $c^{(l)}(x_\nu) \gets \underset{k}{\arg\min} \left\| A^{(l)}(x_\nu) - C^{(l)}_k \right\|^2$ \Comment{Cluster assignment}
\EndFor
\State Construct cluster path: $\gamma(x_\nu) \leftarrow (c^{(0)}, c^{(1)}, \ldots, c^{(L-1)})$ \Comment{Path of cluster IDs}
\State \textbf{Return} $\gamma(x_\nu)$ \Comment{Return path as tuple of cluster IDs}
\State \textbf{End of Phase 2}
\EndProcedure
\end{algorithmic}
\label{algo:path-gen}
\end{algorithm}

\section{Computational and memory cost comparison}\label{sec:computational_complexity}

This section contrasts the asymptotic runtime and storage demands (big-O notation) of cluster path and DkNN when used for post-hoc interpretability.
There are three main parts of this analysis to consider: offline index construction, online per-input query time, and the persistent memory footprint needed to support any necessary data structure.
Offline index construction is the preprocessing run after training your model, in which all training activations are gathered and either clustered (cluster paths) or inserted into a layer-wise Locality-Sensitive Hashing (LSH) table (DkNN).
Its computational cost is a one-time occurrence.
Online per-input query time is the extra computation needed for each inference. 
Cluster paths perform a quick nearest-centroid lookup at each layer, while DkNN hashes the test sample, finds candidate buckets and re-ranks a small neighbor set.
The persistent memory footprint is the data that must be stored to answer queries. 
Cluster paths have centroid matrices per layer, whereas DkNN requires the full set of training activations plus hash keys.

\subsection{Offline index construction}

\textbf{Cluster paths} perform a single sweep over the training set to record layer activations and then apply k-means at every layer $\ell$. 
One iteration of Lloyd's algorithm costs $O(NK_\ell d_\ell)$ and after $I_\ell$ iterations, the total offline cost is $T_{\text{CP-build}}=\sum_{\ell=1}^{L}O(NK_\ell d_\ell I_\ell)=O(NLKdI)$.

\textbf{DkNN} also forwards all $N$ training samples to record their internal representations, but instead of clustering it inserts each vector into $H$ cross-polytope LSH tables, per layer. 
For one table the hash key is calculated using a pseudo-random rotation implemented using the Fast Hadamard Transform in $O(d_\ell\log{d_\ell})$ time. 
Repeating this for all $H$ tables gives a per-activation insertion cost of $O(Hd_\ell\log{d_\ell})$. 
Therefore the overall offline cost is $T_{\text{DkNN-build}}=\sum_{\ell=1}^{L}O\!\bigl(NHd_\ell\log{d_\ell}\bigr)=O(LNHd_{\text{max}}\log{d_{\text{max}}})$ where $d_{\max}=\max_\ell d_\ell$.
Thus both methods are linear in the training-set size, but DkNN replaces the clustering factor $KI$ (e.g. $K=25$, $I=1,000$ yields a factor of $KI=25,000$) with a typically smaller constant $H\log{d}$ (e.g. $H = 200$, $\log{d}\approx9$ for $d=512$, yielding $H\log{d} = 1,800$).

\subsection{Online interpretability query (single test input)}

\textbf{Cluster paths} only need to compute a nearest-centroid lookup at each layer.
Each squared Euclidean norm calculation is $O(d_\ell)$, yielding $T_{\text{CP-query}}=O\!\bigl(LK d\bigr)$.

\textbf{DkNN}, in contrast, hashes the new activation into each of the $H$ tables in $O(Hd_\ell \log{d_\ell})$ and re-ranks the $s\!\ll\!N$ candidates they return. 
The candidates are the subset of training activations that hash to the same key as the query activation in at least one of the $H$ cross-polytope LSH tables. 
For each layer, the query is first rotated and hashed and every hash table then yields the small bucket of stored points with an identical key. 
The union of these buckets, typically $s\!\ll\!N$ points, forms the candidate set $S_\ell$. 
Only these $s$ candidates are subjected to exact cosine (or Euclidean) distance calculations, after which the algorithm keeps the $K$ nearest samples, therefore avoiding a full search among all N training examples.
We check the distance for the $s$ retrieved candidates in $O(sd_\ell)$ and select the $K$ nearest among the candidates using a max-heap in $O(s\log{K})$ 
The final online query time complexity becomes $T_{\text{DkNN-query}}=O\bigl(L(Hd_{\max} \log{d_{\max}}+sd_{\max} + s\log{K} ) \bigr)$, which is independent of N like cluster paths, but linear in both $H$ and the expected bucket size $s$.

\subsection{Memory footprint}

\textbf{Cluster paths} keep only the centroids, more concretely, each of the $L$ layers keeps $K_\ell$ centroids of dimension $d_\ell$.  
With uniform $K$ and $d$ this is
$S_{\text{CP}} = \sum_{\ell=1}^{L} K_\ell d_\ell = O(L K d)$.
For a vision network with $d\!\approx\!512$ and $K\!\le\!50$ the raw array is $L K d \times 4$ bytes which results in approximately $<\!10$ MB when $L\le4$.
For DkNN, each of the N training samples contributes one $d_\ell$-vector per indexed layer, i.e. $O(N L d)$ floats.

\textbf{DkNN} stores a compact key (an integer index of the closest signed basis vector) with cross-polytope LSH and a pointer for that sample in every one of the $H$ tables, for a further $O(N L H)$ words (a word being 4 bytes on a 32-bit architecture or 8 bytes on a 64-bit architecture).
Combining the two gives $S_{\text{DkNN}} = O\bigl(N L (d_{\max} + H)\bigr)$.
With modern datasets ($N\!\sim\!10^6$) and even moderate $L$ this quickly scales to many GB, so the activation store dominates the memory complexity.

\subsection{Practical implications}
Cluster paths front-load work since their $O(NLKdI)$ clustering pass can be slow. 
However, it yields a small, fixed-size auxiliary data storage $\bigl(O(LKd)\bigr)$ that grows irrespective of the dataset $N$. 
Furthermore, we only need $O(LKd)$ extra FLOPs per inference, making the method attractive on devices where RAM and latency are constrained.
DkNN, by contrast, shifts the cost since its build phase is lighter, $O(NLHd\log d)$ with small H, but it must keep every training activation plus H hash entries, an $O\!\bigl(NL(d+H)\bigr)$ store that grows linearly with the dataset and easily reaches many GB.
At inference, DkNN incurs $O\!\bigl(L(Hd\log d+sd+s\log K)\bigr)$ overhead which is reasonable when the expected candidate set s is small.
In summary, cluster paths favor memory-constrained or privacy-sensitive settings where retaining the raw training set is not possible, whereas DkNN suits server-side or research scenarios that can afford large memory. 
Both approaches scale linearly in the number of probed layers $L$ and cluster paths are governed by the tunable centroid term $K$, while DkNN depends on the hash-table size $H$ and candidate size $s$, which in turn reflect dataset size.

\begin{table}[ht]
\centering
\caption{Asymptotic computational and memory requirements for \textbf{cluster paths} and \textbf{DkNN}.
$N$ = \# training examples; $L$ = \# of chosen layers; $d$ = (max) layer-activation dimension;
$K$ = centroids per layer; $I$ = $k$-means iterations; $H$ = LSH tables per layer;
$s$ = expected candidate set size per layer at query time.}
\begin{tabular}{lccc}
\toprule
\textbf{Method} &
\textbf{Offline build time} &
\textbf{Online query time$^{\dagger}$} &
\textbf{Memory} \\[2pt]
\midrule
Cluster paths &
$O\!\bigl(N L K d I\bigr)$ &
$O\!\bigl(L K d\bigr)$ &
$O\!\bigl(L K d\bigr)$ \\[4pt]

DkNN &
$O\!\bigl(N L H d \log d\bigr)$ &
$O\!\bigl(L\,(H d \log d + s d + s \log K)\bigr)$ &
$O\!\bigl(N L (d + H)\bigr)$ \\
\bottomrule
\end{tabular}

\smallskip
\footnotesize{$^{\dagger}$Times are additional to the standard forward pass of the network; both methods still incur the normal inference cost of the underlying model.}
\end{table}

\section{\cifarten spurious cue model explanations}

\subsection{SpuriousCIFAR10 experiment hyperparameters}\label{sec:spurious_cue_hyperparameters}
For our spurious-cue experiments on CIFAR-10, we generated two dataset variants—normal (90\% of images receive a class-consistent patch: red for cat, blue for dog) and corrupted (patch color randomized).  
Patches are 8x8 pixels (one-quarter of the 32x32 image) placed at a uniformly random location, and all randomness (patch placement, train/test splits, clustering subsampling) is seeded to 0 for full reproducibility.

We trained the CIFAR$\_$CatDog model (detailed in Section~\ref{sec:architectures}) for 100 epochs with batch size=16, using the Adam optimizer at a learning rate=0.001 and cross-entropy loss.  All training and inference ran on the MPS backend (device='mps'), and we saved model checkpoints at the end of training.

To extract cluster paths, we recorded activations from four layers (`Conv Block3', FC1, FC2, and the final layer's logits) on a random subset of 10,000 training samples. 
We then applied k-Means (k-means++ init, 10 restarts) with [2, 2, 2, 3] clusters per layer, running up to 300 iterations each.

Finally, we quantified decision-alignment faithfulness by one-hot encoding each sample's path and training a 100-tree Random Forest on an 80/20 split of these features.

\FloatBarrier
\section{Additional \celeba hair attribute cluster path analysis}

\begin{figure*}[ht]
    \centering
    \includegraphics[%
        trim=0pt 0pt 0pt 40pt,%
        clip,%
        width=0.9\linewidth%
    ]{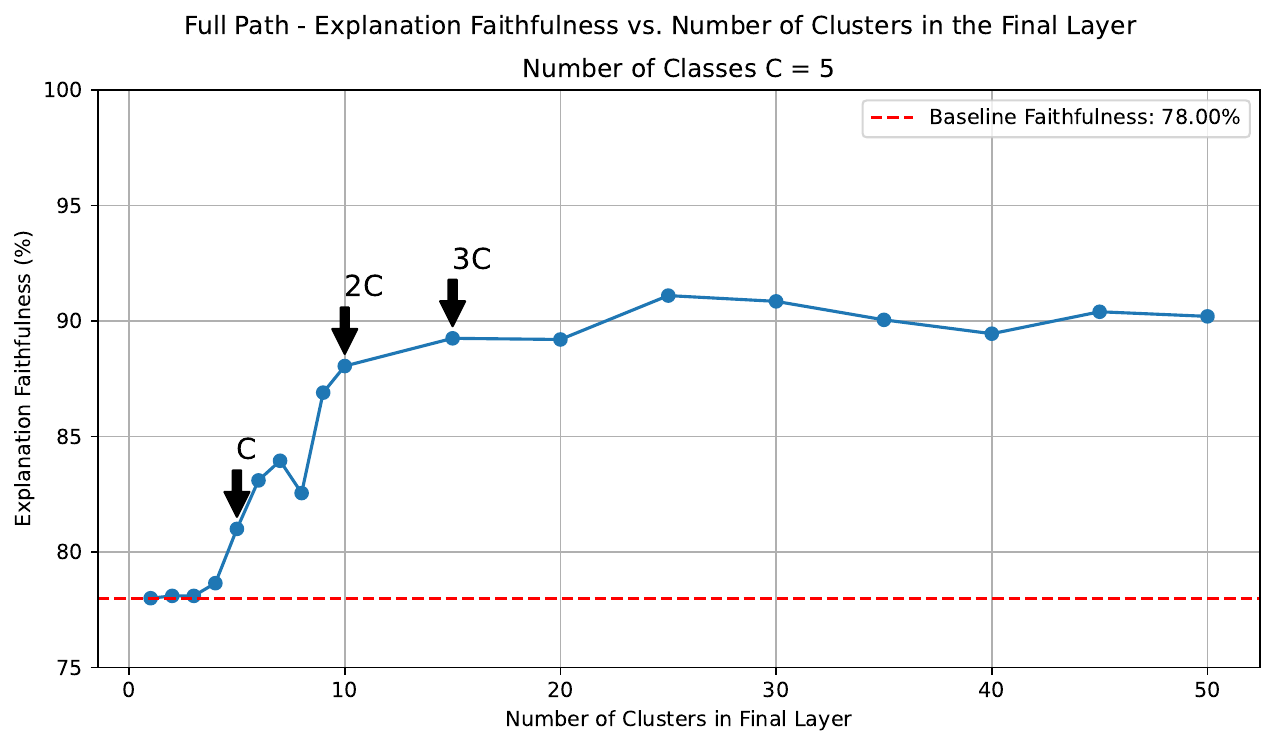}
    \caption{\textit{Full Path} Explanation Faithfulness vs. Number of Clusters in the Final Layer: This figure shows how decision-alignment faithfulness varies with the number of clusters in the final layer, when predicting the neural network's decisions considering \textit{both intermediate and final layers} for the \texttt{HairAttributeCNN} model.}
    \label{fig:faithfulness-full-path}
\end{figure*}

\begin{figure*}[ht]
    \centering
    \includegraphics[%
        trim=0pt 0pt 0pt 40pt,%
        clip,%
        width=0.9\linewidth%
    ]
    {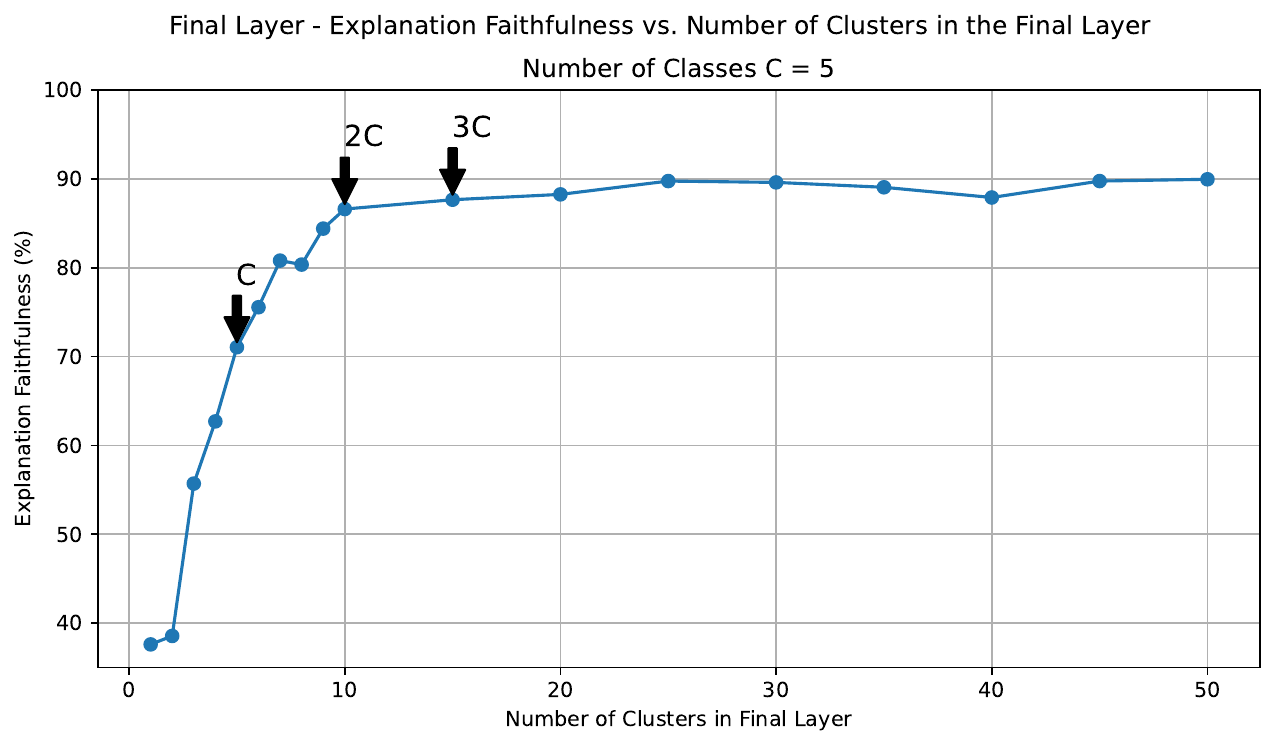}
    \caption{\textit{Final Layer} Explanation Faithfulness vs. Number of Clusters: This figure illustrates how faithfulness varies with the number of clusters in the final layer when predicting the neural network's decisions, \textit{exclusively considering the final layer's clusters} for the \texttt{HairAttributeCNN} model.}
    \label{fig:faithfulness-final-layer}
\end{figure*}

\subsection{\celeba experiment hyperparameters}\label{sec:celeba_hyperparameters}
For our \celeba hair-attribute experiments, we first resized all images to 64x64 and normalized them with ImageNet statistics (mean=[0.485, 0.456, 0.406], std= 0.229, 0.224, 0.225]). 
We used a 5-class ``HairAttributeCNN'' (dropout=0.2) and loaded the best-performing checkpoint after 50 epochs of training with Adam (learning rate=0.001), batch size=128, on the MPS backend (device=`mps'). 
Train/val/test splits were taken from the standard \celeba partition files, and all random seeds were fixed to 0 for reproducibility.

For decision-path clustering, we extracted activations from four layers: flattened conv-out (64x8x8$\to$4,096-dim), FC1 (128-dim), FC2 (64-dim), and the 5-way output logits, over a random subset of 10,000 training samples (seed=0). 
We then ran k-Means with $k=5$ clusters per layer, k-means++ initialization, 10 restarts, and a maximum of 300 iterations. 
Where applicable, we used 80/20 splits for training simple surrogate models (Random Forests) to measure path faithfulness, and we evaluated robustness on the first 1,000 validation samples.

\begin{figure*}[t]
    \centering
    \includegraphics[%
        trim=0pt 0pt 0pt 30pt,%
        clip,%
        width=1.0\linewidth%
    ]{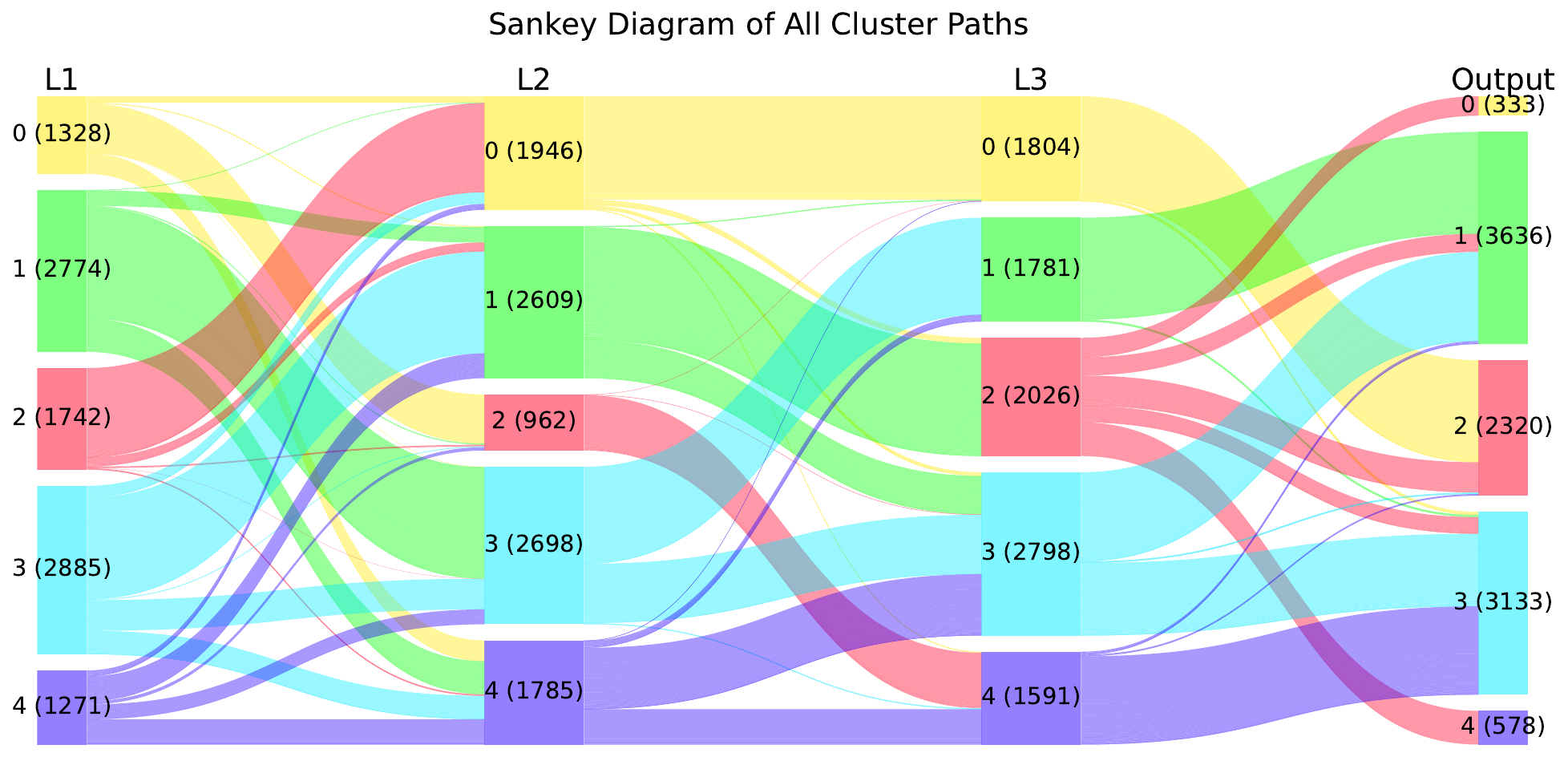}
    \caption{A Sankey diagram illustrating the relative volume of images through that cluster in the cluster path. Each cluster ID is color-coded for ease of visualization, specifically yellow for cluster ID 0, green for 1, red for 2, cyan for 3, and purple for 4. Each number in a row represents a cluster ID for a given layer or column in the diagram. The number in parentheses next to each cluster ID represents the cumulative number of samples visiting that cluster. }
    \label{fig:sankey_all_paths}
    % \todo{#######add in ppt slide edit showing some cluster path examples pointing to the flow int he diagram########}
\end{figure*}

\subsection{Sankey diagrams of cluster paths} \label{sec:sankey}
We enhance our cluster paths analysis using Sankey diagrams.
These diagrams provide a way to represent sample distribution flow from one cluster ID in layer $l$ to $l+1$, in this case from layer `L1,' to `L2,' to `L3,' and finally to the output layer `Output.'
As we can see in Figure~\ref{fig:sankey_all_paths}, dominant, high-volume bands reveal primary decision routes, whereas thin, branching bands demonstrate rare alternatives and divergence points, hinting of sub-types the model may find ambiguous. 
This birds-eye view highlights which paths drive predictions and clarifies how the network's feature space is organized.

\FloatBarrier
\subsection{\celeba Cluster path distribution}
\begin{figure*}[ht]
    \centering
    \includegraphics[%
        trim=0pt 0pt 0pt 20pt,%
        clip,%
        width=1.0\linewidth%
    ]
    {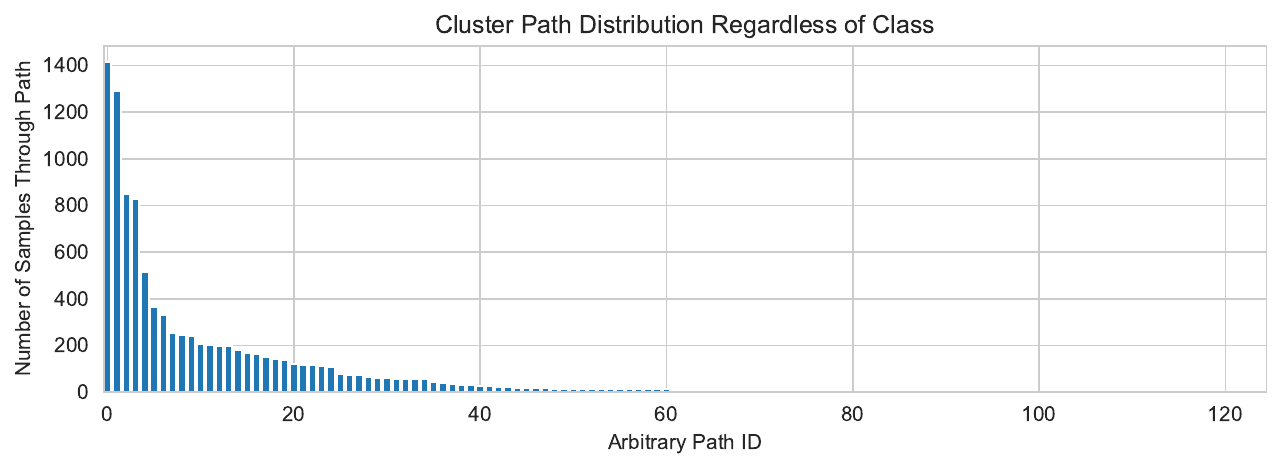}
    \caption{Counts of CelebA samples per cluster path, with path IDs anonymized for plotting and ordered by frequency. The steep drop-off reveals a heavy-tailed, power-law-like distribution in which a few dominant paths carry most images, while many others are rarely traversed.}
    \label{fig:cluster_path_distribution_all}
\end{figure*}

To provide an initial overview of how many images traverse each cluster path, we present their distribution across all predicted classes in Fig.~\ref{fig:cluster_path_distribution_all}.
Of the 131 unique paths traversed, the top 10 and 40 most popular cluster paths contain approximately 60\% and 93\% of samples, respectively.
This skew indicates that a small number of paths heavily utilized, suggesting these paths may capture dominant patterns or features that are critical for the model's decision-making processes.
The existence of many paths with only a few samples each may indicate rare features within the dataset that are not as generalizable as those captured by the more frequented paths.
Additionally, the presence of such sparse paths may highlight the need for further model tuning or training with a more diverse dataset to enhance the model's ability to generalize across a broader range of inputs.

\subsection{\celeba qualitative comparison to pixel-attribution maps}\label{sec:qualitative-comparison}
\begin{figure}[!ht]
    \centering
    \includegraphics[width=1.0\textwidth]{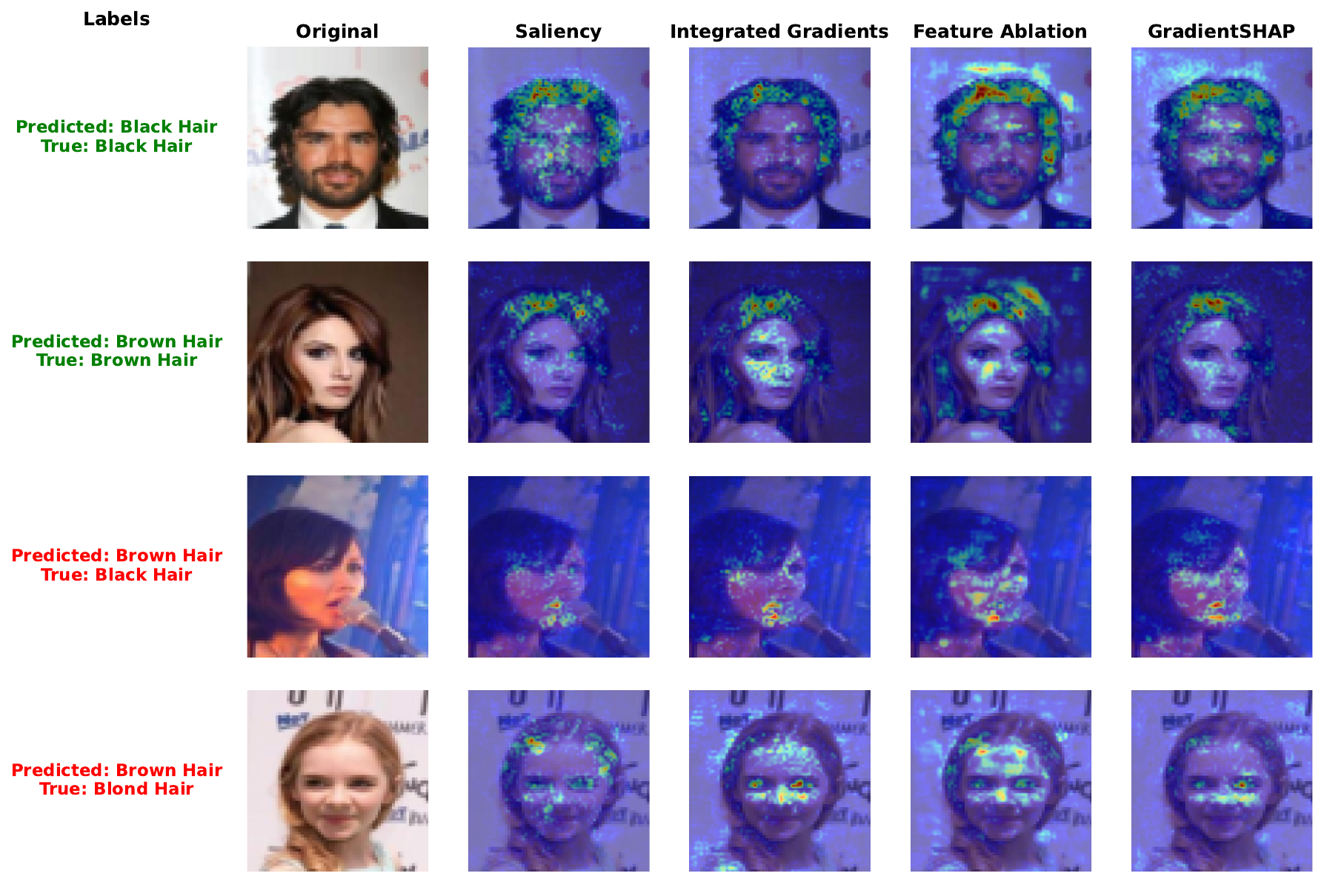}
    \caption{Left to right: predicted/true labels, original image, Saliency, IG, FA, and GS.  
    Rows 1-2 are correctly classified and rows 3-4 are misclassified.  
    Pixel methods applied to correctly predicted images focus on hair cues, whereas incorrectly predicted images focus on different parts of the face like eyes, cheeks, and chin.}
    \label{fig:qualitative-attribution}
\end{figure}

We compare and contrast cluster paths with four SOTA gradient-based attribution baselines, implemented with the Captum interpretability library~\cite{kokhlikyan_captum_2020}.
These attribution methods calculate gradients with respect to input features to explain model predictions.
Our methods for comparison include Vanilla Gradient~\cite{simonyan_deep_2014}, 
Integrated Gradients~\cite{sundararajan_axiomatic_2017},
Feature Ablation, and GradientSHAP~\cite{lundberg_unified_2017}. 
We randomly select four validation images: two that the network classifies correctly and two that it misclassifies.
For each image we compute the attributions listed above (Fig.~\ref{fig:qualitative-attribution}).  
Feature maps outline high-gradient shown with colors closer to red. 
For the correct examples, the gradients generally appear to highlight the regions of hair.
However, for the misclassified samples, all attributions focus predominantly on facial regions-eyes, noses and mouths-rather than hair pixels (Fig.~\ref{fig:qualitative-attribution}, rows~3-4).  
Cluster paths, however, are limited by visual inspection of a cluster path to determine the commonality among the grouped images. 
Therefore, pixel attributions can readily determine \emph{where} the network looks, however cluster paths can reveal \emph{what} the network conceptually attends to.

\FloatBarrier
\section{Vision transformer cluster path results} \label{sec:transformer_clusterpaths}

\subsection{Vision transformer experiment hyperparameters}\label{sec:vit_hyperparameters}
For our ViT interpretability experiments, we used the timm ``vit-base-patch16-224'' model pretrained on ImageNet-21K and fine-tuned on ImageNet-1K. 
All images were rescaled to 224x224 and normalized with the model's default mean=[0.485, 0.456, 0.406] and std=[0.229, 0.224, 0.225]. 
We sampled roughly 100 training images per class (1,000 classes) plus the full validation set (50K images) for a total of 150K images. 
We fixed all seeds (Python, NumPy, Torch) to 0 for reproducibility. 
Inference and activation extraction ran on MPS, with no data augmentation beyond resizing.

To build decision-paths, we registered forward hooks on four ViT layers: blocks 0, 4, 8 and the final classifier head.
Then for each of the 150K images, we extract the mean-pooled CLS token into a 768-dim representation at each block and a 1,000-dim logit vector at the head. 
No additional transforms or normalization were applied to these activations.
Finally, we ran k-Means (k={20 for blocks, 100 for head}, k-means++ init, 10 restarts, 1,000 max-iter, seed=0).

\begin{figure}[!ht]
    \centering
        \includegraphics[%
        trim=0pt 0pt 690pt 30pt,%
        clip,%
        width=1.0\linewidth%
    ]{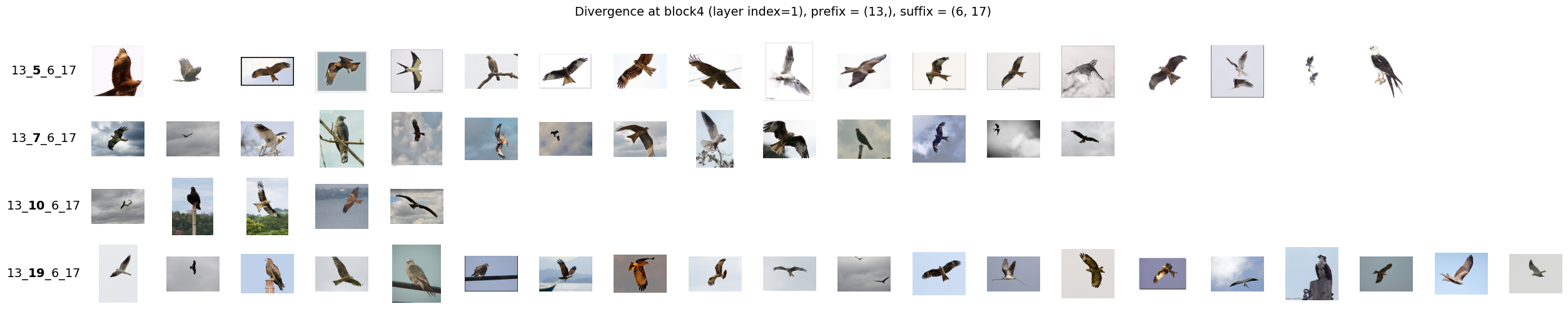}
    \caption{\textbf{Cluster path 13$\to$X$\to$6$\to$17}. Applying our divergence analysis to `Block4' clusters with cluster path prefix `13' and suffix `6$\to$17.' 
    \chatgpt identified the following concepts per layer:
Cluster 5: Clear-sky backdrop,
Cluster 7: Cloudy backgrounds,
Cluster 10: Natural perches, and 
Cluster 19: Manmade perches.}
    \label{fig:transformer_13_X_6_17}
\end{figure}

\begin{figure}[!ht]
    \centering
        \includegraphics[%
        trim=0pt 0pt 660pt 30pt,%
        clip,%
        width=1.0\linewidth%
    ]{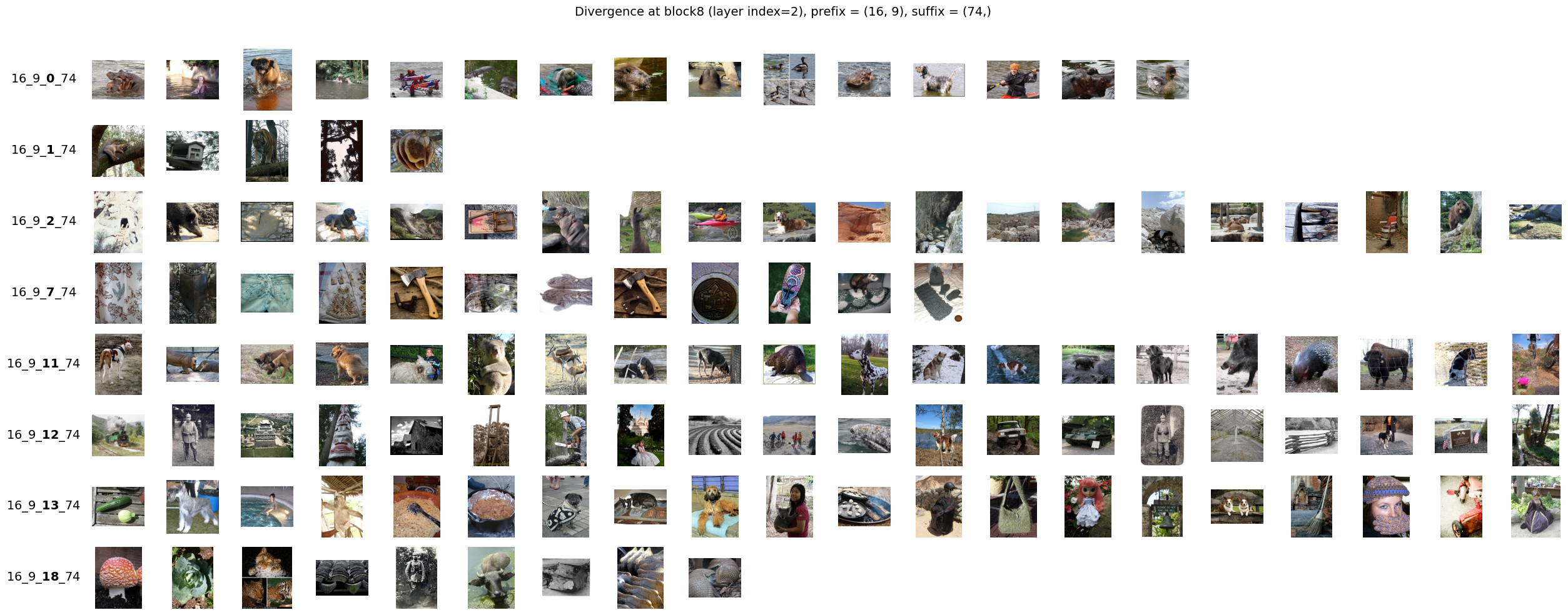}
    \caption{\textbf{Cluster path 16$\to$9$\to$X$\to$74}. Applying our divergence analysis to `Block8' clusters with cluster path prefix `16$\to$9' and suffix `74.' 
    \chatgpt identified the following concepts per layer:
Cluster 0: Water environments,
Cluster 1: Arboreal subjects,
Cluster 2: Textured ground surfaces,
Cluster 7: Isolated object close-ups,
Cluster 11: Side-profile animals,
Cluster 12: Vertical structures,
Cluster 13: Domestic/human-scale objects, and 
Cluster 18: Repetitive patterns.
}
    \label{fig:transformer_16_9_X_74}
\end{figure}

\begin{figure}[!ht]
    \centering
        \includegraphics[%
        trim=0pt 0pt 600pt 20pt,%
        clip,%
        width=1.0\linewidth%
    ]{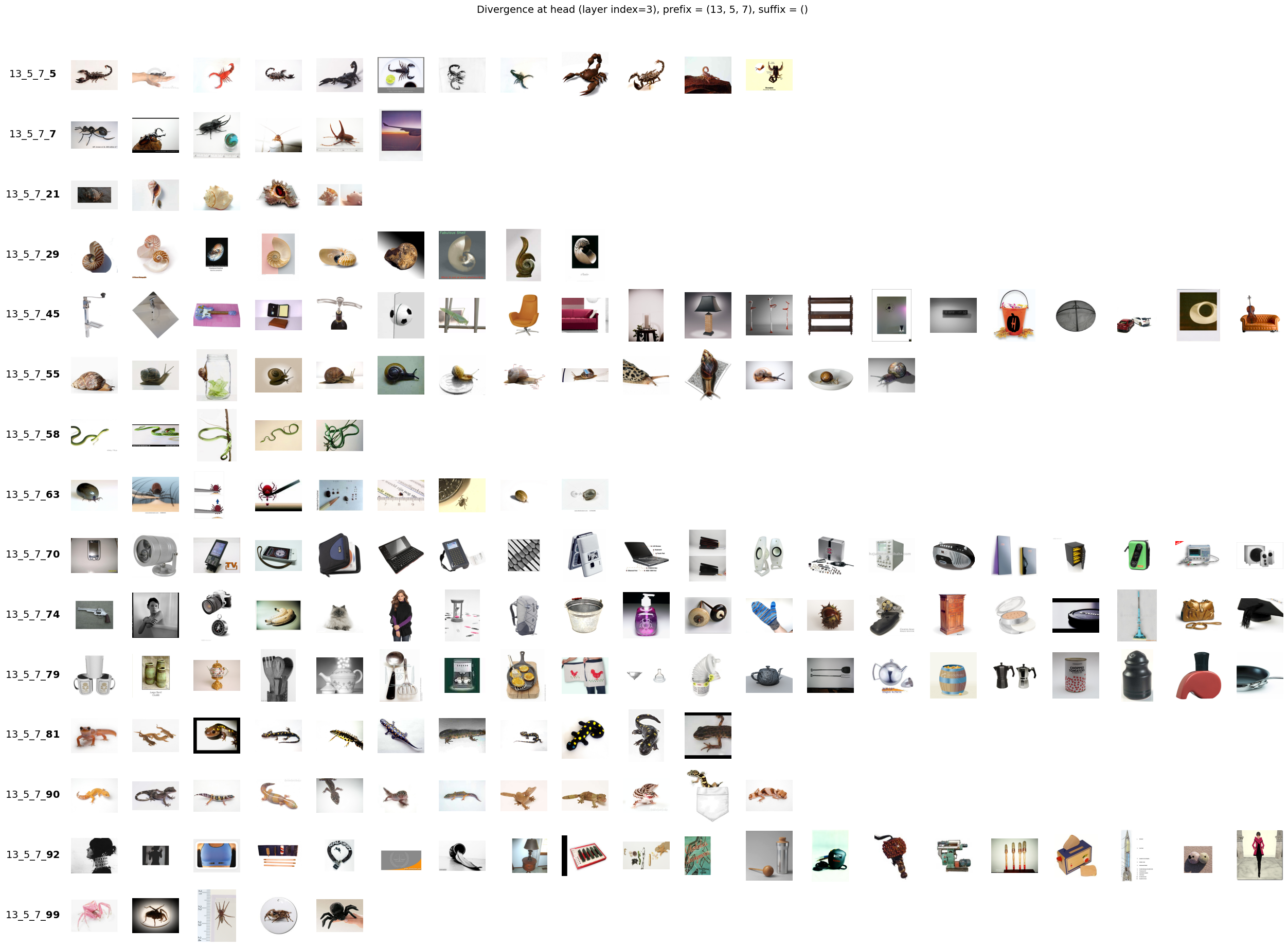}
    \caption{Cluster path \textbf{13$\to$5$\to$7$\to$X}. Applying our divergence analysis to `Head' (or output layer) clusters with cluster path prefix `13$\to$5$\to$7.' 
    \chatgpt identified the following concepts per layer:
Cluster 5: Arched scorpion silhouettes,
Cluster 7: Slender crawling insects,
Cluster 21: Beige spiral shells,
Cluster 29: Chambered-nautilus cross-sections,
Cluster 45: Furniture and fixtures,
Cluster 55: Snails with extended bodies,
Cluster 58: Green coiled snakes,
Cluster 63: Magnified insect close-ups,
Cluster 70: Electronic gadgets,
Cluster 74: Human and animal figures,
Cluster 79: Kitchen vessels,
Cluster 81: Reptile-like creatures,
Cluster 90: Wet-skinned amphibians,
Cluster 92: Minimalist graphic designs, and
Cluster 99: Spindly legged spiders.
}
    \label{fig:transformer_13_5_7_X}
\end{figure}

\begin{figure}[!ht]
    \centering
        \includegraphics[%
        trim=0pt 0pt 0pt 60pt,%
        clip,%
        width=1.0\linewidth%
    ]{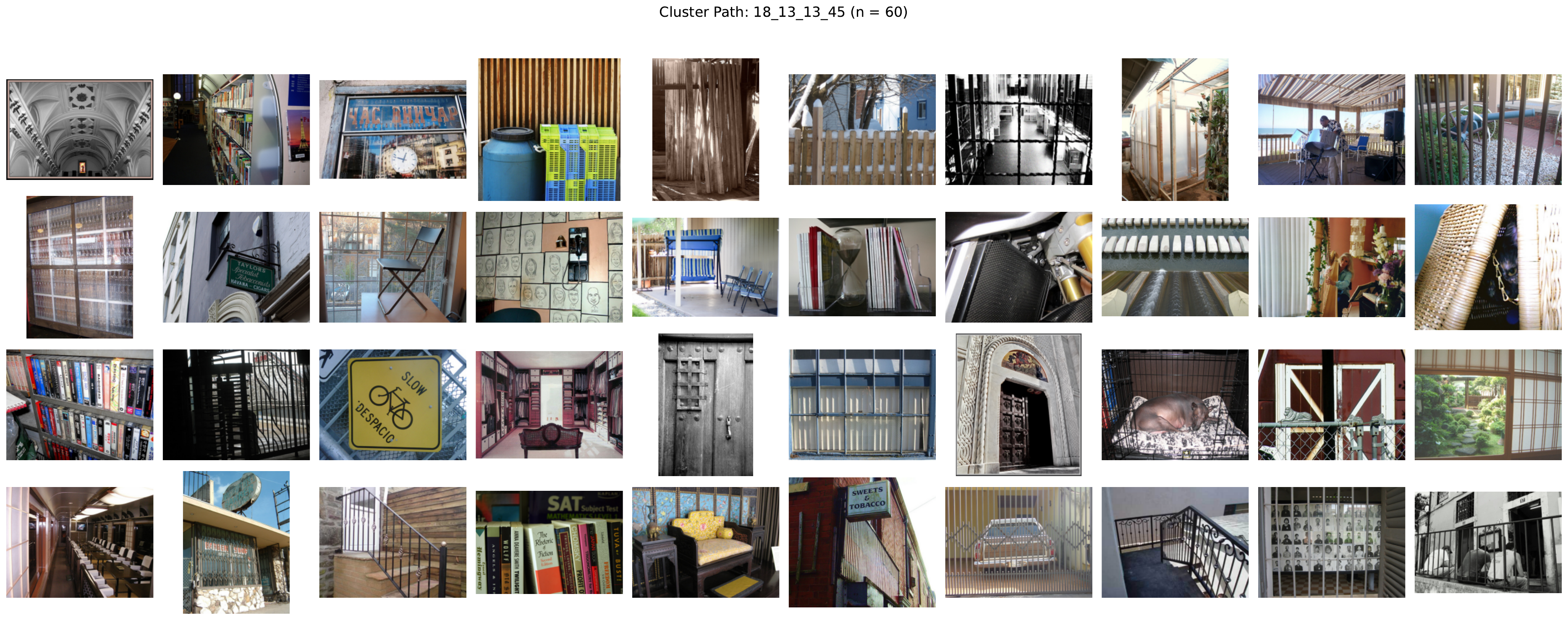}
    \caption{\textbf{Cluster path 18$\to$13$\to$13$\to$45}. This image-grid shows 40 inputs that follow the same cluster path. 
    After divergence and NLP analysis, the identified top-3 concepts per layer are ``\textit{forms | shapes | light $\to$ domestic | urban | contrast urban $\to$ close | domestic | warm $\to$ urban | transportation | community interactions}.'' The | operator represents an OR operation as there can be more than one key word or phrase to describe a cluster's concept.}
    \label{fig:transformer_18_13_13_45}
\end{figure}

\begin{figure}[!ht]
    \centering
        \includegraphics[%
        trim=0pt 0pt 0pt 60pt,%
        clip,%
        width=1.0\linewidth%
    ]{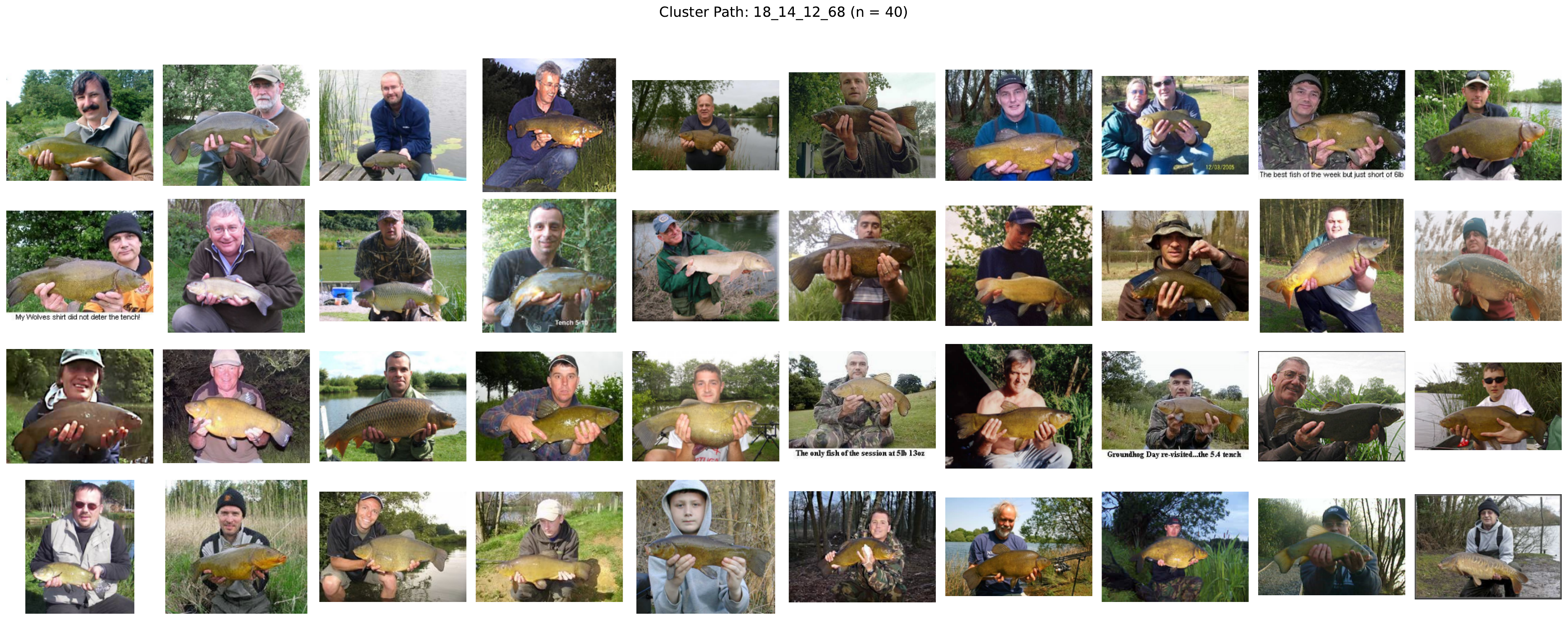}
    \caption{\textbf{Cluster path 18$\to$14$\to$12$\to$68}. This image-grid shows 40 inputs that follow the same cluster path. 
    After divergence and NLP analysis, the identified top-3 concepts per layer are ``\textit{forms | shapes | light $\to$ texture | shapes | dark $\to$ human interaction | interaction | human $\to$ outdoor reflection | leisure | familial}.''
    The | operator represents an OR operation as there can be more than one key word or phrase to describe a cluster's concept.
    }
    \label{fig:transformer_18_14_12_68}
\end{figure}

\FloatBarrier
\section{Architectures}\label{sec:architectures}

\begin{table}[H]
\centering
\caption{Summary of the \texttt{CIFAR\_CatDog} architecture used in the SpuriousCIFAR10 experiment.  Layers are given in PyTorch notation.}
\label{tab:cifar_catdog_arch_summary}
\begin{tabular}{lp{10cm}}
\toprule
\textbf{Layer/Block} & \textbf{Description} \\
\midrule
\textbf{Conv Block 1} & Conv2d: $3 \rightarrow 4$ (3×3, stride=1, pad=1) $\rightarrow$ ReLU. \\[1ex]
\textbf{Conv Block 2} & Conv2d: $4 \rightarrow 8$ (3×3, stride=1, pad=1) $\rightarrow$ ReLU $\rightarrow$ MaxPool2d (2×2, stride=2). \\[1ex]

\textbf{Conv Block 3} & Conv2d: $8 \rightarrow 16$ (3×3, stride=1, pad=1) $\rightarrow$ ReLU $\rightarrow$ MaxPool2d (2×2, stride=2). \\[1ex]
\textbf{Final Layers} & Flatten (from $16\times8\times8$ features) $\rightarrow$ FC1: Linear($1024 \rightarrow 32$) $\rightarrow$ ReLU $\rightarrow$ FC2: Linear($32 \rightarrow 16$) $\rightarrow$ ReLU $\rightarrow$ FC3: Linear($16 \rightarrow 2$). \\
\bottomrule
\end{tabular}
\end{table}

\begin{table}[ht]
\centering
\caption[Summary of the \svhn architecture]{Summary of the \svhn architecture, described in terms of the PyTorch library.}
\label{tab:svhn_arch_summary}
\begin{tabular}{lp{10cm}}
\toprule
\textbf{Layer/Block} & \textbf{Description} \\
\midrule
\textbf{Conv Block 1} & Conv2d: 3 $\rightarrow$ 16 (3×3, stride=1, pad=1) $\rightarrow$ BatchNorm2d (16) $\rightarrow$ ReLU $\rightarrow$ MaxPool2d (2×2, stride=2). \\[1ex]
\textbf{Conv Block 2} & Conv2d: 16 $\rightarrow$ 64 (3×3, stride=1, pad=1) $\rightarrow$ BatchNorm2d (64) $\rightarrow$ ReLU $\rightarrow$ MaxPool2d (2×2, stride=2). \\[1ex]
\textbf{Conv Block 3} & Conv2d: 64 $\rightarrow$ 128 (3×3, stride=1, pad=1) $\rightarrow$ BatchNorm2d (128) $\rightarrow$ ReLU $\rightarrow$ MaxPool2d (2×2, stride=2). \\[1ex]
\textbf{Final Layers} & Flatten (from 4×4×128 features) $\rightarrow$ FC1: Linear(4×4×128 $\rightarrow$ 256) $\rightarrow$ ReLU $\rightarrow$ Dropout (0.1) $\rightarrow$ FC2: Linear(256 $\rightarrow$ 128) $\rightarrow$ ReLU $\rightarrow$ Dropout (0.1) $\rightarrow$ FC3: Linear(128 $\rightarrow$ 10). \\
\bottomrule
\end{tabular}
\end{table}

\begin{table}[ht]
\centering
\caption[Summary of the \texttt{HairAttributeCNN} (\celeba) architecture]{Summary of the \texttt{HairAttributeCNN} (\celeba) architecture, described in terms of the PyTorch library.}
\label{tab:hair_attr_arch_summary}
\begin{tabular}{lp{10cm}}
\toprule
\textbf{Layer/Block} & \textbf{Description} \\
\midrule
\textbf{Conv Block 1} & Conv2d: 3 $\rightarrow$ 16 (3×3, stride=1, pad=1) $\rightarrow$ ReLU $\rightarrow$ MaxPool2d (2×2, stride=2). \\[1ex]
\textbf{Conv Block 2} & Conv2d: 16 $\rightarrow$ 32 (3×3, stride=1, pad=1) $\rightarrow$ ReLU $\rightarrow$ MaxPool2d (2×2, stride=2). \\[1ex]
\textbf{Conv Block 3} & Conv2d: 32 $\rightarrow$ 64 (3×3, stride=1, pad=1) $\rightarrow$ ReLU $\rightarrow$ MaxPool2d (2×2, stride=2). \\[1ex]
\textbf{Final Layers} & Flatten (from 64$\times$8$\times$8 features) $\rightarrow$ Dropout (0.2) $\rightarrow$ FC1: Linear(64$\times$8$\times$8 $\rightarrow$ 128) $\rightarrow$ ReLU $\rightarrow$ Dropout (0.2) $\rightarrow$ FC2: Linear(128 $\rightarrow$ 64) $\rightarrow$ ReLU $\rightarrow$ FC3: Linear(64 $\rightarrow$ 5). \\
\bottomrule
\end{tabular}\label{tab:haircnn}
\end{table}

\begin{table}[ht]
\centering
\caption[Summary of the \cifarten Residual Network architecture]{Summary of the \cifarten Residual Network architecture, described in terms of the PyTorch library.}
\label{tab:resnet_arch_summary}
\begin{tabular}{lp{10cm}}
\toprule
\textbf{Layer/Block} & \textbf{Description} \\
\midrule
\textbf{Conv Block 1} & Conv2d: 3 $\rightarrow$ 64 (3×3, stride=1, pad=1, no bias) $\rightarrow$ BatchNorm2d (64, momentum=0.9) $\rightarrow$ ReLU. \\[1ex]
\textbf{Conv Block 2} & Conv2d: 64 $\rightarrow$ 128 (3×3, stride=1, pad=1, no bias) $\rightarrow$ BatchNorm2d (128, momentum=0.9) $\rightarrow$ ReLU $\rightarrow$ MaxPool2d (2×2, stride=2). \\[1ex]
\textbf{Residual Block} & (128 channels) Two 3×3 Conv2d layers with BatchNorm2d (momentum=0.9) and ReLU activations, connected by a skip connection (with optional downsampling if stride $\neq$ 1), as defined by He et al. \\[1ex]
\textbf{Conv Block 3} & Conv2d: 128 $\rightarrow$ 256 (3×3, stride=1, pad=1, no bias) $\rightarrow$ BatchNorm2d (256, momentum=0.9) $\rightarrow$ ReLU $\rightarrow$ MaxPool2d (2×2, stride=2). \\[1ex]
\textbf{Conv Block 4} & Conv2d: 256 $\rightarrow$ 256 (3×3, stride=1, pad=1, no bias) $\rightarrow$ BatchNorm2d (256, momentum=0.9) $\rightarrow$ ReLU $\rightarrow$ MaxPool2d (2×2, stride=2). \\[1ex]
\textbf{Residual Block} & (256 channels) Two 3×3 Conv2d layers with BatchNorm2d (momentum=0.9) and ReLU, connected by a skip connection. \\[1ex]
\textbf{Final Layers} & ReLU $\rightarrow$ MaxPool2d (2×2, stride=2) $\rightarrow$ Flatten $\rightarrow$ FC1: 1024 $\rightarrow$ 128 (ReLU) $\rightarrow$ FC2: 128 $\rightarrow$ 64 (ReLU) $\rightarrow$ FC3: 64 $\rightarrow$ 10. \\
\bottomrule
\end{tabular}
\end{table}

\subsection{Vision transformer (ViT)} \label{sec:transformer_architecture}
The ViT-Base/Patch-16/224 vision transformer~\cite{dosovitskiy_image_2020} first tokenized 224 x 224 RGB images into non-overlapping 16 x 16 patches, projects them linearly into a $d=768$ dimensional embedding via a Conv2d patch-embedding layer that then appends learnable positional embeddings.
A sequence of $L=12$ encoder blocks process these $N_{\text{patches}}=(224/16)^2=196$ tokens.
Each block consists of a LayerNorm, multi-head self-attention (with 12 heads, each of 64 dimensions, implemented with a $3d \to 3d$ Query-Key-Value linear layer and output projection, residual connections, followed by a two-layer MLP of width $4d=3{,}072$ with GELU activation and dropout.
After the final block, a global classification token's representation is layer-normalized and fed through a linear head to produce 1,000 ImageNet logits.

The ViT-Base model achieves 93\% top-1 training accuracy and 85\% top-1 validation accuracy. 
The extracted activations from the four chosen layers each have dimension $d_\ell\in\{768,768,768,1{,}000\}$, respectively.
We use $50{,}000$ validation set samples and a uniform random sampling of 100 training samples from each class of the $1{,}000$ classes, resulting in a total of $150{,}000$ samples for clustering each layer's activations.

\section{Computational details} \label{sec:computational_details}
All experiments were carried out on a single MacBook Pro equipped with an Apple M2 Max chip and 32 GB of RAM, running macOS and Python 3.10 with PyTorch 2.6.0.  We instantiated the pretrained VisionTransformer via ``model = timm.create$\_$model(`vit$\_$base$\_$patch16$\_$224', pretrained=True)'' (timm package version 1.0.15).

Clustering is sufficiently fast for practitioners as the CelebA forward pass for $10{,}000$ samples takes $<20$ seconds and clustering takes $<30$ seconds.
At scale, bottlenecks of cluster paths include forward passes to store activations at each layer of interest.
The ViT forward pass for $150{,}000$ ImageNet samples (224x224x3 RGB images) took approximately under 1.5 hours and clustering takes $<2$ minutes.

\section{Asset sources and licensing}\label{sec:asset_licensing}
\begin{enumerate}
\item CIFAR-10: We use the CIFAR-10 dataset from \cite{krizhevsky_learning_2009}, distributed under the MIT license (\url{https://www.cs.toronto.edu/~kriz/cifar.html}).

\item CelebA: We use the CelebA face-attribute dataset (Liu et al., 2015), ``CelebA Research-Only License (MMLAB)'' for non-commercial research (\url{https://mmlab.ie.cuhk.edu.hk/projects/CelebA.html} and \url{https://datasetninja.com/celeb-faces-attributes}).

\item ImageNet: Our ViT experiments use the ILSVRC2012 subset of ImageNet~\cite{deng_imagenet_nodate}, which is ``available for free to researchers for non-commercial use'' (\url{https://image-net.org/index.php}).

\item ViT checkpoint: We load a publicly released ViT-Base/patch-16/224 model checkpoint from the official repository, distributed under the Apache 2.0 license(\url{https://github.com/rwightman/pytorch-image-models/blob/master/LICENSE}).

\item PyTorch: All training and evaluation code is built on PyTorch (BSD 3-Clause License - \url{https://github.com/pytorch/pytorch/blob/main/LICENSE}). 
\end{enumerate}

\end{document}